\documentclass[acmtog]{acmart}
\acmSubmissionID{354}

\AtBeginDocument{%
  \providecommand\BibTeX{{%
    \normalfont B\kern-0.5em{\scshape i\kern-0.25em b}\kern-0.8em\TeX}}}

\setcopyright{rightsretained}
\acmJournal{TOG}
\acmYear{2026} \acmVolume{45} \acmNumber{4} \acmArticle{160}
\acmMonth{7} \acmDOI{10.1145/3811308}

\setcitestyle{square}
\usepackage{amsmath}
\usepackage{multirow}

\usepackage[noend]{algpseudocode}
\usepackage{float}
  \floatstyle{ruled}
  \newfloat{algorithm}{thp}{alg}
\floatname{algorithm}{Algorithm}
\usepackage{algorithmicx}
\usepackage{hyperref}
\begin{document}

\title{MOCHI: Motion Enhancement of Collaborative Human-object Interactions}


\author{Jiye Lee}
\affiliation{%
  \department{Department of Computer Science and Engineering}
  \institution{Seoul National University}
  \country{South Korea}}
\city{Seoul}
\email{jiyelee@imo.snu.ac.kr}

\author{Yonghun Choi}
\affiliation{%
  \department{Department of Computer Science and Engineering}
  \institution{Seoul National University}
  \country{South Korea}}
\city{Seoul}
\email{yongdori@imo.snu.ac.kr}

\author{Jungdam Won}
\authornote{Corresponding author}
\affiliation{%
  \department{Department of Computer Science and Engineering}
  \institution{Seoul National University}
  \country{South Korea}}
\city{Seoul}
\email{jungdam@imo.snu.ac.kr}

\begin{teaserfigure}
    \begin{center}
    \includegraphics[width=1.0\linewidth,trim={0 1.5cm 0 0cm} ]{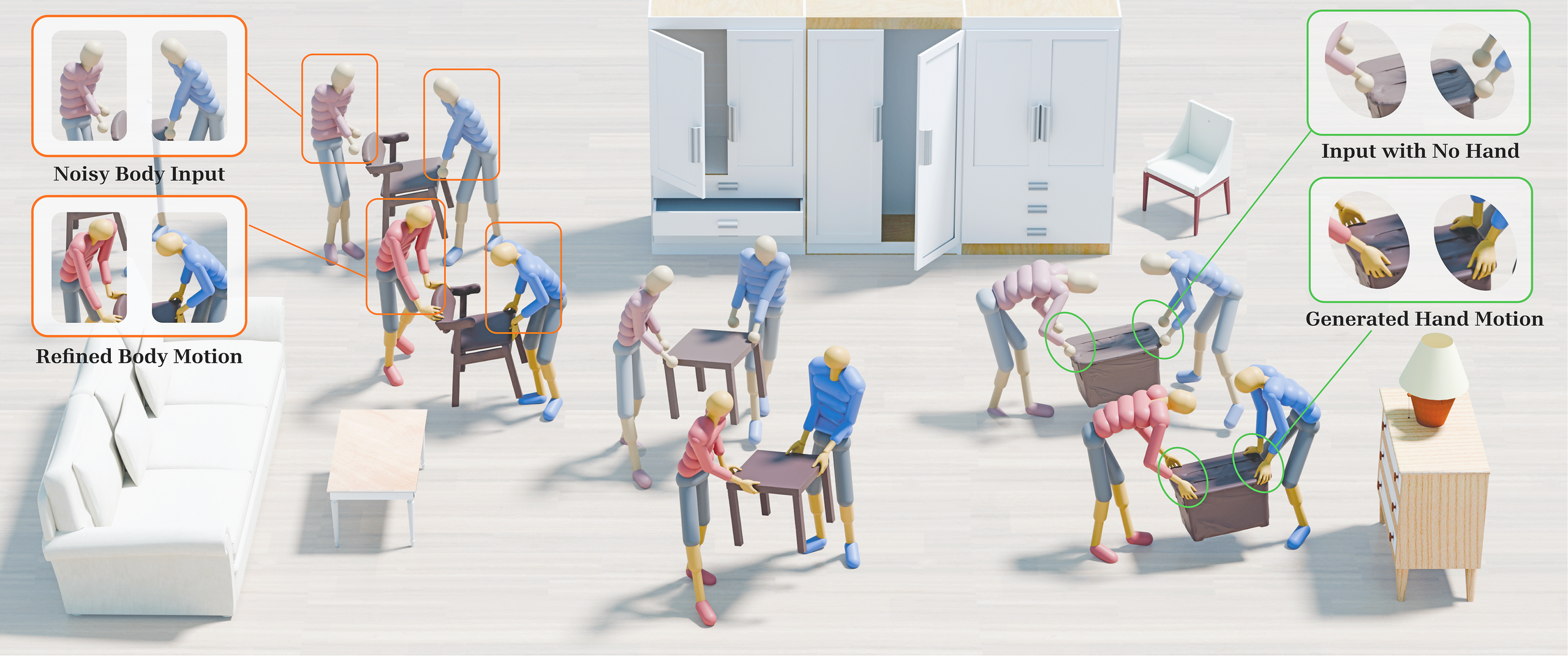}
    \end{center}
\caption{Our method enhances noisy multi-human object interaction (MHOI) data (\textcolor[RGB]{248,176,185}{\textbf{soft pink}} and \textcolor[RGB]{160,210,236}{\textbf{skyblue}}) into clean MHOI data (\textcolor[RGB]{227,112,140}{\textbf{pink}} and \textcolor[RGB]{111,155,210}{\textbf{blue}}), automatically generating plausible hand and finger motions while preserving the original interaction semantics. The method is applicable to diverse scenarios and objects, including lifting a chair (left), moving a table (middle), and placing a box down (right). 
Project page: \url{https://jiyewise.github.io/projects/MOCHI/}.
}
\end{teaserfigure}


\begin{abstract}
Collaborative human-object interaction shows dynamic and complex movements that require mutual anticipation and continuous adjustment between participants and the shared object. 
Understanding and modeling such collaborative multi-human object interaction (MHOI) scenarios requires high-quality data acquisition as a foundational step, however, this is a challenging task due to the inherent complexity of MHOI scenarios where human-human and human-object interactions occur simultaneously. 
Such complexity leads to noisy MHOI captures characterized by several artifacts: contact misalignment between hands and objects, motion jitter and temporal inconsistencies in the captured sequences, and missing or incomplete finger-level articulation details.
To address these challenges, we present MOCHI (\textbf{MO}tion Enhancement of \textbf{C}ollaborative \textbf{H}uman-object \textbf{I}nteractions), a two-stage framework for enhancing noisy MHOI data. 
Our approach first generates physically plausible hand grasps through optimization from noisy body input, producing grasps that are both physically plausible and semantically consistent with the body pose, where these optimized grasps are extended into complete hand-object interaction sequences. Consequently, the full-body motion for all participants are refined through a diffusion-based noise optimization framework that uses single-person motion priors. 
During the optimization process, we introduce optimization objectives to encode human-object and human-human interaction information within these single-person priors.
Experimental results demonstrate the effectiveness of our pipeline across diverse MHOI data, either acquired by existing capture methods or synthesized by generative models.
We further show robustness of our system across varying numbers of participants and types of interactions, and demonstrate various applications including keyframe-based MHOI creation and data augmentation through varying object geometries. 

\end{abstract}



\maketitle

\section{Introduction}
\label{sec:intro}


Collaborative human-object interaction scenarios, such as carrying a couch through a narrow hallway, assembling furniture together, or loading a box into a car, are routine activities in human life.
Yet despite their commonality, these scenarios involve complex coordination patterns: the participants' actions must be carefully synchronized around the shared object, requiring fluid communication, mutual anticipation, and continuous physical adjustment between partners and with the object. 
Understanding such human-object and human-human coordination dynamics is central to a wide range of applications including human-robot collaboration, humanoid manipulation, character animation, and immersive VR.

While single-person human-object interactions have seen substantial progress in understanding interaction dynamics and synthesizing natural human motions, multi-human object interaction (MHOI) scenarios where humans collaboratively manipulate a shared object remain much less explored despite their importance and potential applicability. 
The initial step towards modeling and understanding MHOI is collecting high-quality and scalable datasets. 
However, MHOI data collection is challenging due to its inherent complexity: first, frequent occlusions occur between participants and with the interacting object, which leads to tracking failures during capture. 
Furthermore, both dynamic body movements and fine-grained finger-level details should be captured simultaneously, as both are critical for understanding the human-human and human-object interactions that coexist in MHOI. 
However, this is a difficult task as body-level movements involve large-scale motions while finger manipulation requires detailed tracking at much smaller scale.

Because of such challenges, existing motion capture methods fall short in capturing and scaling up high-quality MHOI data. 
Traditional marker-based motion capture in multi-view setups is limited to constrained environments, prone to occlusion-induced tracking failures, and requires complex and costly hardware setups.
More lightweight solutions, such as monocular camera-based motion capture, offer greater scalability but demonstrate reduced accuracy and artifacts due to depth ambiguities and single-view constraints.
Consequently, MHOI data captured with existing methods suffers from artifacts such as inconsistent contact, temporal discontinuities, and unnatural motions, and missing detailed finger motion.
Moreover, scaling up data collection remains challenging due to the high cost and complex hardware requirements of multi-view setups.

In this paper, we propose MOCHI (\textbf{MO}tion Enhancement of \textbf{C}ollaborative \textbf{H}uman-object \textbf{I}nteractions), a method to enhance the quality of imperfectly captured MHOI data. 
Specifically, given an input MHOI motion sequence that may contain artifacts such as hand–object misalignment, jittery motion, and missing finger movements, our goal is to generate a visually and physically plausible MHOI sequence with both body and hand motions while preserving the interaction semantics encoded in the original data.
To this end, we develop a two-stage framework.
In the first stage, hand-object interactions are synthesized using an optimization pipeline that generates physically plausible grasps from noisy body input. These grasps are then extended into full hand-object interaction sequences guided by hand-object interaction priors.
In the second stage, the motions of all human subjects are refined by jointly considering the synthesized hand-object interactions and the input body motion sequence, leveraging a diffusion-based single-person motion prior. 
Notably, we propose a formulation that effectively encodes multi-person interaction semantics to enable generating interaction motions with a single-person motion prior.

Our contributions are summarized as follows:
\begin{itemize}
    \item 
    We present the first approach that directly addresses the problem of MHOI enhancement where both human-object and human-human interactions coexist. Our approach enables the generation of physically plausible and natural finger-level and full-body interaction motions from imperfectly captured MHOI data.
    \item We propose a novel two-stage framework that decomposes the complex MHOI refinement problem into hand–object interaction estimation via optimization within a bounded search space and full-body motion refinement that leverages single-person diffusion-based motion priors to model multi-person interaction semantics.
    \item We demonstrate that our enhancement pipeline can serve as a creation tool for MHOI data through varying interacting objects from existing MHOI data and generating from sparse keyframes, while maintaining robustness across diverse scenarios such as generative model outputs and varying number of participants.
\end{itemize}

\section{Related Work}
\label{sec:relwork}

\subsection{Interaction Motion Synthesis}
\paragraph{Full Body Human-Object Interaction}
Synthesizing human-object interactions is a long-standing topic in computer graphics. One line of work studies human motion in static 3D scenes~\cite{jiang2024trumans, hassan2021samp, zhang2022couch, wang2021synthesizing, wang2022towards, hassan2023synthesizing, wang2022humanise, zheng2022gimo}, while another focuses on dynamic interactions where humans manipulate objects. Early methods targeted specific objects~\cite{starke2019neural, merel2020catch, eom2019mpc}, and more recent work generalizes to diverse interactions with unseen objects~\cite{li2023omomo, xu2023interdiff}. Specifically, OMOMO~\cite{li2023omomo} and InterDiff~\cite{xu2023interdiff} use hand-object contact as an intermediate representation in synthesizing plausible object interaction motion.
 Recent approaches also present synthesizing object manipulation within complex environments~\cite{lee2023lama, jiang2024lingo, li2023chois}. 
Another line of research focuses on leveraging deep reinforcement learning (DRL) to physically control characters to interact with the object~\cite{merel2020catch, pan2025tokenhsi, xu2025intermimic, yu2025skillmimicv2, wang2023physhoi, xu2025learningtoball}.
However, these efforts are limited to single-human scenarios, while we aim to support multi-human object interactions (MHOI).
To address this, recent approaches such as TeamHOI~\cite{lionar2026teamhoi} and CooHOI~\cite{gao2024coohoi} began extending physics-based character control to MHOI settings; however, their demonstrations remain limited to a small set of specific objects and tasks.
Meanwhile, Uni-Inter~\cite{liu2025uni} presented a unified model that supports interaction generation of various contexts including human-human, human-object, and human-scene interactions via a unified interaction volume representation.

Recent advances in human-object interaction synthesis have been driven by the development of interaction datasets and data capture systems. 
BEHAVE~\cite{bhatnagar2022behave} and FORCE~\cite{zhang2024force} 
capture full-body interactions using hybrid multi-view and IMU-based setups, but remain limited to controlled environments and struggle with occlusions. CIRCLE~\cite{araujo2023circle} and TRUMANS~\cite{jiang2024trumans} use VR to collect diverse interactions in synthetic scenes, yet only consider interactions of a single participant. HOI-M3~\cite{zhang2024hoim3} captures the behavior of multiple participants in contextual environments but lacks collaborative interactions among them.
CORE4D~\cite{liu2024core4d} addresses this by capturing multi-human collaborative interactions with a hybrid inertial-optical system, although challenges such as scalability, motion artifacts, and occlusion-induced tracking errors still exist. 
Our method complements such efforts by refining noisy or implausible motions and serving as a pipeline that supports data augmentation and creation, potentially alleviating scalability constraints.

\paragraph{Human-Human Interaction}
Human-human interaction synthesis is also an active research area, which ranges from collaborative (e.g., dance, social actions) to competitive scenarios (e.g., sports, fights). 
The diversity and complexity of these behaviors and the scarcity of large-scale multi-person datasets have driven diverse computational approaches to mitigate such challenges. 
Early work used motion patches~\cite{won2014generating, shum2008interactionpatch, lee2006motionpatch, kim2012tiling} or character-wise motion optimization~\cite{shum2007simulating, shum2008simulatingspace, liu2006composition}. Recent methods leverage deep generative models~\cite{shafir2023priormdm, liang2024intergen, xu2024interx, ghosh2024remos, starke2021neural} and physics-based controllers trained via reinforcement learning~\cite{won2021control, zhang2023ig}, using two-person data or interaction-specific rewards to generate realistic behaviors.

\paragraph{Hand-Object Interaction}
Generating hand-object interaction motions has also been extensively studied in computer graphics. Optimization based methods are a traditional approach to synthesize grasps~\cite{pollard2005physically, li2007datagrasp, liu2009dextrous, wang2022dexgraspnet, liu202fc}, which have been extended into full hand-object manipulation sequences~\cite{ye2012synthesishand}. 
More recent methods leverage hand-object interaction datasets~\cite{taheri2020grab} to train deep neural networks to synthesize hand-object manipulation motions, for example from wrist and object trajectory~\cite{zhang2021manipnet, taheri2023grip}, or reconstructing clean hand-object manipulation motions from jittery sequences as~\cite{zhou2022toch, liu2024geneoh}.

Hand-object interaction research has also being extended to cover both body and hand motion while the person is interacting with an object~\cite{ghosh2023imos, taheri2022goal, lu2025choice, wu2025hoifhli, ron2025hoidini}. 
Notably,~\citet{wu2025hoifhli} utilized both optimization-based and generative model-based approaches to enable synthesis for various objects, but only focuses on single-person interaction motion synthesis. Furthermore, their goal differs from ours in that they assume high-level control signals generated from an LLM-based planning module as input, rather than our setting which assumes noisy motion as input. 
Another relevant work is HOIDiNi~\cite{ron2025hoidini}, which proposes a single framework that synthesizes both body and hand motions interacting with the object using diffusion noise optimization. Similar to~\citet{wu2025hoifhli},
HOIDiNi focuses on single-person interactions. 
It also addresses a different problem setting, assuming interaction context provided as text inputs, while our approach takes noisy motion as input.

To provide physical plausibility, recent approaches also focus on physics-based character control with deep reinforcement learning (DRL)~\cite{bae2023pmp, luo2024omnigrasp, tessler2025maskedmanipulator, yu2025skillmimicv2, xu2025learningtoball, wang2023physhoi}.
To train a policy for character control with noisy reference motion as input, PhysHOI~\cite{wang2023physhoi} constructs contact graphs, while 
Skillmimic-V2~\cite{yu2025skillmimicv2} uses trajectory graphs to enable the DRL policy to search for plausible trajectories as training progresses.
While these methods consider finger-level detail and whole-body dynamics, they primarily address domain-specific interactions (e.g., basketball) or single-hand manipulation of small objects (e.g., reaching and picking up an object) with minimal dynamic body movements. Moreover, these methods mainly focuses on cases where a single person interacts with an object.

\subsection{Motion Priors for Motion Tasks}
Recent advances in deep learning have led to extensive research on building motion priors using diverse architectures, including variational autoencoders (VAEs)~\cite{ling2020motionvae}, phase neural networks~\cite{holden2017pfnn}, phase manifolds~\cite{starke2022deepphase}, neural motion fields~\cite{he2022nemf} and diffusion models~\cite{tevet2022mdm}. 
The most relevant work to ours on the usage of the motion priors is refining motion from noisy motion input~\cite{wang2025stablemotion, zhang2024rohm, holden2018robust}. RoHM~\cite{zhang2024rohm} leverages diffusion-based priors built with a motion dataset where noisy and clean motions are paired; 
StableMotion~\cite{wang2025stablemotion} addresses this with a prior that simultaneously learns motion and evaluates motion quality.
While such approaches can refine general common artifacts, those methods are not directly applicable to interaction scenarios as they do not address target-specific artifacts that arise from interaction targets.
Another relevant direction is to enhance motion capture outputs from monocular videos using motion priors. Due to depth ambiguity and temporal inconsistency in per-frame pose estimates, several methods use motion priors to reconstruct natural motions~\cite{zhang2021lemo, rempe2021humor, ye2023slahmr, zhang2024rohm, shi2023phasemp}.
Other approaches apply generative priors to reconstruct plausible motions from sparse signals, such as joint keypoints~\cite{du2023agrol, xie2024omnicontrol, starke2024categorical} or keyframes~\cite{cohan2024flexible}. For navigation tasks like maze traversal and path following, reinforcement learning combined with VAE-defined priors has been used to produce optimal actions~\cite{ling2020motionvae}. 
%
Beyond specific tasks, recent work leverages diffusion models as flexible priors, either through guided generation during sampling~\cite{karunratanakul2023gmd} or optimization in noise space~\cite{karunratanakul2024dno}, while GENMO~\cite{li2025genmo} designs diffusion models to operate with diverse conditioning signals for flexible control through various input modalities.

\begin{figure*}
\centering
\includegraphics[width=1.0\linewidth, trim={0 0.3cm 0 0}]{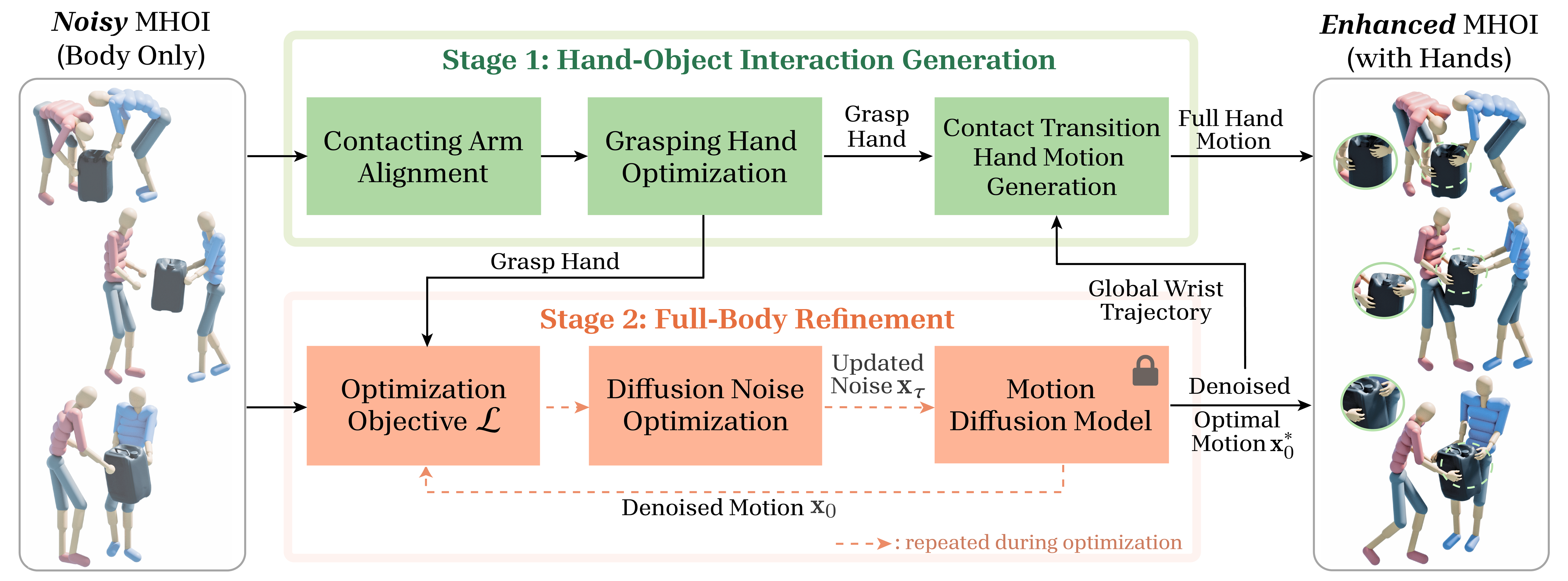}
    \caption{System Overview.}
    \label{fig:sys_overview}
\end{figure*}

\section{Method}
\label{sec:method}

We aim to develop a general method to enhance the quality of collaborative multi human object interaction (MHOI) motion sequences, applicable to any data including noisy capture results or erroneous outputs.
We have examined existing MHOI data and identified three common shortcomings.
First, in human object manipulation scenarios (e.g., object rearrangement) which form a dominant category of human object interactions, detailed hand motions are often missing due to the inherent complexity of finger movements and the technical challenges of hand motion capture.
Second, full-body motions frequently exhibit unnatural artifacts, primarily due to difficulties in capturing MHOI scenarios such as occlusions.
Third, the complexity and diversity of real world MHOI scenarios far exceed what existing MHOI datasets can represent, resulting in limited variability.

To address these challenges, we decompose the MHOI data enhancement task into two stages (see Figure~\ref{fig:sys_overview}). First, we synthesize plausible grasping hand motions for hand-object interactions. Second, we enhance the body motion quality by incorporating the generated hand grasping and contact information from the first stage while preserving the original interaction semantics.
An additional advantage of this decomposition is that it allows us to readily leverage existing hand-object interaction priors and single person motion priors, for which high quality data is relatively abundant and data collection is more feasible than for MHOI-specific cases.

\subsection{Data Representation and Preprocessing}
An MHOI motion sequence can be represented as $\mathcal{I} = \{ \mathbf{M}^i, \mathbf{M}^j, \mathbf{O}, \mathbf{G}_{\text{obj}} \}$. Here, $\mathbf{M} = \{ \mathbf{m}_t \}_{t=0}^{T}$ denotes the motion sequence of a person, where $\mathbf{m}_t$ represents the full-body pose at time $t$, expressed either as 3D joint positions or as a combination of global translation and rotation of the root joint and local rotations of the other joints, 
while $T$ is the length of the sequence. 
$\mathbf{O} = \{ \mathbf{o}_t \}_{t=0}^{T}$ indicates the object pose trajectory, where $\mathbf{o}_t \in SE(3)$ describes the object pose at time $t$. $\mathbf{G}_{\text{obj}}$ denotes for the object geometry, represented by a mesh with a set of vertices and faces.
Note that we assume the object geometry $\mathbf{G}_{\text{obj}}$ is unchanged throughout the process.
Our method takes as input a motion sequence $\mathcal{I}$ that may contain artifacts and produces a \textit{enhanced} motion sequence $\hat{\mathcal{I}} = \{ \hat{\mathbf{M}}^i, \hat{\mathbf{M}}^j, \hat{\mathbf{O}}, \mathbf{G}_{\text{obj}}, \mathbf{H}^i, \mathbf{H}^j \}$. Here, $\hat{\mathbf{M}} = \{ \hat{\mathbf{m}}_t \}_{t=0}^{T}$ denotes the refined body motion for each subject, and $\hat{\mathbf{O}} = \{ \hat{\mathbf{o}}_t \}_{t=0}^{T}$ denotes the refined object trajectory. 
Notably, the enhanced sequence $\hat{\mathcal{I}}$ include synthesized finger motions interacting with the object, which are represented as $\mathbf{H} = \{ \mathbf{h}^L_t, \mathbf{h}^R_t \}_{t=0}^{T}$. 
We adopt the hand geometries and pose parameters from SMPL-X~\cite{pavlakos2019smplx} and reparameterize the hand pose parameters so that new parameters $\mathbf{h}_t \in \mathbb{R}^{51}$ consist of the wrist orientation and translation expressed in the object space (i.e., the object-centric coordinate system), along with local joint rotations of all finger joints represented in angle-axis form.
For brevity, we describe the method assuming two human subjects, but our method can be straightforwardly extended to scenarios involving more than two subjects.

As a preprocessing step before the motion enhancement process, we first remove artifacts from the input object trajectory, where common cases include jittery object movement and instances where the object abruptly stops and then moves sharply again.
We compute temporal accelerations between consecutive frames and identify noisy object poses based on abrupt changes in acceleration, where the missing object poses are filled via interpolation.

\subsection{Stage 1: Hand-Object Interaction Generation}
\label{sec:contact_refine}

When the hands are in contact with an object, the grasping hand poses and their spatial locations relative to the object tend to remain consistent.
Based on this observation, we divide the input MHOI sequence into three contact-related phases: a pre-contact phase, during which the hands approach the object; a contact phase, during which they maintain consistent contact with the object; and a post-contact phase, during which the hand-object interaction concludes and the contacts are released.
Accordingly, we formulate hand-object interaction generation as a two-step process: we first generate stable grasping hand poses during the contact phase, and then generate hand motions for contact transitions in the pre-contact and post-contact phases. 
This generation process is repeatedly applied to all hands that are involved in the interaction during the given MHOI sequence.

\subsubsection{Identifying Contact Phase.}
To identify the contact phase, we first generate contact labels by empirically setting a distance threshold of 8cm, where a contact label is assigned as true if the minimum distance between the input wrist pose and the object falls below this threshold. 
Since hand poses in the input data are noisy, we employ sliding window convolution with a window size of six frames (at 30 FPS) to identify consecutive contact frames.
We then refine contact labels with the rule that when an object is in motion, at least one person must be in contact with it. We analyze object acceleration to identify frames in which the object is moving and compare these with the detected contact labels to determine the final contact phases depending on the scenario.
For cases in which participants jointly manipulate a shared object, such as object rearrangement or rotation, the contact phases of two participants should be consistent with the frames during which the object is moving.
For handover cases, object manipulation follows a sequential process in which the object is transferred from one person to another. In such cases, the contact phases of the two participants should have a short temporal overlap, and their union should align with the frames during which the object is moving.

\begin{figure}
\centering
\includegraphics[width=1.0\linewidth, trim={0 0.5cm 0 0.3cm}]{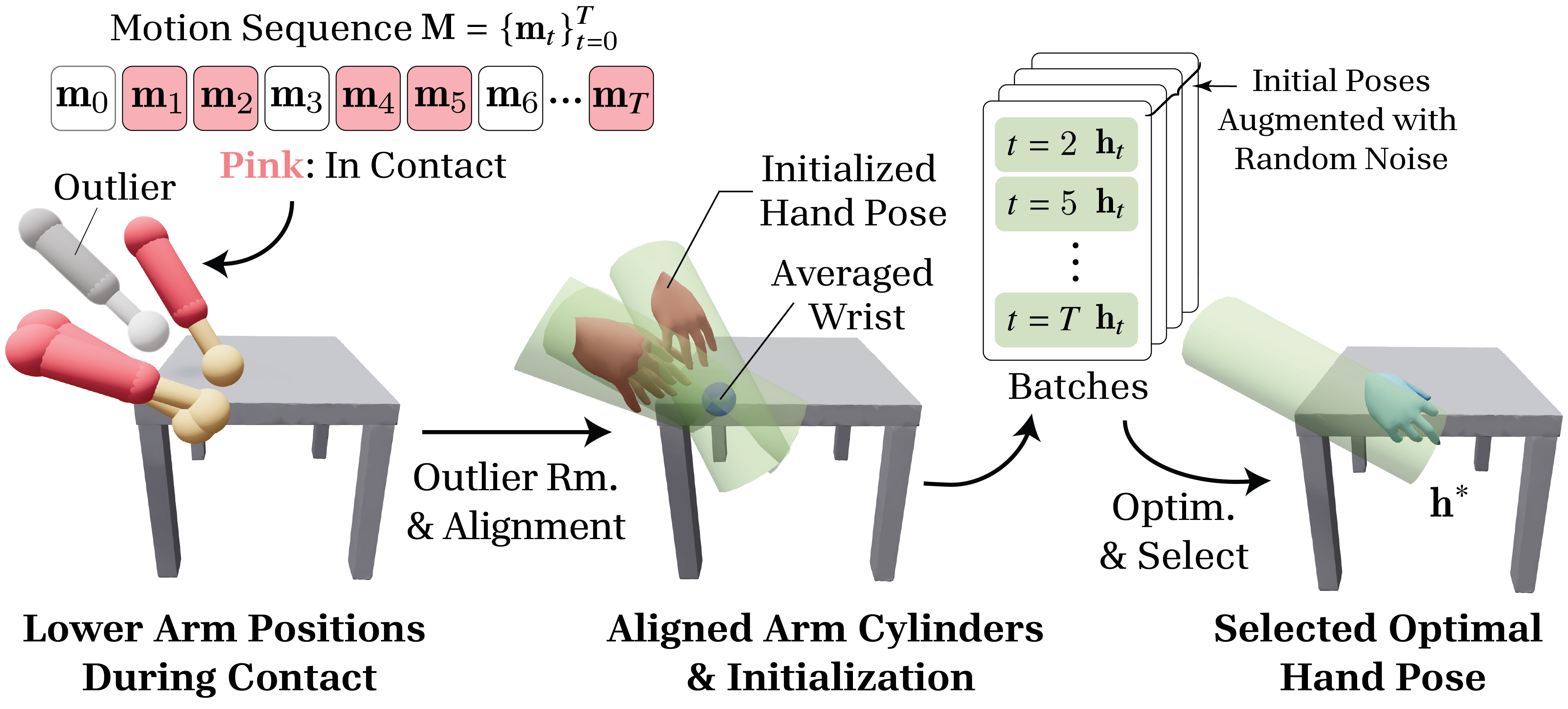}
    \caption{An overview of grasping hand pose generation. (Left) Lower arms in contact, where the pink arms are inliers and the gray arms are outliers. 
    (Middle) The inlier lower arms are aligned around the averaged wrist position. 
    A cylindrical bounding volume is constructed for each aligned inlier lower arm, with its orientation aligned to the lower arm direction and its lower base centered at the average wrist location.
    The initial solutions for the grasp optimization are set with the wrist position at the cylinder's upper base and the wrist orientation aligned with the cylinder axis.
    (Right) Multiple candidate hand poses are generated from different initial hand poses with added random noise, from which the optimal hand pose is selected.}
    \label{fig:batch_optim}
\end{figure}

\subsubsection{Grasping Hand Pose Generation}
The goal of this stage is to find an optimal grasping hand pose $\mathbf{h}^*$ during the contact phase that is most plausible given the object geometry and the noisy body motion within the phase. Although the ground truth hand pose in object space remains consistent during contact, the noisy input motion may cause the input hand poses to be inconsistent during contact. 
As a result, performing the optimization based on a single arbitrary frame within the contact phase may lead to suboptimal solutions.

To address this issue, for each hand we perform a batched optimization by constructing a batch from all frames whose contact label is \textit{true} within the contact phase. 
To make the optimized grasp be consistent with the body motion,
we construct the batch of initial solutions for optimization using the lower arm positions during contact, where those initial solutions are further augmented by applying random perturbations to the wrist orientation to improve exploration of the solution space. 
We then optimize across the entire batch and evaluate the quality of the resulting candidates to select the optimal hand pose. 
Figure~\ref{fig:batch_optim} illustrates this process, where the details of batch construction and optimization are described in paragraph~\ref{sec:batch_optim}.


The optimization process follows the formulation below:
\begin{equation} \label{eq:opt_hand_pose}
\mathbf{h}^* = \arg\min_{\mathbf{h}} \lambda_\text{fc}\mathcal{L}_\text{fc} + \lambda_\text{obj}\mathcal{L}_\text{obj} + 
\lambda_\text{hp}\mathcal{L}_\text{hp} + 
\lambda_\text{reg}\mathcal{L}_\text{reg},
\end{equation}
which consists of four loss terms, with the first three inspired by the formulation in~\cite{wang2022dexgraspnet, liu202fc}, while the fourth is a novel term specifically designed for noisy MHOI. $\lambda$s are weights to balance the loss terms.

The first term, $\mathcal{L}_{\text{fc}}$, is a differentiable force closure loss that encourages the generation of stable grasps on the object, defined as
\begin{equation}
\mathcal{L}_{\text{fc}} = \lVert \mathcal{G} \cdot \mathbf{N}{\text{c}} \rVert,
\end{equation}
where $\mathcal{G} \in \mathbb{R}^{6 \times 3N}$ denotes the grasp matrix that maps contact forces to the net wrench on the object,
and $\mathbf{N}_{\text{c}} \in \mathbb{R}^{N \times 3}$ is the concatenation of $N$ contact normals $\{\mathbf{n}_{\text{c}}\}$ corresponding to $N$ contact points $\{\mathbf{p}_{\text{c}}\}$. 
At each optimization iteration, the contact points are obtained by sampling $N$ vertices from candidate vertices on the hand surface using the Metropolis–Hastings algorithm. The sampling process is initially near uniform, producing random samples in the early iterations, and gradually converges to stable contact points as the optimization progresses.

The object interaction loss $\mathcal{L}_{\text{obj}}$ consists of a distance loss $\mathcal{L}_{\text{dist}}$ and a hand–object penetration loss $\mathcal{L}_{\text{pen}}$, which together encourage the hand to remain close to the object while preventing interpenetration.
\begin{equation}
    \mathcal{L}_\text{dist} = \sum_{\mathbf{p} \in \{\mathbf{p}_{\text{c}}\}} d(\mathbf{p},\mathbf{G}_{\text{obj}}), \
    \mathcal{L}_\text{pen} = \sum_{\mathbf{v} \in S(\mathbf{G}_{\text{obj}})} [\mathbf{v}\in\mathbf{G}_{\text{hand}}]d(\mathbf{v}, \mathbf{G}_{\text{hand}})
\end{equation}
$d(\cdot,\cdot)$ is the point-to-mesh distance measured using a signed distance function (SDF), $S(\cdot)$ is the surface point cloud of the input mesh, and $[\mathbf{v} \in \mathbf{G}_{\text{hand}}] = 1$ if the point $\mathbf{v}$ lies inside $\mathbf{G}_{\text{hand}}$, and $0$ otherwise.

The hand plausibility loss $\mathcal{L}_{\text{hp}}$ encourages plausible hand poses and natural hand shapes. It consists of a self penetration loss $\mathcal{L}_{\text{spen}}$, which penalizes interpenetration among vertices of the hand geometry, and a pose prior loss $\mathcal{L}_{\text{prior}}$, which encourages the hand pose parameters to be within a pretrained hand pose distribution modeled as a multivariate Gaussian.
Each term is formulated as:
\begin{equation}
    \begin{split}
        \mathcal{L}_{\text{spen}} &= \sum_{\mathbf{v}_i, \mathbf{v}_j \in S(\mathbf{G}_{\text{hand}})} [\mathbf{v}_i \neq \mathbf{v}_l] \max\big(\delta - \|\mathbf{v}_i - \mathbf{v}_j\|^2, 0\big) \\
        \mathcal{L}_{\text{prior}} &= \left\| \text{diag}(\boldsymbol{\sigma})^{-1} \left( \mathbf{h} - \boldsymbol{\mu} \right) \right\|^2
    \end{split}
\end{equation}
where $\delta$ is the minimum distance threshold, $\boldsymbol{\mu} \in \mathbb{R}^d$ and $\boldsymbol{\sigma} \in \mathbb{R}^d$ are the mean and standard deviation of the pretrained hand pose distribution, respectively.

\begin{figure}
\centering
\includegraphics[width=1.0\linewidth, trim={0 0.5cm 0 0.3cm}]{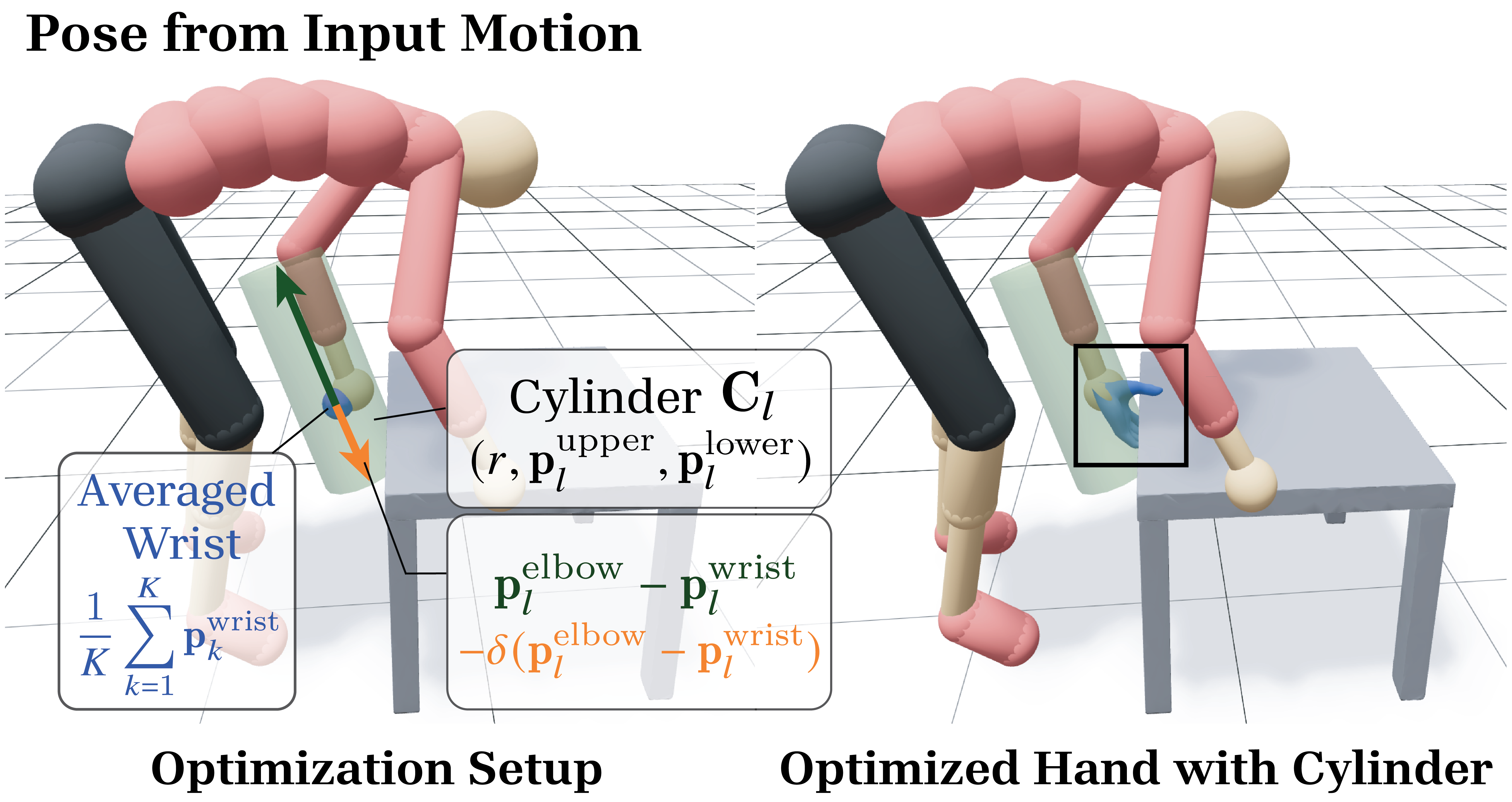}
    \caption{An example of a cylindrical bound for grabbing a table. (Left) The cylindrical bound is constructed near the lower arm. (Right) The optimized hand grasp pose is constrained to remain within the cylindrical bound.}
    \label{fig:hand_cylinder}
\end{figure}

\paragraph{Regularization.}
Since the three terms above consider only plausibility in hand-object interaction, optimizing with these terms alone may result in hand poses that are inconsistent with the input body motion. 
To address this, we construct a bounded constraint on the solution space to preserve the semantics encoded in the input body motion while generating natural grasps. 
Since the input body motion's positional information is noisy despite containing important semantic cues, we establish a bounded search space rather than directly relying on exact positional values.
Specifically, from the input motion we derive feasible wrist and arm information during the contact phase to construct a cylindrical bound $\mathbf{C} = (r, \mathbf{p}^\text{upper}, \mathbf{p}^\text{lower})$, where $r$ is the radius, and $\mathbf{p}^\text{upper}$ and $\mathbf{p}^\text{lower}$ are the centers of the upper and lower bases defined by the elbow and wrist positions in object space (see Figure~\ref{fig:hand_cylinder} (left)). 
From a batch of frames where the arm is identified to be in contact with the object, we compute the average wrist position across all $K$ frames after removing outliers using Mahalanobis distance with an 80\% confidence threshold.
Each cylinder per frame in the batch is then aligned such that its wrist position $\mathbf{p}^\text{wrist}$ matches the averaged wrist position. 
This process is visually illustrated in Figure~\ref{fig:batch_optim} (left and middle). 

The cylinder parameters $\mathbf{C}_l = (r, \mathbf{p}^\text{upper}_l, \mathbf{p}^\text{lower}_l)$ for frame $l$ can be formally expressed as follows:
\begin{equation}
\begin{aligned}
    \mathbf{p}^\text{upper}_l &= \Big( \frac{1}{K} \sum_{k=1}^{K} \mathbf{p}^\text{wrist}_k \Big) + (\mathbf{p}^\text{elbow}_l - \mathbf{p}^\text{wrist}_l), \\ 
    \mathbf{p}^\text{lower}_l &= \Big( \frac{1}{K} \sum_{k=1}^{K} \mathbf{p}^\text{wrist}_k \Big) - \delta (\mathbf{p}^\text{elbow}_l - \mathbf{p}^\text{wrist}_l),
\end{aligned}
\end{equation}
$\mathbf{p}^{\text{wrist}}_k$ denotes the wrist position at the $k$-th contact frame, $\mathbf{p}^{\text{wrist}}_l$ and $\mathbf{p}^{\text{elbow}}_l$ 
denote the wrist and elbow positions at the current frame $l$. 
The radius $r$ and elongation factor $\delta$ are chosen to ensure the hand geometry remains within the cylinder.
Using this cylinder constraint, we define the regularization loss $\mathcal{L}_{\text{reg}}$ as
\begin{equation}
    \mathcal{L}_\text{reg} = \sum_{\mathbf{v} \in S(\mathbf{G}_{\text{hand}})} [\mathbf{v} \notin \mathbf{C}] d(\mathbf{v}, \mathbf{C})
\label{eq:cylinder_regloss}
\end{equation}
which enforces the hand geometry derived by the optimized hand pose parameters to remain within the cylinder. 
In the later optimization steps, only the lower part of the cylinder (near and below the wrist) is used for calculation to constrain the hand to this region.
Fig.~\ref{fig:hand_cylinder} (right) illustrates an example where the optimized grasping hand remains within the cylinder constraint after the optimization process.

\paragraph{Initialization and Optimization.}
\label{sec:batch_optim}
Hand pose optimization is performed in parallel for all $K$ contact frames in the batch, where the initial hand pose is set with the wrist position at the center of the corresponding cylinder's upper base $\mathbf{p}^\text{upper}$ 
and the wrist orientation is initialized by aligning the wrist orientation from the input body pose to the cylinder axis.
Each initialized batch is augmented $B$ times by perturbing the input wrist orientations with random noise in the range of 15 degrees. 
During optimization, the object 6DoF pose is perturbed with annealing noise (maximum rotation up to 11.5 degrees (0.2 radians) and translation up to 10cm) in the early 20\% of iterations to avoid local minima.
After all optimization processes are done, among all candidate solutions in the augmented batch we select the optimal hand pose $\mathbf{h}^*$ with the lowest loss value in Eq.~\ref{eq:opt_hand_pose}.
 The initialization and optimization process are visualized in Figure~\ref{fig:batch_optim} (middle and right). 

\subsubsection{Hand Motion Generation for Contact Transitions}
\label{sec:hand_motion_seq}

In this step, we generate hand motions for the pre-contact and post-contact phases to create natural transitions between rest and grasp poses, thereby forming a complete hand-object interaction sequence. 
We utilize a conditional diffusion model trained on hand–object interaction datasets to learn interaction priors as in~\cite{wu2025hoifhli}.
Technically, the model takes a hand–object proximity feature $\Pi \in \mathbb{R}^{T \times 100}$ and a target grasping pose $\mathbf{h}^{\text{grasp}}$ as input, and generates a full finger motion sequence $\{\mathbf{h}_t\}_{t=0}^{T}$, where $T$ denotes the sequence length ($T=30$ in our experiments, at 30 FPS).
We use the optimized grasping pose as the target grasping pose, while the proximity feature $\Pi$ is computed from a wrist 6DoF (position and orientation) trajectory, obtained from the refined full-body motion (Sec.~\ref{sec:motion_refine}). Specifically, to compute the hand-object proximity feature $\Pi$ we sample 100 palm side vertices from the hand mesh, set the wrist position and orientation to follow the input wrist 6DoF trajectory with the fingers in a rest pose, and compute the minimum distances from these vertices to the object surface.


\subsection{Stage 2: Full Body Motion Refinement}
\label{sec:motion_refine}
The goal of this stage is to refine the input full-body motions $\{ \mathbf{M}^i, \mathbf{M}^j \}$ such that the refined motions $\{ \hat{\mathbf{M}}^i, \hat{\mathbf{M}}^j \}$ are consistent with the grasping hand poses generated in the previous stage, while preserving the original interaction semantics and removing underlying artifacts.
We formulate this as a conditional generation process guided by a diffusion-based motion prior.
However, learning MHOI-specific priors using diffusion models is not suitable for our setting due to the limited quality and scalability of available training data for MHOI scenarios.
To address these limitations, we propose a method that leverages a motion prior trained solely on single-person motion data, which is relatively abundant, diverse, and of high quality.

\subsubsection{Diffusion Model Basics.}
Diffusion models~\cite{ho2020denoising} consist of a forward process and a reverse process. 
In the forward process, Gaussian noise is incrementally added to clean data $\mathbf{x}_0$. This process is formulated as a Markov chain, where noise is added over $N$ steps according to a fixed variance schedule $\beta_\tau$:
\begin{equation}
q(\mathbf{x}_\tau \mid \mathbf{x}_{\tau-1}) = \mathcal{N}(\mathbf{x}_\tau; \sqrt{1 - \beta_\tau} \, \mathbf{x}_{\tau-1}, \beta_\tau \mathbf{I}),
\end{equation}
where $\tau$ denotes the diffusion step.
The reverse process aims to reconstruct the original data $\mathbf{x}_0$ from noise $\mathbf{x}_N \sim \mathcal{N}(0, \mathbf{I})$
with a neural network trained to approximate the distribution $p_\theta(\mathbf{x}_{\tau-1} \mid \mathbf{x}_\tau, \mathbf{c})$, optionally conditioned on auxiliary information $\mathbf{c}$ (e.g., text or action labels). This denoising process is modeled as:
\begin{equation}
p_\theta(\mathbf{x}_{\tau-1} \mid \mathbf{x}_\tau, \mathbf{c}) = \mathcal{N}(\mathbf{x}_{\tau-1}; \mu_\theta(\mathbf{x}_\tau, \tau, \mathbf{c}), (1 - \alpha_\tau)\mathbf{I}),
\end{equation}
where $\mu_\theta(\mathbf{x}_\tau, \tau, \mathbf{c})$ denotes the predicted mean from the trained neural network $m_\theta$ with parameters $\theta$.
In practice, the neural network predicts the clean motion $\mathbf{x}_0$ (denoted $m_\theta(\mathbf{x}_\tau, \tau, \mathbf{c})$), and the mean is computed by:
\begin{equation}
\mu_\theta(\mathbf{x}_\tau, \tau, \mathbf{c}) = \frac{\sqrt{\bar{\alpha}_{\tau-1}} \beta_\tau}{1 - \bar{\alpha}_\tau} m_\theta(\mathbf{x}_\tau, \tau, \mathbf{c}) + \frac{\sqrt{\alpha_\tau} (1 - \bar{\alpha}_{\tau-1})}{1 - \bar{\alpha}_\tau} \mathbf{x}_\tau,
\end{equation}
with $\bar{\alpha}_\tau = \prod_{i=1}^\tau \alpha_i$ and $\beta_\tau = 1 - \alpha_\tau$. The model is trained by minimizing a reconstruction loss computed between the predicted and ground-truth output.

\subsubsection{Diffusion Noise Optimization.}
\label{sec:diffusion_opt}
We adopt a noise optimization approach~\cite{karunratanakul2024dno}, which explicitly optimizes the latent noise $\mathbf{x}_\tau$ such that its subsequent denoising trajectory, defined by a pretrained diffusion model, yields data that satisfies the given criteria. Specifically, the optimization is performed iteratively: first by denoising the noise $\mathbf{x}_\tau$ to obtain $\mathbf{x}_0$, then computing the loss using an objective function $\mathcal{L}$ based on $\mathbf{x}_0$, and finally backpropagating gradients through the entire denoising trajectory to update $\mathbf{x}_\tau$. 
During each optimization step, we employ DDIM~\cite{song2020ddim} for the reverse process with gradient normalization for stable optimization.
Additionally, we introduce small random perturbations to the updated $\mathbf{x}_\tau$ to encourage exploration.

In our pipeline, we employ a motion diffusion model~\cite{tevet2022mdm} trained on a single-person motion dataset. 
Consequently, the data $\mathbf{x}$ consists of a pose sequence of a single-person, where each pose is represented as a 263-dimensional tuple according to the format used in HumanML3D~\cite{bensabath2024humanml3d}. 
To generate the motions of all human subjects in a given MHOI scenario, we perform optimization incrementally: we fully complete all optimization steps for one subject before proceeding to the next.

\paragraph{Optimization Objective.}
For each person $i$, our goal is to find the optimal noise $\mathbf{x}_\tau^*$ such that denoising it through the pretrained diffusion model $\mathcal{D}$ yields the desired refined motion $\hat{\mathbf{M}}^i$. This can be formulated by
\begin{equation}
\mathbf{x}_\tau^* = \arg\min_{\mathbf{x}_\tau} \mathcal{L}(\text{ReverseProcess}(\mathcal{D}, \mathbf{x}_\tau)),
\end{equation}
where $\text{ReverseProcess}(\mathcal{D}, \mathbf{x}_\tau)$ denotes the denoising process from noise $\mathbf{x}_\tau$ to clean motion $\mathbf{x}_0$, following the reverse trajectory obtained by $\mathcal{D}$.
The objective is computed over the entire sequence $L=\sum_{t=0}^T l_t$, where $T$ denotes the sequence length of $\mathbf{x}_0$. The per-frame objective $l_t$ is computed as follows:
\begin{equation}
    l_t = \lambda_\text{pose}\mathcal{L}_\text{pose}
    + \lambda_\text{contact}\mathcal{L}_\text{contact}
    + \lambda_\text{IG}\mathcal{L}_\text{IG}
    + \lambda_\text{col}\mathcal{L}_\text{col}
    + \lambda_\text{em}\mathcal{L}_\text{em}
\end{equation}
where each term is weighted by $\lambda$, and we describe each objective in detail below.

\paragraph{Pose Objective} 
The pose objective $\mathcal{L}_{\text{pose}}$ penalizes excessive deviations from the input motion, which is designed to remove underlying artifacts existing in the input motion while preserving its original semantics.

\begin{equation}
    \begin{split}
         \mathcal{L}_\text{pose} &= 
         \lambda_\text{body} \mathcal{L}_\text{body} + 
         \lambda_\text{foot} \mathcal{L}_\text{foot} 
         +
         \lambda_\text{fs} \mathcal{L}_\text{fs}
         \\
         \mathcal{L}_\text{body} &= \sum_{j\in \mathcal{J}} \left\| \hat{\mathbf{q}}_j - (\mathbf{q}_j + \sigma) \right\|^2 \\
         \mathcal{L}_\text{foot} &= \sum_{f \in \mathcal{F}} \max(0, -\hat{\mathbf{q}}_\text{f,y}) +  \max(0, \min_{f \in \mathcal{F}} \hat{\mathbf{q}}_\text{f,y} - \delta) \\
\mathcal{L}_\text{fs} &= \sum_{f \in \mathcal{F}} \|\mathbf{v}_{f,t}\|, \quad \text{if }\|\mathbf{v}_{f,t}\| < \tau
    \end{split}
\label{eq:pose_loss}
\end{equation}
where $\hat{\mathbf{q}}_j$ and $\mathbf{q}_j$ are $j$-th joint position derived from the input motion and the reverse process, respectively, $\mathcal{J}$ is the set of joint indices. 
To avoid overfitting to potentially erroneous input poses, we introduce a small Gaussian noise $\sigma$ to the input joint positions. 
Moreover, to ensure plausible foot contact, we include $\mathcal{L}_\text{foot}$ that penalizes cases where foot joints penetrate the ground or both feet hover above a specified height threshold $\delta$.
Here, $\hat{\mathbf{q}}_\text{f,y}$ denotes the predicted vertical (y-axis) position of each foot joint, and $\mathcal{F}$ represents the set of foot joint indices. To further mitigate foot slip artifacts, we introduce a foot slip loss $\mathcal{L}_{\text{fs}}$, which penalizes foot velocity when the foot is in contact with the ground. Contact is determined when the foot velocity falls below a predefined threshold $\tau$. This term is included only in the later iterations during optimization, as contact is often unreliable in the early iterations when the motion may still contain artifacts.

\paragraph{Contact Objective}
This loss ensures the body motion maintains consistent and stable contact with the object. Specifically, during the contact phase, the generated wrist positions is enforced to align with the target wrist positions derived from the grasp hand pose in the previous stage.
\begin{equation}
\label{eq:body_refine_contact}
    \mathcal{L}_\text{contact} = \sum_{k \in C} \left\| \hat{\mathbf{q}}_k - \hat{\mathbf{o}}_t \cdot \mathbf{p}^\text{wrist}_k
    \right\|^2
\end{equation}
where $C$ denotes the indices of hand joints that are in contact with the object. Note that each target wrist position $\mathbf{p}^{\text{wrist}}_k$ is transformed into the global space using the refined object pose $\hat{\mathbf{o}}_t$ at the current frame.

When the input motion is noisy, conflicts can arise between optimization objectives, such as maintaining contact with the object while the input body is far from the object. 
In such cases, the pose objective ($\mathcal{L}_\text{pose}$ in Eq.~\ref{eq:body_refine_contact}) is implemented to automatically detect such cases and switch to root-relative measurement in those frames, allowing contact-related objectives to take priority.

\begin{figure}
\centering
\includegraphics[width=1.0\linewidth]{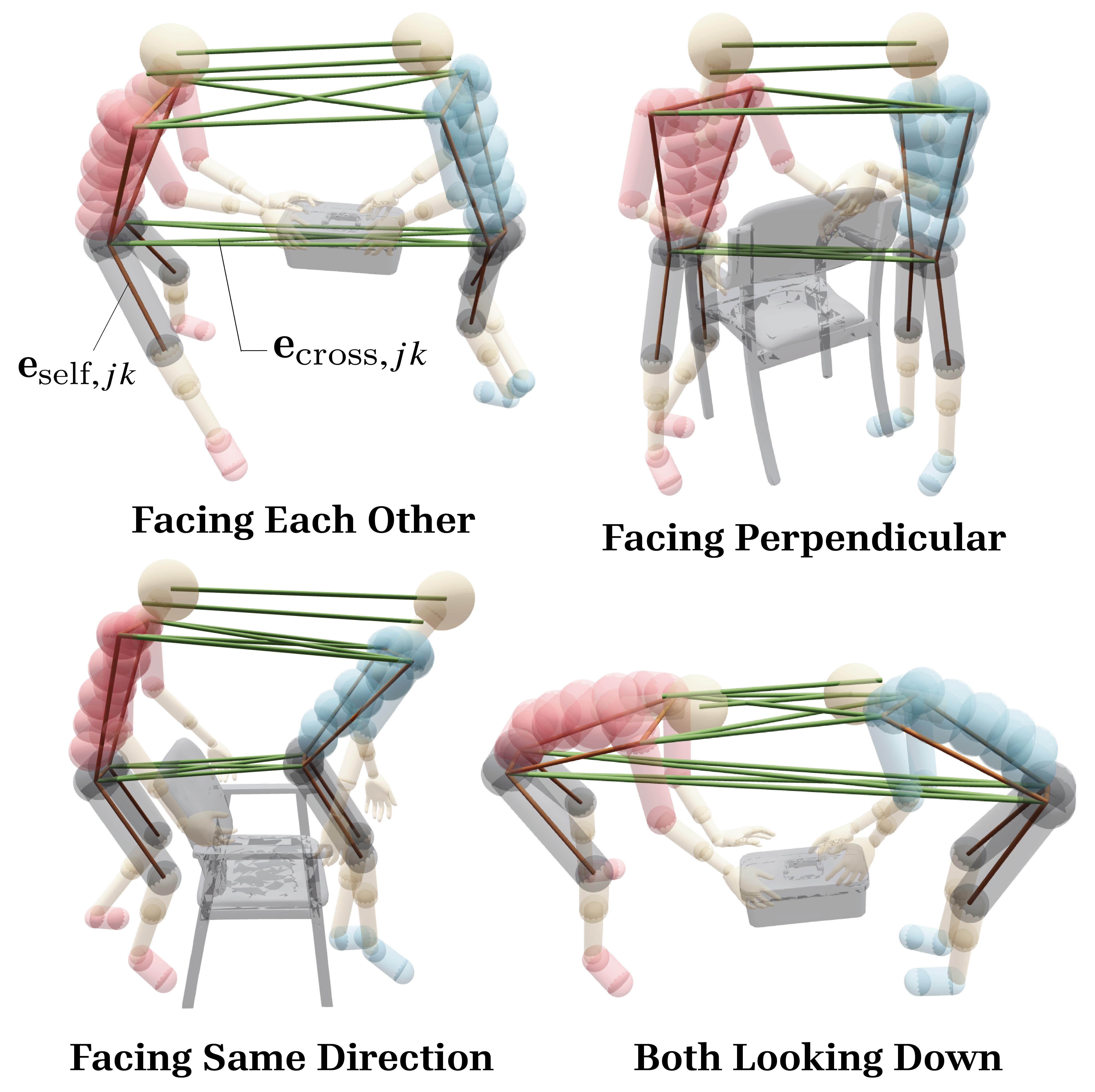}
    \caption{Visualization of the interaction graph showing self-edges $\mathbf{e}_{\text{self},jk}$ (red) and cross-edges $\mathbf{e}_{\text{cross},jk}$ (green) across four common human-human spatial relationships in MHOI scenarios.
    }
    \label{fig:ig_vis}
\end{figure}

\paragraph{Interaction-Graph Objective}
In collaborative MHOI scenarios, humans exhibit coordinated movements around a shared object.
During collaboration, participants must not only maintain spatial proximity but also coordinate their orientations and movement dynamics based on their partner's movement. 
For instance, when two people face each other while moving an object together, they must maintain consistent velocities, while their facing directions must remain aligned toward each other to ensure proper collaboration.
To capture and quantify such relationships between the participants, we construct an interaction graph $\Gamma$ inspired by~\cite{zhang2023ig, ho2010spatial} to encode the spatial configurations between human subjects.

Specifically, in the interaction graph $\Gamma$, nodes are defined as the set of joint positions $\mathbf{q}_j$ and edges $\mathbf{e}_{jk} = (\mathbf{q}_k - \mathbf{q}_j) \in \mathbb{R}^3$ are defined by the set of vectors between two joint positions $\mathbf{q}_j$ and $\mathbf{q}_k$. Here, the graph contains self-edges $\mathbf{e}_{\text{self},jk}$ where both joints $j$ and $k$ belong to the same human subject, and cross-edges $\mathbf{e}_{\text{cross},jk}$ where joint $j$ belongs to one subject while joint $k$ belongs to the other subject. 
The self-edges connect shoulder, hip, and knee joints due to the high noise in end effectors, while cross-edges connect shoulder and hip joints where spatial relationships such as facing direction are most prominently encoded. 
Additionally, we include connections between head and neck joints for cross-edges to preserve mutual gaze and attention patterns commonly seen in collaborative scenarios.

Using the interaction graph, we design the objective $\mathcal{L}_{\text{IG}}$ to preserve directional and velocity alignment of edges as follows:
\begin{equation}
\begin{split}
\mathcal{L}_{\text{IG}} &=
    \lambda_{\text{dir}}
    \frac{1}{|\mathcal{E}|} 
    \sum_{\mathbf{e} \in \mathcal{E}}
    \cos^{-1}\left(
    \frac{ \hat{\mathbf{e}} }
    {\| \hat{\mathbf{e}} \|}
    \cdot
    \frac{ \mathbf{e} }
    {\| \mathbf{e} \| }
    \right) \\
& +
    \lambda_{\text{vel}}
    \frac{1}{|\mathcal{E}|}
    \sum_{\mathbf{e} \in \mathcal{E}}
    \left\|
    (\hat{\mathbf{e}}_{t+1} - \hat{\mathbf{e}}_t)
    -
    (\mathbf{e}_{t+1} - \mathbf{e}_t)
    \right\|_2^2
\end{split}    
\end{equation}
where $\mathcal{E}$ is the set of edges of the graph $\Gamma$, and $\hat{\mathbf{e}}$ is derived from the reverse process with $\mathbf{x}_0$ while $\mathbf{e}$ is computed from the input motions.
The first term enforces directional alignment to preserve the relative orientations between interacting individuals, while the second term enforces velocity alignment to maintain the temporal dynamics of their spatial relationships. 
In contrast to the setting considered in~\cite{zhang2023ig, ho2010spatial}, we assume that our input motions may contain substantial noise. As positional error is more sensitive to such noise, we exclude it from our interaction graph objective. 
This design choice is also reasonable as positional alignment is addressed by other objectives (Eq.~\ref{eq:pose_loss}) in the optimization framework.

\paragraph{Collision Objective}
We introduce a collision objective $\mathcal{L}_{\text{col}}$ to prevent collisions between characters. 
In many collaborative MHOI scenarios, we observed that collisions primarily occur in the head and shoulder regions, for example when two people bend down together to pick up a small box. To address this, we set up collision geometries to detect potential inter-person collisions. Specifically, we place a spherical bounding volume with a 5cm radius at each shoulder and collar joint, and a larger sphere with a 10cm radius at the head joint.
The collision objective is then measured using the signed distance function (SDF) between these spherical bounding volumes. Note that this objective is inactive during the optimization of the first human subject.
Additionally, we observed that in several cases, such as two people carrying a large box together, collisions may occur between humans and objects. 
In such cases, the collision objective is optionally extended to human-object collisions by placing spherical bounding volumes on the head (radius 10cm), knee (radius 6cm) and shoulder (radius 4cm) joints. Similar to inter-person collisions, the collisions between humans and objects are detected using SDF values between the object vertices and the spherical bounding volumes. 
The collision objective is applied in the latter steps of optimization, as we empirically found this leads to more stable convergence.

\paragraph{Energy Minimization Objective.}
We empirically observe that the optimized results can exhibit motions with excessively large movements as the optimization aggressively satisfies the imposed constraints. To mitigate this, we introduce an energy minimization objective, which is formulated as:
\begin{equation}
\mathcal{L}_{\text{em}} = \sum_{j\in\mathcal{J}} w_j \cdot \|\mathbf{a}_{j}\|^2
\end{equation}
where $\mathbf{a}_{j}$ denotes the acceleration of joint $j$ at the current frame, computed via finite differences. The weight $w_j$ assigns higher importance to joints closer to the pelvis (root joint) and decreases as the joint approaches the end effectors.

\section{Results}
\label{sec:results}

In this section, we present both qualitative and quantitative results of our method across various types of noisy MHOI data and diverse application scenarios to demonstrate its applicability. 
To evaluate both Stage 1 (hand-object interaction generation, Sec.~\ref{sec:contact_refine}) and Stage 2 (body motion refinement, Sec.~\ref{sec:motion_refine}), we focus on collaborative object manipulation scenarios where multiple human subjects jointly interact with a shared object. 
We further demonstrate that our pipeline can be extended to a wider range of MHOI scenarios such as increasing the number of participants or including static human-object interactions such as sitting on a chair (Sec.~\ref{sec:result_extend}). 
Additionally, we perform ablation studies to assess the contribution of each technical component in our pipeline.
Note that the qualitative results are best viewed in the supplementary video. 

\subsection{Experimental Setup}
\paragraph{Platform}
The implementation and experiments are conducted using a single NVIDIA 5070 Ti Super GPU with PyTorch 2.9 and CUDA 12.8. 
Grasp optimization is performed once per contact phase and takes 3 to 4 minutes per hand for 6000 steps, with its computation time dominated by the number of contact phases rather than the duration of each phase.
The diffusion noise optimization for full-body refinement is processed in 6-second windows, taking approximately 2 minutes per window for 500 steps, and scales proportionally with the number of windows. 

\paragraph{Dataset and Implementation Details.} For training the contact transition module (Sec.~\ref{sec:hand_motion_seq}) in Stage 1, which generate full hand motion sequences from wrist trajectories and grasping hand poses, we follow the settings in~\citet{wu2025hoifhli} using the GRAB dataset~\cite{taheri2020grab} for training. For the diffusion-based motion prior used in body motion refinement (Sec.~\ref{sec:diffusion_opt}), a motion diffusion model was trained on the HumanML3D~\cite{bensabath2024humanml3d} dataset. 
The hyperparameters used in the experiments are provided in the appendix.

\paragraph{Metrics.}
Throughout the experiments, we use the following quantitative metrics to evaluate the performance of our method from two perspectives: (1) how well the model recovers plausible contact from noisy input, and (2) how natural the enhanced motion is. 

For evaluating contact plausibility, we adopt metrics from \cite{li2023omomo, li2023chois}, including \textbf{precision ($C_{prec}$)}, \textbf{recall ($C_{rec}$)}, and \textbf{F1 score}. Following previous work, we apply an empirically determined contact threshold (6cm) between the palm joint and the object to generate frame-by-frame contact labels. 
We apply the same procedure to the ground-truth hand positions and compute the metrics based on true positive, false positive, and false negative counts with respect to the ground truth. Higher values for these metrics indicate more plausible contact behaviors.

To evaluate the quality of grasping hand poses during hand-object contact, we use the following metrics adopted from prior work~\cite{grady2021contactopt}.
\textbf{Contact Coverage (CC)} measures the percentage of hand vertices close to the object surface.
\textbf{Penetration} quantifies hand-object interpenetration by measuring SDF values of object vertices inside the hand mesh (multiplied by -1 for positive values).

To evaluate motion naturalness, we use the following metrics: \textbf{Foot Skate (FS)} quantifies foot sliding artifacts during ground contact. \textbf{Jitter} measures motion smoothness by computing acceleration changes averaged across all body joints. 
\textbf{InterPene} indicates the percentage of frames with interpersonal collisions, where collisions are detected using the collision geometries placed at the joints. Lower values across these metrics correspond to more natural motion.
%

\subsection{Improving Existing MHOI Dataset}
\label{sec:existing_improve}
To evaluate the effectiveness of our approach, we apply our method across different types of MHOI data acquired from existing mocap systems and conduct a systematic analysis of motion artifacts that exist in the data, as well as an analysis of how much our pipeline improves the data quality.
\paragraph{Dataset Setup.}
We leverage the CORE4D dataset~\cite{liu2024core4d} which features multi-human interactions involving diverse collaborative object manipulation scenarios such as jointly carrying, rotating and handing over objects.
To comprehensively evaluate how our pipeline handles varying data quality and possible artifacts that arise from different capture methods, we analyze our approach using three variants of the CORE4D dataset which are as follows:
\begin{itemize}
    \item \textit{CORE4D-Original}, where human poses are recorded with an IMU-based mocap suit and retargeted into SMPL-X parameters through per-frame optimization, while object trajectories are tracked using optical markers within a multi-view capture setup.
    \item \textit{CORE4D-Noisy}, a synthetic variant where we deliberately introduce additional noise to simulate lower-quality capture conditions. Specifically, we apply Gaussian-filtered uniform noise (15cm std) to global root translations of the human and filtered uniform noise (2cm std) to joint positions.
    \item \textit{CORE4D-HMR}, obtained by applying monocular video-based motion capture methods to the allocentric (exo-centric) third-person videos provided in the CORE4D dataset. We use SLAHMR~\cite{ye2023slahmr} that supports multi-person tracking to extract human motion from the monocular videos, and use root positions in the XZ plane of \textit{CORE4D-Original} to align SLAHMR results to the same coordinates with the object trajectory.
\end{itemize}


\begin{figure*}
\centering
\includegraphics[width=1.0\linewidth, trim={0 0.0cm 0 0}]{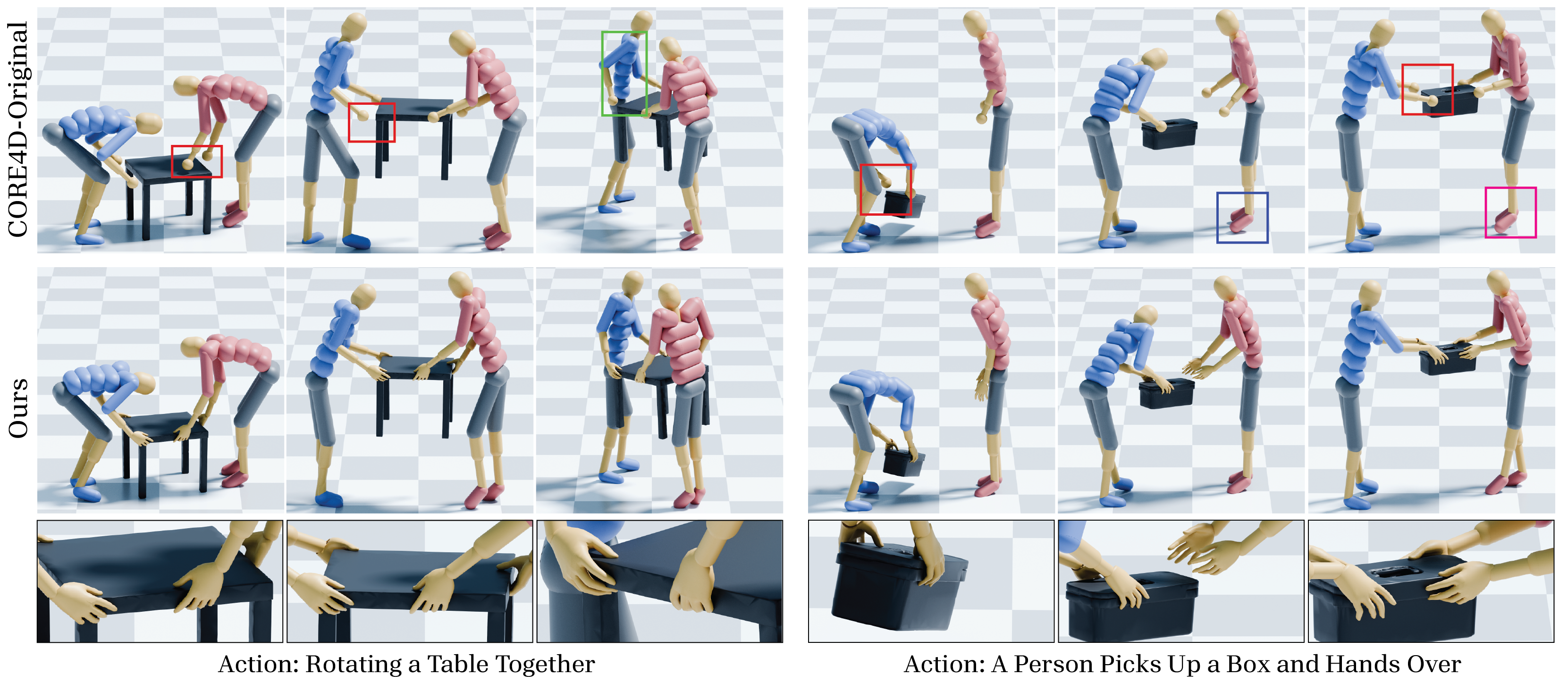}
    \caption{Qualitative comparison: CORE4D-\textit{Original} and Ours. }
    \label{fig:result_core4dorig}
\end{figure*}

\begin{figure*}
\centering
\includegraphics[width=1.0\linewidth, trim={0 0.0cm 0 0}]{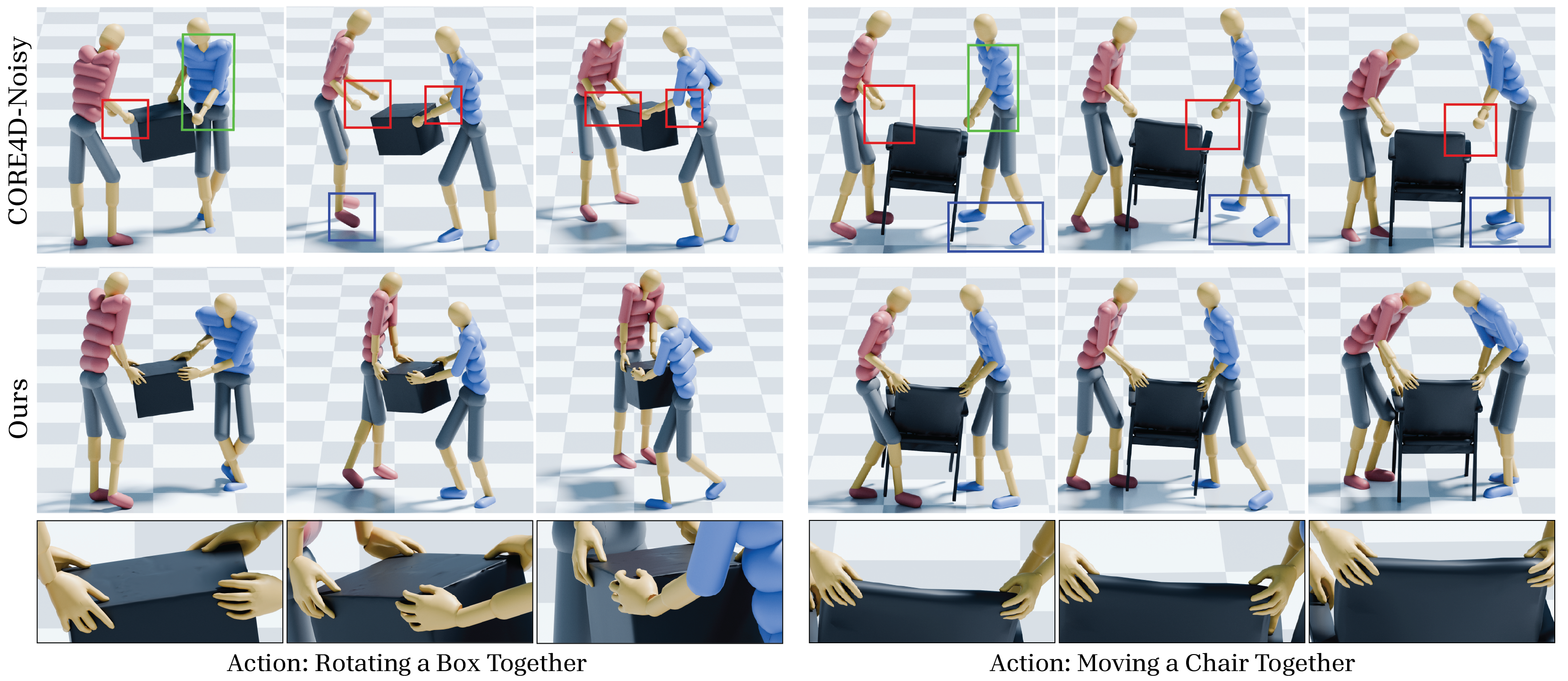}
    \caption{Qualitative comparison: CORE4D-\textit{Noisy} and Ours. }
    \label{fig:result_core4dnoisy}
\end{figure*}


\begin{table}
\footnotesize
    \centering
    \caption{(Top) Quantitative comparison between existing MHOI datasets (CORE4D, CORE4D-Noisy) and our pipeline results. CORE4D-N denotes CORE4D-Noisy. (Bottom) Quantitative comparison between generative model outputs (OMOMO) and our pipeline results.}
    \begin{tabular}{l|c|c|c|c|c|c}
    \toprule
        Method & $C_{prec}$$\uparrow$ & $C_{rec}$$\uparrow$ & F1$\uparrow$ & FS$\downarrow$ & Jitter $\downarrow$&InterPene$\downarrow$\\
        \midrule
 CORE4D& -& -& - & 0.56 & 71.25 & \textbf{0.00} \\
     CORE4D + Ours& -& -& - & \textbf{0.17} & \textbf{19.89} & 0.40\\
     \midrule
    CORE4D-N & 0.77 & 0.53 & 0.79 & 0.93 & 73.05 & \textbf{0.71} \\
    CORE4D-N + Ours& \textbf{0.87} & \textbf{0.98} & \textbf{0.91} & \textbf{0.21} & \textbf{19.79}& 0.75\\
    \midrule
    CORE4D-HMR & \textbf{0.76}& 0.36& 0.44& 1.27& 83.04& \textbf{0.32}\\
    CORE4D-HMR + Ours & 0.74& \textbf{0.67}& \textbf{0.66}& \textbf{0.38}& \textbf{22.87}& 0.66\\
    \bottomrule
    \toprule
    OMOMO-2P & 0.72 & 0.70 &  0.71 & 0.51 &113.01&2.49\\
    OMOMO-2P + Ours & \textbf{0.76}& \textbf{0.85}&  \textbf{0.78}&\textbf{0.16} & \textbf{15.40} &\textbf{0.69}\\
    \bottomrule
    \end{tabular}
    \label{tab:quant_c4d_variants}
\end{table}

\begin{table}
    \centering
    \small
    \caption{Quantitative comparison between CORE4D hand data and hand motions generated by our method.}
    \begin{tabular}{l|c|c}
    \toprule
        Method & Contact Coverage $\uparrow$& Penetration $\downarrow$ \\
    \midrule
 CORE4D w/ Hand Mocap & 6.72 & 0.53 \\
Ours w/ CORE4D Body Input & \textbf{17.13} & \textbf{0.48} \\
    \bottomrule
    \end{tabular}
    \label{tab:c4d_hand_comp}
\end{table}

\begin{figure}
\centering
\includegraphics[width=1.0\linewidth, trim={0 0.0cm 0 0}]{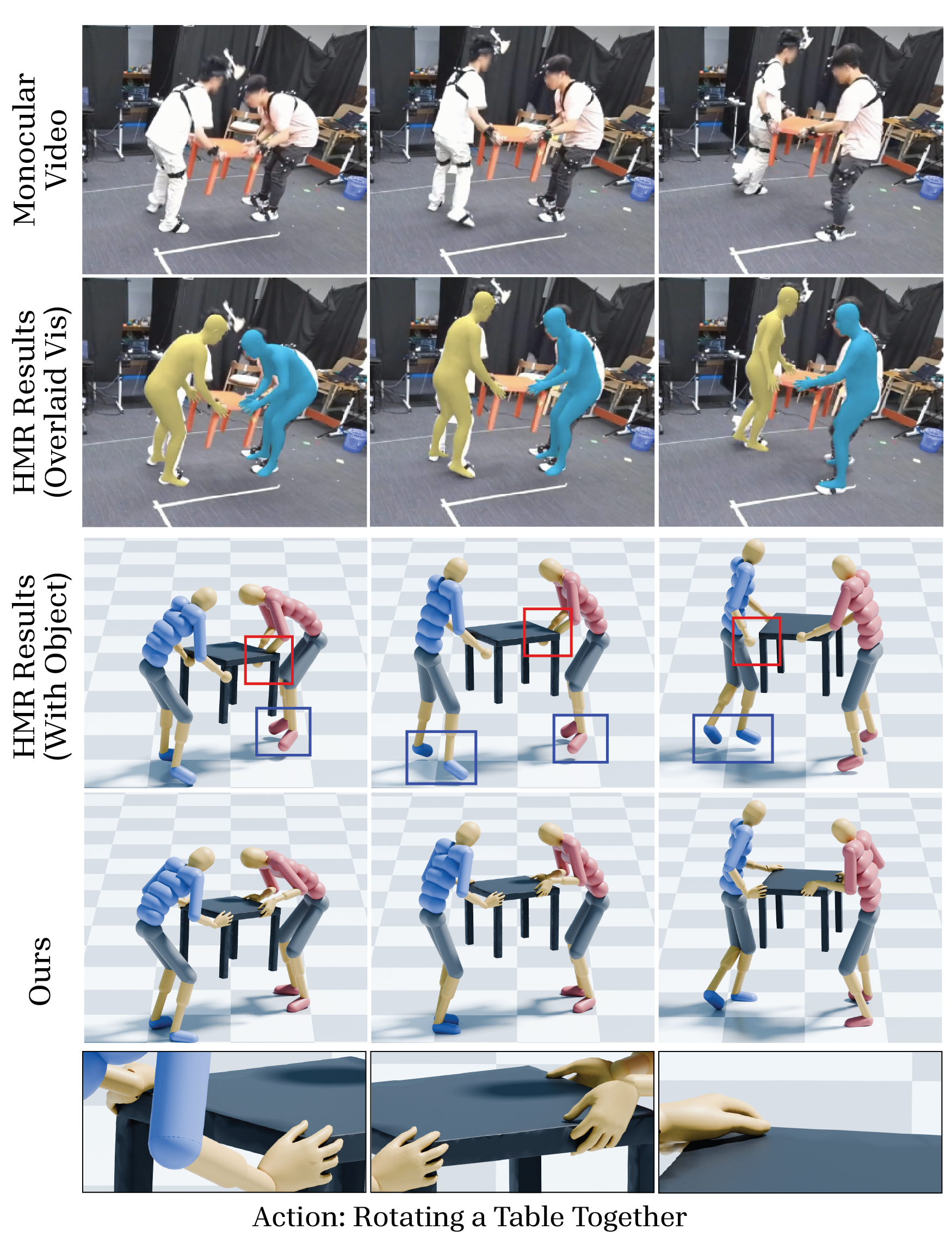}
    \caption{Qualitative comparison among CORE4D-\textit{HMR} and Ours. 
    The first row shows monocular video input from the CORE4D dataset. The second row presents SLAHMR results overlaid onto the input images for visualization, while the third row displays the MHOI motion obtained by aligning HMR results to the same coordinate system as the object trajectories in the corresponding CORE4D data.}
    \label{fig:result_slahmr}
\end{figure}

\paragraph{Results for Body Refinement}
Table~\ref{tab:quant_c4d_variants} (top) presents the quantitative comparison results across all three variants.
For \textit{CORE4D-Original}, we omit contact metrics since we use its contact labels as ground truth.
The original data exhibits foot skating and jitter artifacts, primarily due to the process of retargeting raw mocap data to SMPL-X~\cite{pavlakos2019smplx} parameters via per-frame optimization, which are significantly reduced with our method.
For \textit{CORE4D-Noisy}, the artifacts are more severe due to the synthetic noise added to the data, showing not only increased foot skating and jitter but also temporally inconsistent and physically implausible contacts caused by the added noise.
\textit{CORE4D-HMR} exhibits similar levels of artifacts as \textit{CORE4D-Noisy}.
Even for these more challenging capture setups, our method consistently improves both contact plausibility and motion naturalness.

Visual comparison results are presented in Figure~\ref{fig:result_core4dorig}, Figure~\ref{fig:result_core4dnoisy}, and Figure~\ref{fig:result_slahmr}. 
The original MHOI data exhibit several artifacts, including foot hovering and slipping (shown in blue), unnatural body poses (shown in green), and inconsistent contact states (shown in red). 
Consistent with the quantitative analysis, our method effectively corrects these issues, producing more natural poses and refining inconsistent contacts and foot sliding into physically plausible behavior.
Although the results look plausible where the HMR-estimated human poses overlaid on the video in Figure~\ref{fig:result_slahmr}, depth ambiguities inherent to single-view images lead to inconsistent contacts and foot artifacts in 3D space, which are also effectively addressed by our refinement process.

\begin{figure}
\centering
\includegraphics[width=1.0\linewidth, trim={0 0.0cm 0 0}]{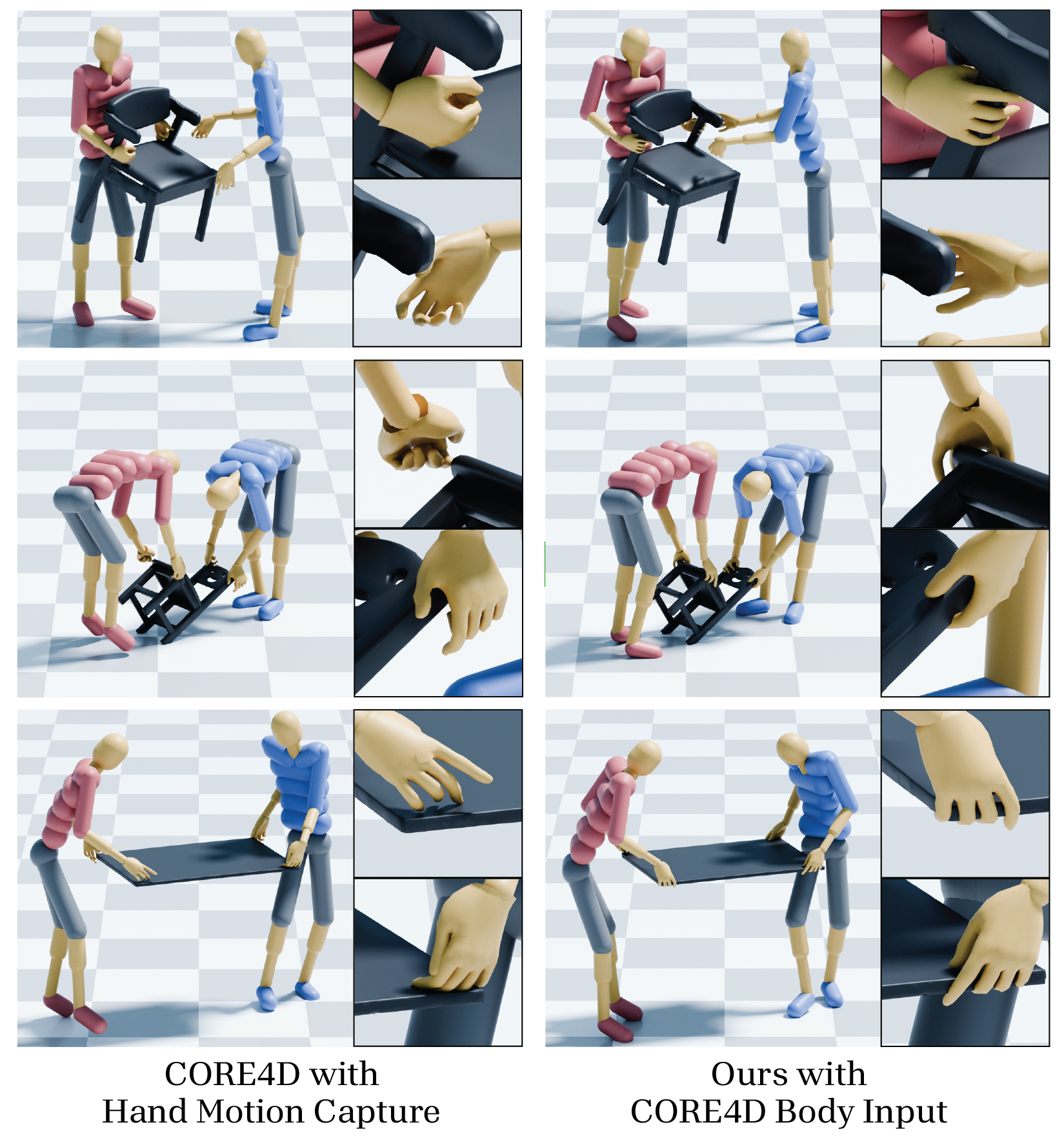}
    \caption{Qualitative comparison between CORE4D data with hand motion capture with hand motion synthesized by our method with body-only inputs.}
    \label{fig:result_handcomp}
\end{figure}

\paragraph{Results for Hand-Object Interaction Generation}
Using the hand motion data in the CORE4D dataset as a comparison baseline, we evaluate whether our method can generate plausible hand motions from body motion alone. The CORE4D dataset includes finger movements captured using motion capture gloves and integrated with body motion through an optimization-based retargeting process that transforms them into global coordinates consistent with the object trajectory.
Specifically, we use the body motion from \textit{CORE4D-Original} as input to our method and compare the generated hand motions against the original CORE4D hand data for a fair comparison. 
Table~\ref{tab:c4d_hand_comp} presents quantitative results, where our synthesized hand motions outperform the original CORE4D data in both contact coverage and penetration metrics, with particularly substantial improvements in contact coverage. 
Figure~\ref{fig:result_handcomp} provides visual comparisons: due to retargeting and coordinate alignment errors, the hands captured in the CORE4D data exhibit distorted fingers. 
Notably, they show incorrect wrist poses and orientations that cause the fingers to float above objects rather than maintaining proper contact, which is consistent with the low contact coverage scores in Table~\ref{tab:c4d_hand_comp}.
In contrast, our method generates hands that achieve plausible and stable grasps with proper object contact.

\begin{figure}
\centering
\includegraphics[width=1.0\linewidth, trim={0 0.0cm 0 0}]{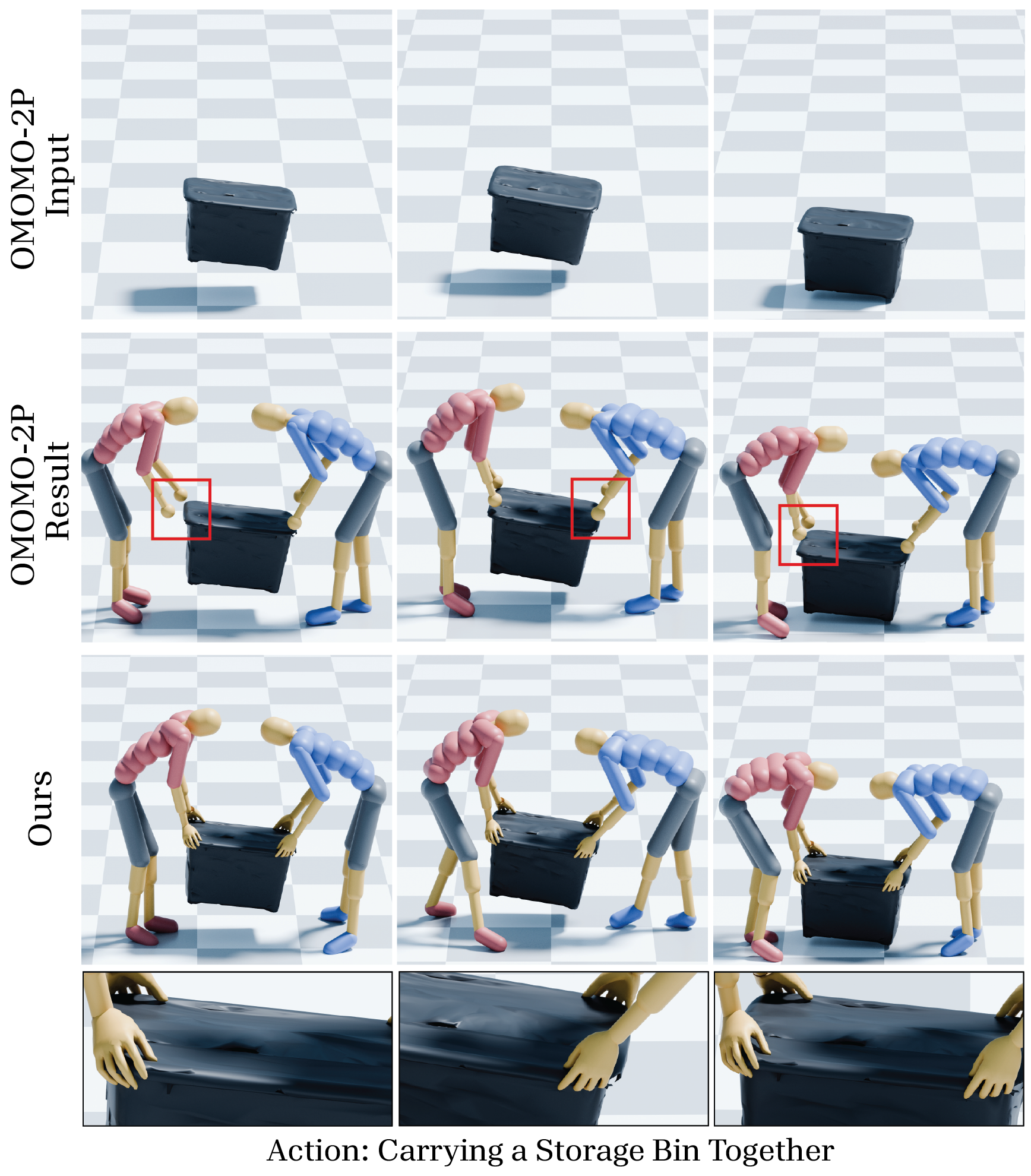}
    \caption{Qualitative results of enhancing OMOMO model's MHOI output. }
    \label{fig:result_omomo}
\end{figure}

\subsection{Post-processing Pipeline for Generative Models}
\label{sec:gen_model_process}
Our method can also be used as a post-processing pipeline to enhance outputs of generative models, allowing us to generate unlimited high-quality MHOI data from limited training samples.

\paragraph{Model Setup.}
The integration of our pipeline into generative models can be formally described as follows. Given any conditional input $\mathbf{C}$ and a generative model $\mathcal{F}$ that produces an MHOI sequence $\mathcal{I}$, augmenting the model pipeline with our module yields an enhanced MHOI sequence $\hat{\mathcal{I}}$.
In our experiment, we use OMOMO~\cite{li2023omomo} as our generative model $\mathcal{F}$. 
OMOMO originally synthesize single-person object interactions from object trajectory $\mathbf{O}$ only via a two-stage diffusion model. We modify the model's output to generate two-person motions $\mathbf{M}^i$ and $\mathbf{M}^j$ and train the model with the CORE4D dataset.
Note that the OMOMO model was trained to produce the body motion only, which serves as input to our enhancement pipeline.

\paragraph{Results.}
Table~\ref{tab:quant_c4d_variants} (bottom) shows quantitative comparison results, where precision, recall, and F1 scores are computed against the ground-truth sequences provided by the dataset. A visual comparison is shown in Figure~\ref{fig:result_omomo}. Both results demonstrate that our method improves the results: the body motion is refined with consistent contact and fewer artifacts, while plausible hand motion is synthesized despite only body motion being provided as input.



\subsection{Augmentation to Object Variations}
\paragraph{Formulation.}
Our method can also serve as an data augmentation pipeline by replacing a given object in a scene with different, yet class-consistent alternatives.
Specifically, given a noisy MHOI data $\mathcal{I} = \{\mathbf{M}^i, \mathbf{M}^j, \mathbf{O}, \mathbf{G}_\text{obj} \}$, the input for augmentation is defined by $\mathcal{I}'_k = \{\mathbf{M}^i, \mathbf{M}^j, \mathbf{O}, \mathbf{G}'_\text{obj}\}$, where $\mathbf{G}'_\text{obj}$ is a new object different from the original object.
Then, our method is applied to the new input to generate the augmented data $\hat{\mathcal{I}}' = \{\hat{\mathbf{M}}^i, \hat{\mathbf{M}}^j, \hat{\mathbf{O}}, \mathbf{G}'_\text{obj}, \mathbf{H}^i, \mathbf{H} ^j\}$. This process can be repeated for many different objects.

\begin{figure}
\centering
\includegraphics[width=0.9\linewidth, trim={0 0.0cm 0 0}]{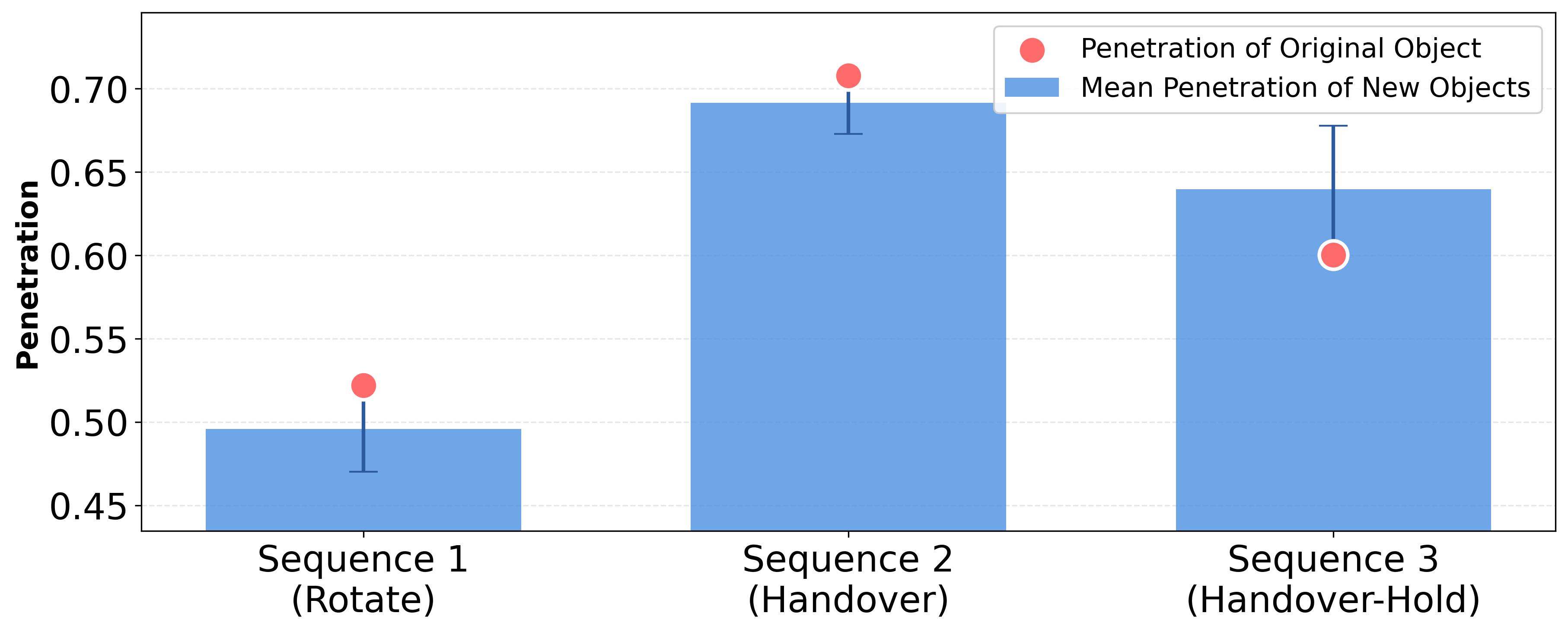}
    \caption{Quantitative results (penetration metric) of augmenting chair-related MHOI sequences (rotate, handover, handover-hold) into new chair objects in the ShapeNet dataset. Blue bars indicate the mean penetration score of new objects per sequence, with error bars representing the standard deviation. Red dots show the penetration of the original object in each sequence. The results demonstrate that the penetration values of the original objects are similar to those of the new objects.}
    \label{fig:result_obj_retarget_quant}
\end{figure}

\begin{figure*}
\centering
\includegraphics[width=1.0\linewidth, trim={0 0.0cm 0 0}]{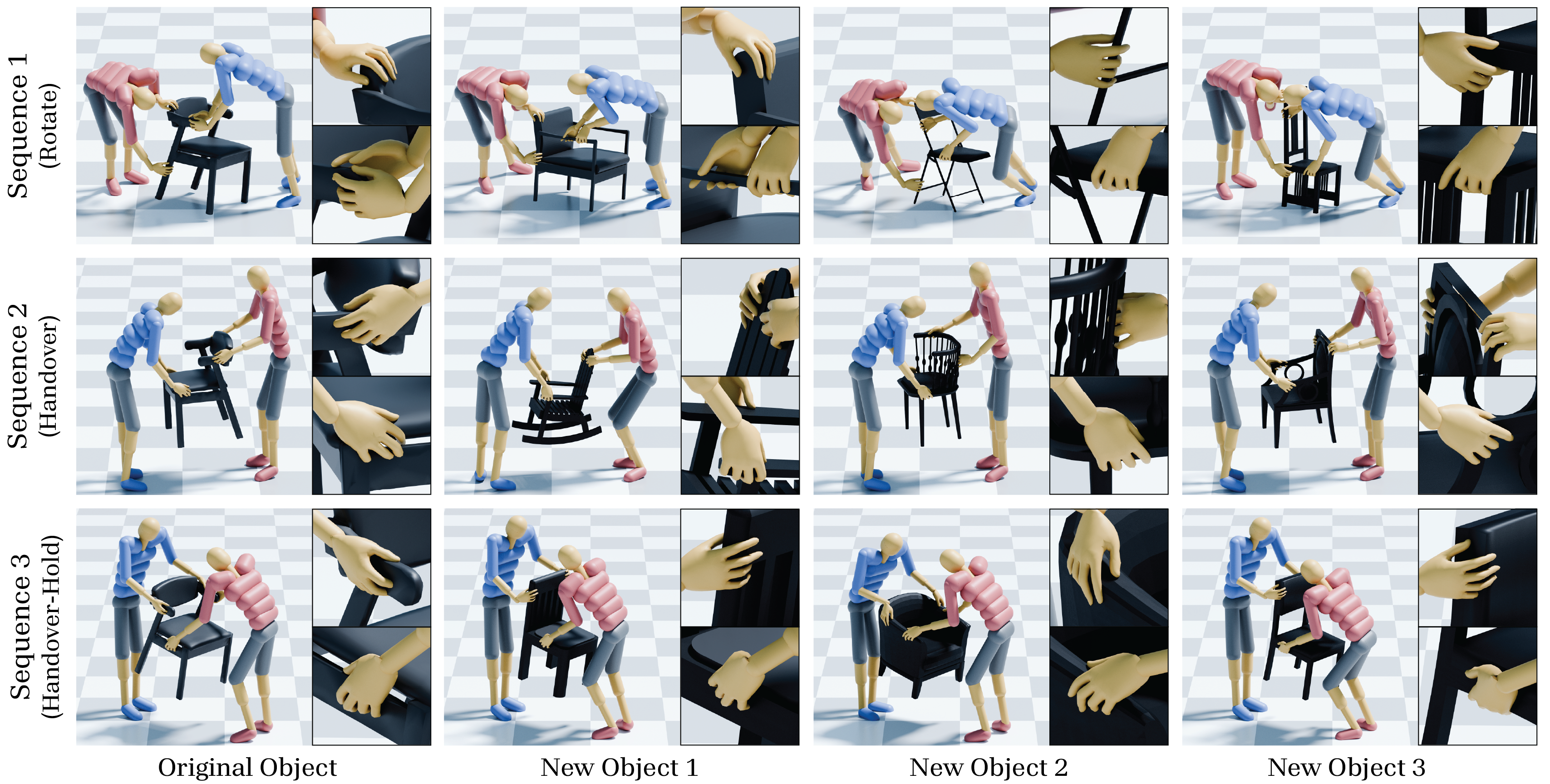}
    \caption{Qualitative results of augmenting chair-related MHOI sequences (rotate, hold, handover) into new chair objects in the ShapeNet dataset. }
    \label{fig:result_obj_retarget}
\end{figure*}

\paragraph{Results.} For evaluation, we utilize the chair sequence from the \textit{CORE4D-Original} data and apply our pipeline to three types of interactions with a chair-shaped object: "Rotate" (rotating a chair lying on the floor to an upright position), "Handover" (one person handing a chair to another), and "Handover-Hold" (one person hands over the chair, both hold it together briefly, then the receiver carries it away).
For quantitative evaluation, we augment each sequence with 9 different chair geometries from the ShapeNet~\cite{chang2015shapenet} dataset and measure penetration metrics compared to the original. 
The results in Figure~\ref{fig:result_obj_retarget_quant} show that the new objects achieve penetration scores within a similar range to the original object.\footnote{We do not consider contact coverage in this case, as different object geometries naturally lead to different plausible palm contact regions.} The visual results in Figure~\ref{fig:result_obj_retarget} demonstrate that our pipeline successfully generates plausible hand motions adapted to new objects while preserving the original interaction semantics.

\begin{figure*}
\centering
\includegraphics[width=0.95\linewidth, trim={0 0.0cm 0 0}]{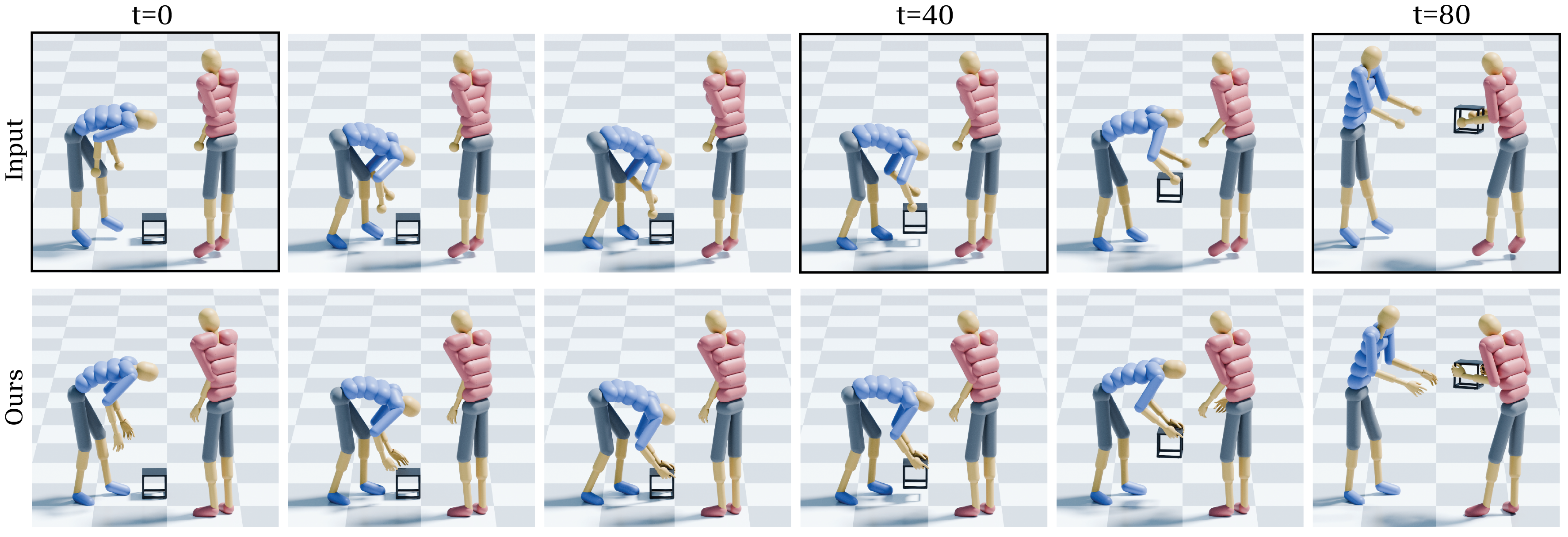}
    \caption{(Top) Initial motion constructed from keyframe interpolation. (Bottom) Enhanced MHOI motion generated by our pipeline. Frames with black boxes indicate the keyframes where poses were provided by the human annotator. The object was generated using the prompt "mini-size cube shaped table".}
    \label{fig:result_keyframe_boxhandover}
\end{figure*}

\begin{figure*}
\centering
\includegraphics[width=0.95\linewidth, trim={0 0.0cm 0 0}]{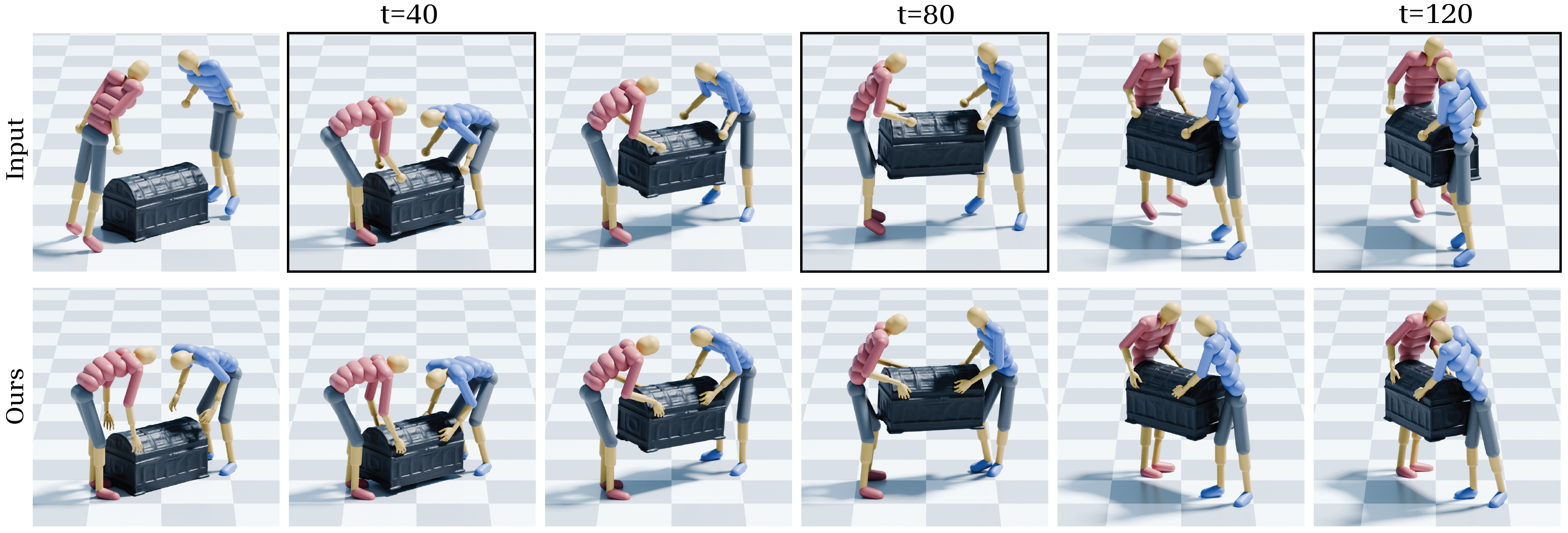}
    \caption{(Top) Initial motion constructed from keyframe interpolation. (Bottom) Enhanced MHOI motion generated by our pipeline. Frames with black boxes indicate the keyframes where poses were provided by the human annotator. The object was generated using the prompt "a large gothic style chest".}
    \label{fig:result_keyframe_tableturn}
\end{figure*}



\subsection{MHOI Creation with Keyframing}
\label{sec:result_keyframing}
In this section,  we demonstrate that our system can also serve as an (human-assisted) data creation pipeline for MHOI data and demonstrate the generalization ability of our method to arbitrary objects. 
In the creation process, 3D objects are generated using off-the-shelf text-to-3D models~\cite{lumaai}, and 6DoF object poses along with human poses are roughly specified by a human annotator at sparse keyframes with 40-frame intervals at 30 FPS. 
The object trajectory and human motion are initially constructed through keyframe interpolation in Blender~\cite{blender}. 
While this interpolation-based initialization produces unnatural skating artifacts and lacks coordination between object trajectory and human pose interpolation, the visual results in Figure~\ref{fig:result_keyframe_boxhandover} and Figure~\ref{fig:result_keyframe_tableturn} show that our pipeline successfully refines the unnatural body motion into natural interactions with the object. Furthermore, the results demonstrate that our method generates natural interaction motions even for arbitrary objects created through text-to-3D models.

\subsection{Extensions to Diverse MHOI Scenarios}
\label{sec:result_extend}
\begin{figure}
\centering
\includegraphics[width=1.0\linewidth, trim={0 0.0cm 0 0}]{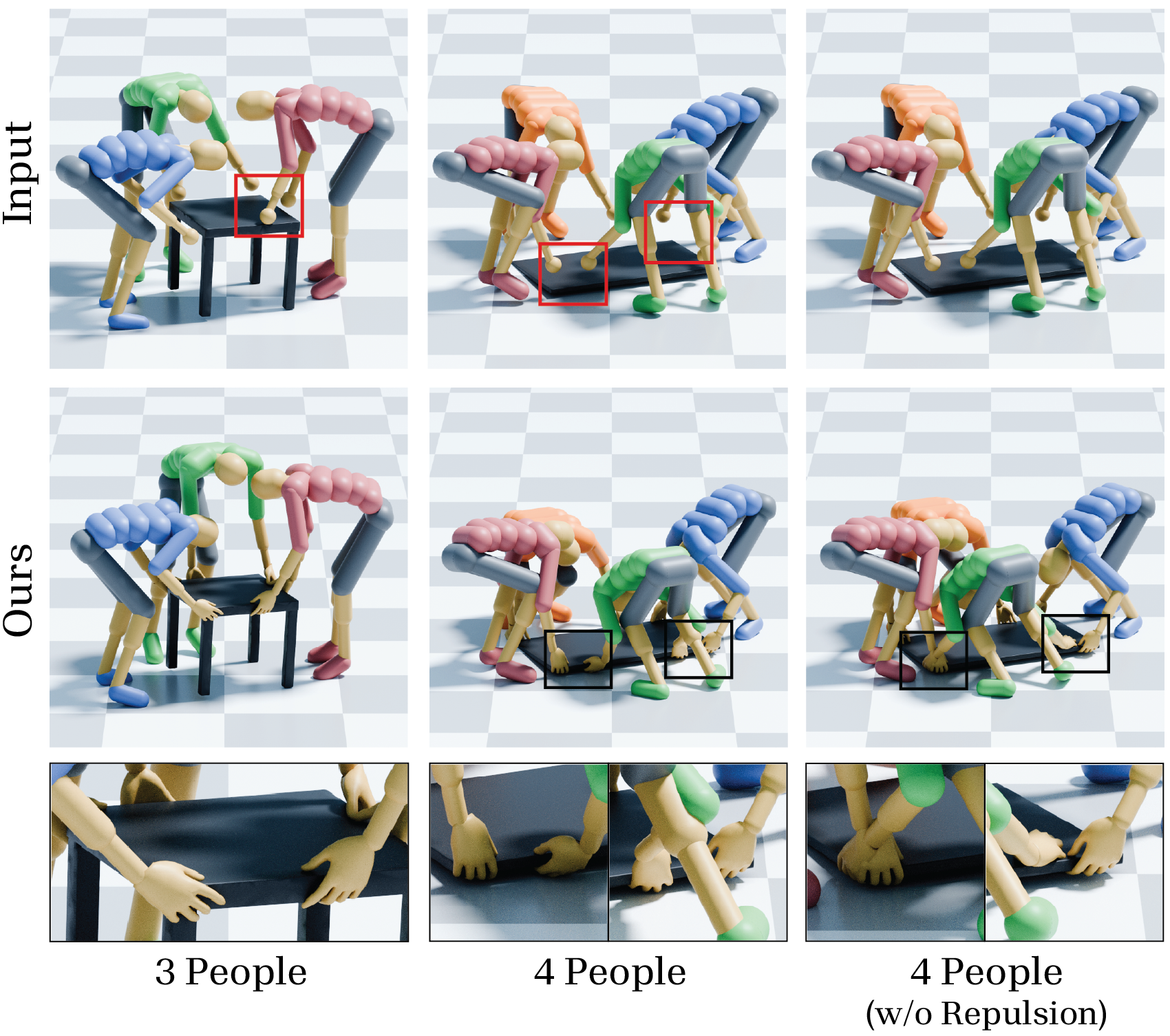}
    \caption{Results of enhancing MHOI motions with more than two participants. The first row shows results with three participants, the second row with four participants, and the third row with four participants without adding repulsion vectors to the cylinders during grasp optimization. 
    }
    \label{fig:result_multip}
\end{figure}

\paragraph{Multi-Person Collaboration (>2 People).}
While previous experiments focused on two-person collaborative object manipulation, this experiment demonstrates that our pipeline is also applicable to MHOI scenarios with three or more people.
When multiple people (e.g., four) simultaneously manipulate a single object, collisions between their grasping hands become more likely due to the increased number of participants. 
To address this, we introduce an additional processing step during arm cylinder construction in the hand pose initialization of the grasp generation stage (Stage 1, Section~\ref{sec:batch_optim}). 
When arm cylinders overlap, we apply repulsion vectors to the wrist positions to push them away from collision regions. 
This produces updated, collision-free cylinders from which we perform initialization and grasping pose optimization.
Figure~\ref{fig:result_multip} shows visual results of 3-4 people collaboratively moving an object. 
In Figure~\ref{fig:result_multip} (middle), the repulsion vectors prevent hand overlap between the green and pink colored subjects, whereas without repulsion vectors (Figure~\ref{fig:result_multip}, right), hand overlap produces unnatural results. 
This demonstrates the effectiveness of our approach to prevent inter-person collisions during grasping.

\paragraph{Mixed Static and Dynamic Object Interactions.}
Since our method encompasses both hand–object interaction generation and body motion refinement, previous experiments primarily focused on object manipulation and rearrangement tasks such as carrying and handovers. In this experiment, we demonstrate that our approach extends to a broader range of human–object interactions, including both static (e.g., sitting on a chair) and dynamic cases (e.g., object carrying).

\begin{figure}
\centering
\includegraphics[width=1.0\linewidth, trim={0 0.0cm 0 0}]{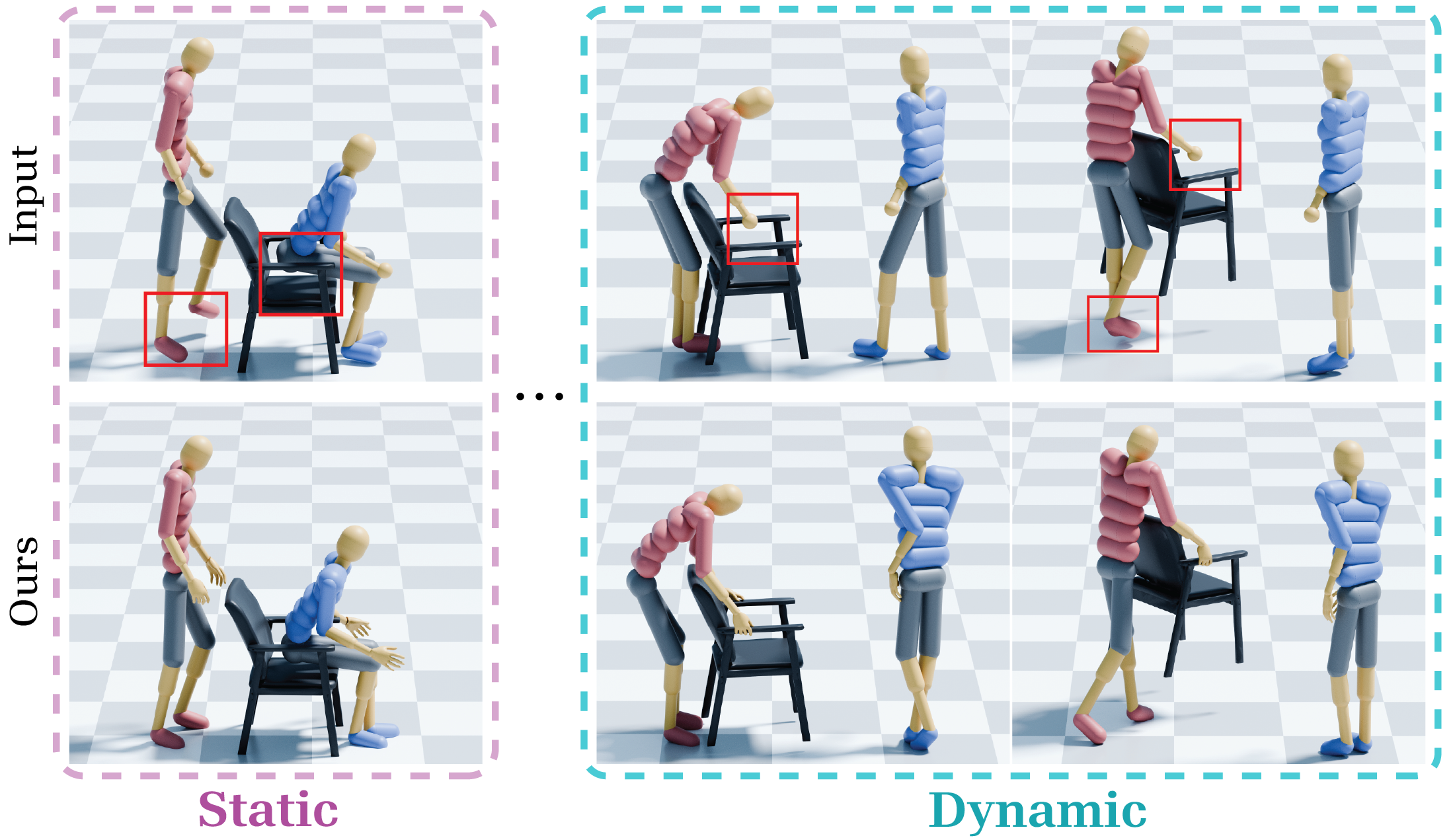}
    \caption{Qualitative results with MHOI motion where static object interaction (sitting on a chair) and dynamic interaction (moving the chair) are both included. 
    }
    \label{fig:result_stadyn}
\end{figure}

We evaluate our method on sequences involving seated interactions and object manipulation. 
Figure~\ref{fig:result_stadyn} shows a sequence where one person (blue) sits on a chair while another (pink) approaches. The blue person then stands up, and the pink person moves the chair away.
For the sitting interaction, we modify the contact objective in Eq.~\ref{eq:body_refine_contact} by replacing the wrist joint with the hip joint to enforce contact between the chair and the seated person. 
As shown in Figure~\ref{fig:result_stadyn}, the input motion shows artifacts with the blue person hovering above the chair. Our method produces stable seated contact and removes the foot hovering artifact of the pink person. 
Furthermore, our approach successfully enhances the latter part of the sequence, which involves dynamic object interaction, where the pink person carries the chair. This example demonstrates the generalizability of our method to scenarios combining human–human interaction with both static and dynamic object interactions.

\subsection{Extensions to Single-Person HOI}

Although our method primarily targets multi-person HOI scenarios as the most challenging setting, it can be naturally extended to single-person HOI with minor modifications. 

\paragraph{Experimental Setup.}
Specifically, for single-person HOI enhancement, we disable the human-human interaction terms (the interaction graph and the inter-character collision term) during the full-body motion refinement stage by setting their weights to zero.
 
To evaluate our pipeline in the single-person setting, we adopt a setup as in Sec.~\ref{sec:gen_model_process}: we take body motion output from an existing single-person HOI generation method and use it as input to our pipeline. 
We set HOIDiNi~\cite{ron2025hoidini} as the generation baseline, which is relevant to ours in that it also leverages diffusion noise optimization during the generation process.
It is notable that the two methods address fundamentally different tasks: HOIDiNi is a \textit{synthesis} method that takes text as input and generates interaction semantics (e.g., contact points and object trajectories) from scratch, while our method assumes a noisy HOI motion as input and extracts interaction semantics from it for \textit{enhancement}. 
Due to this difference in input modality and problem formulation, a direct comparison under equivalent inputs is nontrivial, thus leading to adapt the complementary setting as in Sec.~\ref{sec:gen_model_process}.
For our experiments, we use the GRAB dataset~\cite{taheri2020grab} which HOIDiNi is trained on, and we use its official implementation and demo sequences from the test split. 

\begin{figure}
    \centering
    \includegraphics[width=1.0\linewidth, trim={0 0.0cm 0 0}]{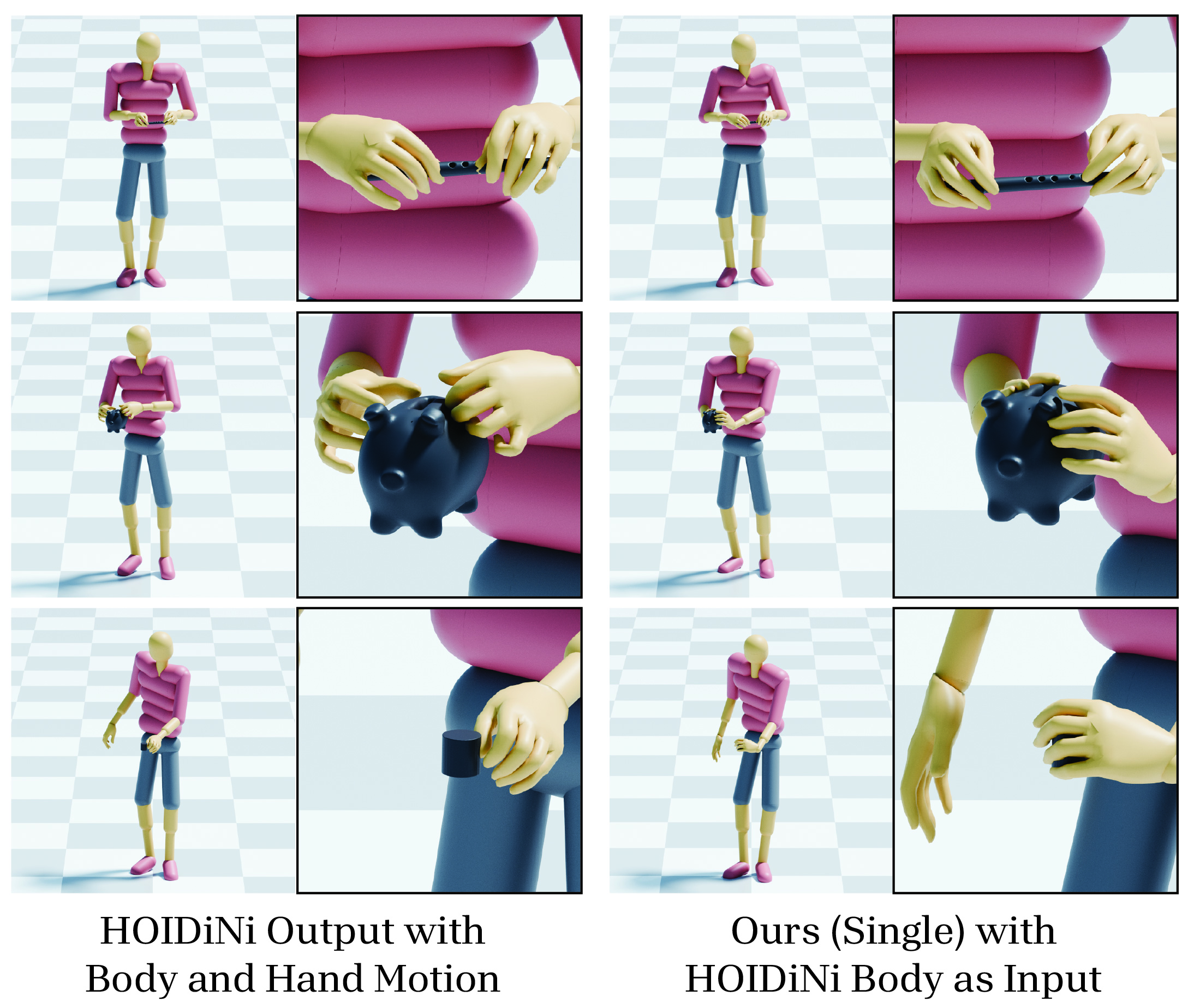}
    \caption{
    Left: HOIDiNi output with jointly generated body and hand motion. Right: our enhanced result using our single-person variant with HOIDiNi's body motion as input. 
    }
    \label{fig:single_hoidini}
\end{figure}

\paragraph{Results}
Fig.~\ref{fig:single_hoidini} (right column) presents visual results of our single-person HOI enhancement, demonstrating that our pipeline generalizes well beyond the multi-person setting and produces plausible hand-object interactions.

\paragraph{Hand-Object Interaction Comparison.}
Although a direct comparison under equivalent inputs is nontrivial as discussed above, HOIDiNi jointly generates body and hand motion within a single framework, 
which allows us to compare hand motion quality during object interaction.
We therefore adopt an evaluation setup as in Table~\ref{tab:c4d_hand_comp}, comparing the quality of hand motion during object interaction between our enhanced outputs and HOIDiNi's output motion.
As shown in Table~\ref{tab:single_hoidini_comp}, the hand-object interactions generated by our method achieves higher contact coverage and lower penetration. 
Fig.~\ref{fig:single_hoidini} further illustrates this comparison: the left column shows HOIDiNi's output while the right column shows our enhanced result, where our output exhibits more plausible hand-object contact and reduced interpenetration. 
The results show that our decomposition approach which consists of hand-object interaction and full-body refinement leads to better interaction quality.

\begin{table}
    \centering
    \small
    \caption{Quantitative comparison on single-person HOI between HOIDiNi and our enhancement pipeline applied to HOIDiNi body output.}
    \begin{tabular}{l|c|c}
    \toprule
        Method & Contact Coverage $\uparrow$ & Penetration $\downarrow$ \\
    \midrule
        HOIDiNi Output (Body + Hands) & 22.40 & 0.25 \\
        Ours (Single) w/ HOIDiNi Body  & \textbf{24.54} & \textbf{0.22} \\
    \bottomrule
    \end{tabular}
    \label{tab:single_hoidini_comp}
\end{table}

\subsection{Ablations}

\begin{table}
    \centering
    \small
    \caption{Ablations on using wrist poses (from grasping hand optimization) during contact in noise optimization.}
    \begin{tabular}{l|c|c|c}
    \toprule
        Method & $C_{prec}$$\uparrow$ & $C_{rec}$$\uparrow$&  F1 Score$\uparrow$\\
    \midrule
 Ours w/o Hand Stage and $\mathcal{L}_\text{contact}$& 0.93& 0.73& 0.81\\
 Ours & \textbf{0.94}& \textbf{0.94}& \textbf{0.94}\\
    \bottomrule
    \end{tabular}
    \label{tab:abl_contact}
\end{table}

\begin{table}
    \centering
    \small
    \caption{Ablations on optimization objectives in noise optimization. 
    }
    \begin{tabular}{l|c|c|c|c|c}
    \toprule
    Method & MPJPE$\downarrow$  & Root PE$\downarrow$ &FS$\downarrow$& Jitter$\downarrow$&InterPene$\downarrow$ \\
    \midrule
    GT& -& -& 1.04& 65.11&0.43\\
    Ours w/o $\mathcal{L}_\text{pose}$&  20.13& 12.57& 0.23& \textbf{14.12}& 1.12\\
    Ours w/o $\mathcal{L}_\text{IG}$ & 9.44& 7.82& 0.20& 
 \underline{18.49}& 0.72\\
    Ours w/o $\mathcal{L}_\text{col}$&  \textbf{9.03}&  \underline{7.45}& \underline{0.20}& 19.05& \underline{0.27}\\
    Ours &  \underline{9.16}& \textbf{7.40}& \textbf{0.19}& 18.52& \textbf{0.24}\\
    \bottomrule
    \end{tabular}
    \label{tab:abl_dno}
\end{table}

\begin{figure}
\centering
\includegraphics[width=1.0\linewidth, trim={0 0.0cm 0 0}]{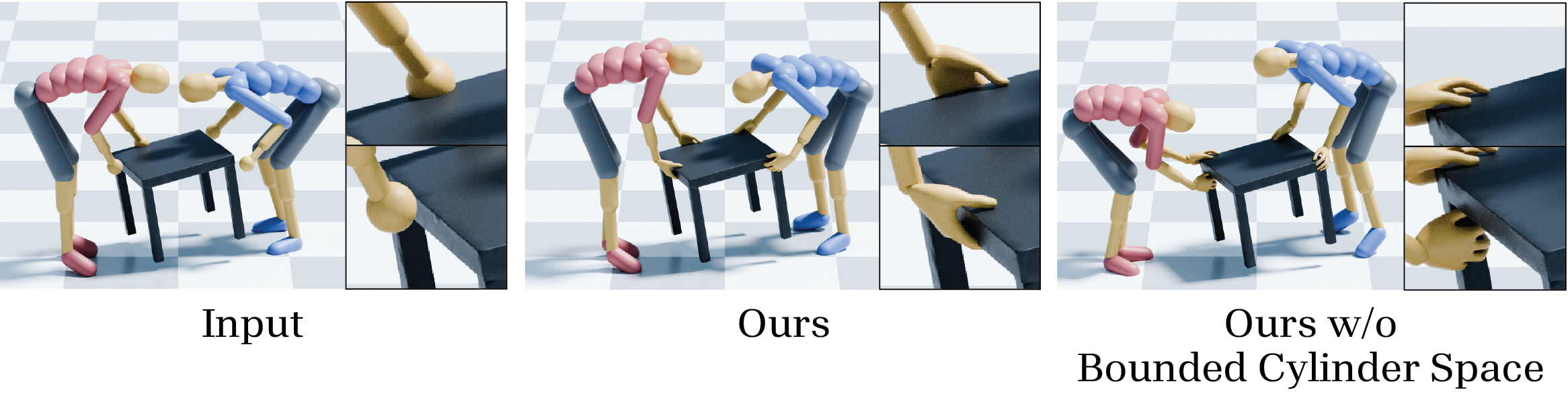}
    \caption{Qualitative comparison between our method (middle) and random initialization baseline (right). 
    }
    \label{fig:abl_cylinder}
\end{figure}

\paragraph{Using Cylindrical Bounds for Grasp Optimization.}
Constructing cylindrical volumes to bound the solution space is crucial for generating hand-object interactions that are consistent with the body motion in our grasp generation optimization (Sec.~\ref{sec:contact_refine}). We conduct an ablation study to validate its effectiveness.
As an ablative baseline, we skip the cylinder construction and instead initialize hand positions using random points sampled 3-5cm away from the averaged input wrist position with random orientations. We then optimize these positions using all other loss terms, excluding the cylinder-based regularization loss (Eq.~\ref{eq:cylinder_regloss}). 
Figure~\ref{fig:abl_cylinder} shows that the baseline produces grasps inconsistent with the input body motion. While the pink person in the input motion grasps the middle of the table, the baseline generates a grasp on the table leg. Although this grasp is physically plausible, it is semantically inconsistent with the input body motion. Consequently, when body motion refinement is performed using these baseline grasps, the output motion exhibits a different interaction context from the input. In contrast, our full pipeline successfully preserves the original interaction semantics throughout the refinement process.

\begin{figure}
\centering
\includegraphics[width=1.0\linewidth, trim={0 0.0cm 0 0}]{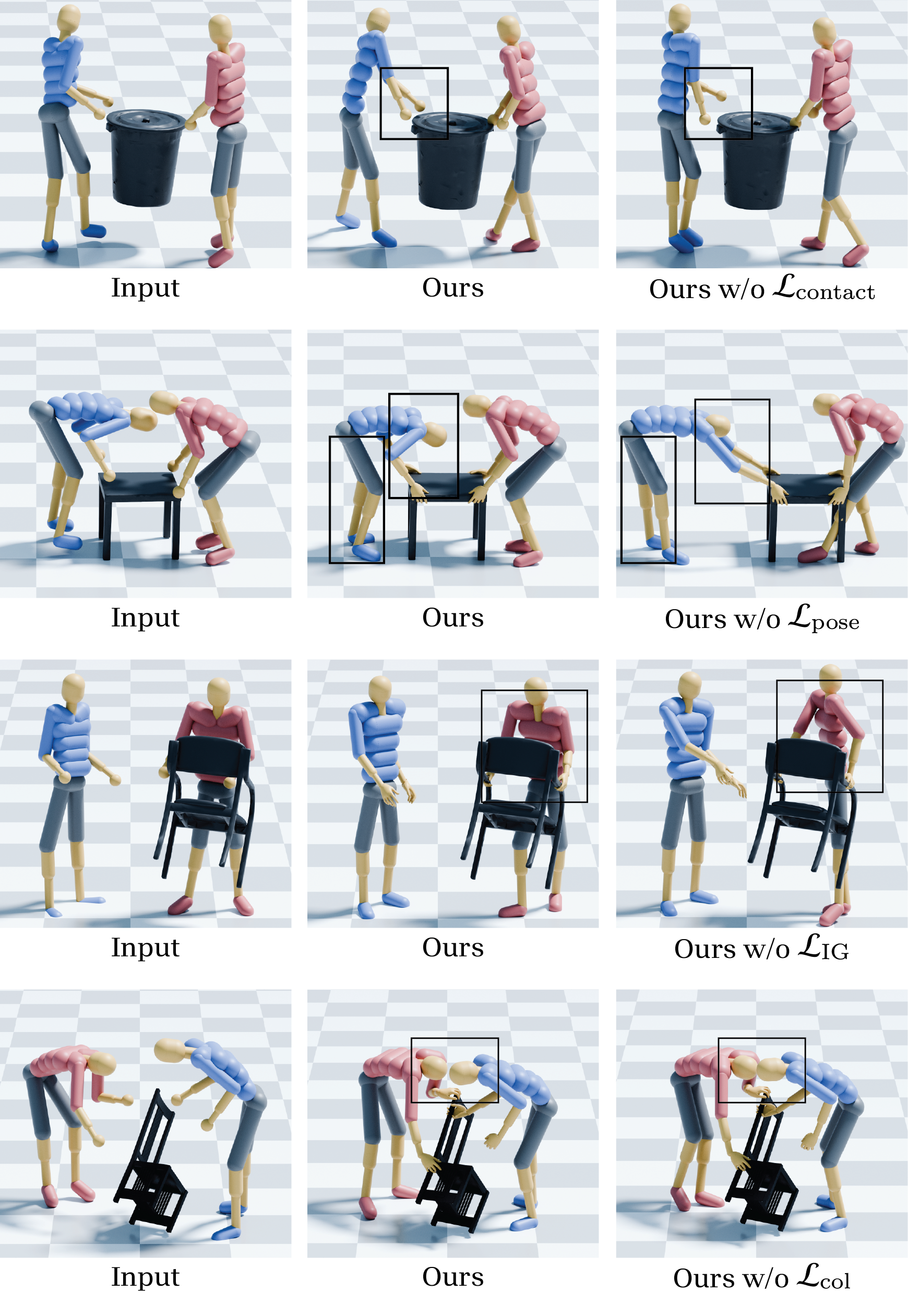}
    \caption{Qualitative ablation study of diffusion optimization objectives. From top to bottom: (1st row) without $\mathcal{L}_\text{contact}$, (2nd row) without $\mathcal{L}_\text{pose}$, (3rd row) without $\mathcal{L}_{\text{IG}}$, and (4th row) without $\mathcal{L}_{\text{col}}$.}
    \label{fig:abl_dno}
\end{figure}

\paragraph{Contact.} \label{ablatation:contact}
We evaluate the contribution of generating grasping hand motions in the hand-object interaction generation stage (Stage 1) and incorporating them into the motion refinement stage (Stage 2). Specifically, we compare our full pipeline with an ablative baseline where Stage 1 is removed along with the contact objective $\mathcal{L}_\text{contact}$ during motion refinement. 
The quantitative results in Table~\ref{tab:abl_contact} demonstrate that incorporating contact information derived from Stage 1 through the optimization objective $\mathcal{L}_{\text{contact}}$ leads to higher scores across contact-related metrics. 
Visual results are shown in the first row of Figure~\ref{fig:abl_dno}.\footnote{Note that we do not visualize finger motions for the full pipeline's result in Figure~\ref{fig:abl_dno} for a fair comparison.} Consistent with the quantitative results, our full method shows superior contact quality: in the full pipeline (ours) both the blue and red characters grasp the barrel, whereas the blue character's hands are detached from the object in the ablative baseline.
This demonstrates that hand motion generation in Stage 1 and adding $\mathcal{L}_{\text{contact}}$ in Stage 2 are essential for generating realistic human-object interactions in MHOI.

\paragraph{Diffusion Optimization Objectives.}
We quantitatively evaluate the contribution of each objective used in the diffusion-based optimization during the motion refinement stage (Stage~2), which aims to preserve both individual and interaction semantics while maintaining naturalness in the resulting motions.
Note that we exclude $\mathcal{L}_{\text{contact}}$ from this analysis, as its contribution is evaluated separately in Sec.~\ref{ablatation:contact}.
We use the \textit{CORE4D-Original} data for evaluation, and the quantitative results are presented in Table~\ref{tab:abl_dno}. In addition to FS, Jitter, and InterPene, we also report MPJPE (Mean Per Joint Position Error, in~$cm$) and Root PE (Root Position Error, in~$cm$) to evaluate how effectively the refinement preserves pose semantics relative to the ground truth. 
Additionally, Figure~\ref{fig:abl_dno} provides qualitative results illustrating the contribution of each objective.
\begin{itemize}
    \item Without $\mathcal{L}_\text{pose}$: The optimization produces natural motions but deviates significantly from the initial input (high MPJPE and Root PE), failing to preserve the original semantics of the MHOI sequence. As shown in the second row of Figure~\ref{fig:abl_dno}, while the baseline achieves reasonable human-object contact and approximate facing directions, the resulting poses substantially differ from the input motion.
    
    \item Without $\mathcal{L}_{\text{IG}}$: The interaction graph objective plays a crucial role in maintaining interaction semantics, as evidenced by increased MPJPE and Root PE when removed. As shown in the third row of Figure~\ref{fig:abl_dno}, given an input motion where two individuals face perpendicularly, the baseline without $\mathcal{L}_{\text{IG}}$ produces poses (pink person) with an unnaturally twisted torso. 
    
    \item Without $\mathcal{L}_{\text{col}}$: When the collision objective $\mathcal{L}_{\text{col}}$ is excluded, the method yields marginally lower MPJPE. This is because the optimization prioritizes adherence to the initial motion without considering collisions. In contrast, our full approach achieves superior collision avoidance. The fourth row of Figure~\ref{fig:abl_dno} shows head collisions between individuals when $\mathcal{L}_{\text{col}}$ is absent.
\end{itemize}

\section{Discussion}
\label{sec:discussion}
We present a two-stage framework for enhancing noisy multi-human object interaction (MHOI) data. Our approach first generates physically plausible hand grasps from noisy body poses through optimization, producing grasps that are both physically plausible and consistent with the given input body motion. These optimized grasps are then extended into complete hand-object interaction sequences via hand-object interaction priors. Full-body motion refinement for all participants is done in the next stage using a diffusion-based noise optimization framework built on single-person motion priors with interaction-aware optimization objectives.
Our experimental results demonstrate the effectiveness and versatility of our pipeline across diverse MHOI scenarios. We show that our method successfully refines noisy MHOI data from various sources, including existing capture methods and outputs from generative models. Furthermore, our framework shows robustness across varying numbers of participants and different types of interactions. We also present a pipeline that can create high-quality MHOI data from sparse keyframes, and a pipeline to augment MHOI data with different object geometries. 

\paragraph{Limitations} While our method demonstrates strong performance on various types of noisy MHOI data, several limitations exist. First, our approach would fail with excessively noisy input where interaction semantics cannot be reliably identified, such as completely incorrect object trajectories or body poses. Second, our method struggles with motions that exhibit extreme speed variations, as these fall outside the distribution of the single-person motion prior. Third, our current formulation cannot handle highly complex cases where contacts continuously change during the interaction (e.g., tossing an object in the air and catching it with a different hand at a new grasp location).

\paragraph{Future Work}
We believe the results of our enhancement pipeline can serve as strong kinematic priors for physics-based character control methods and humanoid robot control systems. Developing physics-based methods that enable characters to track the enhanced MHOI motions would be an interesting future direction of our work. 
Another promising direction would be retargeting our refined MHOI data to characters with different skeletal properties, such as the Unitree G1 humanoid robot, which would allow deployment of collaborative interaction behaviors across diverse robotic platforms.

\begin{acks}
This work was supported by the National Research Foundation (NRF) of Korea grant funded by the Korea government (MSIT) (RS-2024-00450647). This work was also supported by the Institute of Information \& Communications Technology Planning \& Evaluation (IITP) grant funded by MSIT under Grant Nos. RS-2025-25442338 (AI Star Fellowship Support Program, Seoul National University), RS-2021-II211343 (Artificial Intelligence Graduate School Program, Seoul National University), and IITP-2026-RS-2020-II201460 (ITRC: Information Technology Research Center).
\end{acks}

\bibliographystyle{ACM-Reference-Format}
\bibliography{reference}


\begin{thebibliography}{90}


\ifx \showCODEN    \undefined \def \showCODEN     #1{\unskip}     \fi
\ifx \showDOI      \undefined \def \showDOI       #1{#1}\fi
\ifx \showISBNx    \undefined \def \showISBNx     #1{\unskip}     \fi
\ifx \showISBNxiii \undefined \def \showISBNxiii  #1{\unskip}     \fi
\ifx \showISSN     \undefined \def \showISSN      #1{\unskip}     \fi
\ifx \showLCCN     \undefined \def \showLCCN      #1{\unskip}     \fi
\ifx \shownote     \undefined \def \shownote      #1{#1}          \fi
\ifx \showarticletitle \undefined \def \showarticletitle #1{#1}   \fi
\ifx \showURL      \undefined \def \showURL       {\relax}        \fi
\providecommand\bibfield[2]{#2}
\providecommand\bibinfo[2]{#2}
\providecommand\natexlab[1]{#1}
\providecommand\showeprint[2][]{arXiv:#2}

\bibitem[\protect\citeauthoryear{Ara{\'u}jo, Li, Vetrivel, Agarwal, Wu,
  Gopinath, Clegg, and Liu}{Ara{\'u}jo et~al\mbox{.}}{2023}]%
        {araujo2023circle}
\bibfield{author}{\bibinfo{person}{Joao~Pedro Ara{\'u}jo},
  \bibinfo{person}{Jiaman Li}, \bibinfo{person}{Karthik Vetrivel},
  \bibinfo{person}{Rishi Agarwal}, \bibinfo{person}{Jiajun Wu},
  \bibinfo{person}{Deepak Gopinath}, \bibinfo{person}{Alexander~William Clegg},
  {and} \bibinfo{person}{Karen Liu}.} \bibinfo{year}{2023}\natexlab{}.
\newblock \showarticletitle{Circle: Capture in rich contextual environments}.
  In \bibinfo{booktitle}{\emph{CVPR}}.
\newblock


\bibitem[\protect\citeauthoryear{Bae, Won, Lim, Min, and Kim}{Bae
  et~al\mbox{.}}{2023}]%
        {bae2023pmp}
\bibfield{author}{\bibinfo{person}{Jinseok Bae}, \bibinfo{person}{Jungdam Won},
  \bibinfo{person}{Donggeun Lim}, \bibinfo{person}{Cheol-Hui Min}, {and}
  \bibinfo{person}{Young~Min Kim}.} \bibinfo{year}{2023}\natexlab{}.
\newblock \showarticletitle{Pmp: Learning to physically interact with
  environments using part-wise motion priors}. In \bibinfo{booktitle}{\emph{ACM
  SIGGRAPH 2023 Conference Proceedings}}.
\newblock


\bibitem[\protect\citeauthoryear{Bensabath, Petrovich, and Varol}{Bensabath
  et~al\mbox{.}}{2024}]%
        {bensabath2024humanml3d}
\bibfield{author}{\bibinfo{person}{L{\'e}ore Bensabath},
  \bibinfo{person}{Mathis Petrovich}, {and} \bibinfo{person}{Gul Varol}.}
  \bibinfo{year}{2024}\natexlab{}.
\newblock \showarticletitle{A cross-dataset study for text-based 3D human
  motion retrieval}. In \bibinfo{booktitle}{\emph{Proceedings of the IEEE/CVF
  Conference on Computer Vision and Pattern Recognition}}.
  \bibinfo{pages}{1932--1940}.
\newblock


\bibitem[\protect\citeauthoryear{Bhatnagar, Xie, Petrov, Sminchisescu,
  Theobalt, and Pons-Moll}{Bhatnagar et~al\mbox{.}}{2022}]%
        {bhatnagar2022behave}
\bibfield{author}{\bibinfo{person}{Bharat~Lal Bhatnagar},
  \bibinfo{person}{Xianghui Xie}, \bibinfo{person}{Ilya~A Petrov},
  \bibinfo{person}{Cristian Sminchisescu}, \bibinfo{person}{Christian
  Theobalt}, {and} \bibinfo{person}{Gerard Pons-Moll}.}
  \bibinfo{year}{2022}\natexlab{}.
\newblock \showarticletitle{Behave: Dataset and method for tracking human
  object interactions}. In \bibinfo{booktitle}{\emph{CVPR}}.
\newblock


\bibitem[\protect\citeauthoryear{Blender}{Blender}{2025}]%
        {blender}
\bibfield{author}{\bibinfo{person}{Blender}.} \bibinfo{year}{2025}\natexlab{}.
\newblock \bibinfo{booktitle}{\emph{Blender - a 3D modelling and rendering
  package}}.
\newblock Blender Foundation, Stichting Blender Foundation, Amsterdam.
\newblock
\urldef\tempurl%
\url{http://www.blender.org}
\showURL{%
\tempurl}


\bibitem[\protect\citeauthoryear{Chang, Funkhouser, Guibas, Hanrahan, Huang,
  Li, Savarese, Savva, Song, Su, et~al\mbox{.}}{Chang et~al\mbox{.}}{2015}]%
        {chang2015shapenet}
\bibfield{author}{\bibinfo{person}{Angel~X Chang}, \bibinfo{person}{Thomas
  Funkhouser}, \bibinfo{person}{Leonidas Guibas}, \bibinfo{person}{Pat
  Hanrahan}, \bibinfo{person}{Qixing Huang}, \bibinfo{person}{Zimo Li},
  \bibinfo{person}{Silvio Savarese}, \bibinfo{person}{Manolis Savva},
  \bibinfo{person}{Shuran Song}, \bibinfo{person}{Hao Su}, {et~al\mbox{.}}}
  \bibinfo{year}{2015}\natexlab{}.
\newblock \showarticletitle{Shapenet: An information-rich 3d model repository}.
\newblock \bibinfo{journal}{\emph{arXiv preprint arXiv:1512.03012}}
  (\bibinfo{year}{2015}).
\newblock


\bibitem[\protect\citeauthoryear{Cohan, Tevet, Reda, Peng, and van~de
  Panne}{Cohan et~al\mbox{.}}{2024}]%
        {cohan2024flexible}
\bibfield{author}{\bibinfo{person}{Setareh Cohan}, \bibinfo{person}{Guy Tevet},
  \bibinfo{person}{Daniele Reda}, \bibinfo{person}{Xue~Bin Peng}, {and}
  \bibinfo{person}{Michiel van~de Panne}.} \bibinfo{year}{2024}\natexlab{}.
\newblock \showarticletitle{Flexible Motion In-betweening with Diffusion
  Models}.
\newblock \bibinfo{journal}{\emph{arXiv preprint arXiv:2405.11126}}
  (\bibinfo{year}{2024}).
\newblock


\bibitem[\protect\citeauthoryear{Du, Kips, Pumarola, Starke, Thabet, and
  Sanakoyeu}{Du et~al\mbox{.}}{2023}]%
        {du2023agrol}
\bibfield{author}{\bibinfo{person}{Yuming Du}, \bibinfo{person}{Robin Kips},
  \bibinfo{person}{Albert Pumarola}, \bibinfo{person}{Sebastian Starke},
  \bibinfo{person}{Ali Thabet}, {and} \bibinfo{person}{Artsiom Sanakoyeu}.}
  \bibinfo{year}{2023}\natexlab{}.
\newblock \showarticletitle{Avatars grow legs: Generating smooth human motion
  from sparse tracking inputs with diffusion model}. In
  \bibinfo{booktitle}{\emph{CVPR}}.
\newblock


\bibitem[\protect\citeauthoryear{Eom, Han, Shin, and Noh}{Eom
  et~al\mbox{.}}{2019}]%
        {eom2019mpc}
\bibfield{author}{\bibinfo{person}{Haegwang Eom}, \bibinfo{person}{Daseong
  Han}, \bibinfo{person}{Joseph~S Shin}, {and} \bibinfo{person}{Junyong Noh}.}
  \bibinfo{year}{2019}\natexlab{}.
\newblock \showarticletitle{Model predictive control with a visuomotor system
  for physics-based character animation}.
\newblock \bibinfo{journal}{\emph{ACM Trans. Graph.}} \bibinfo{volume}{39},
  \bibinfo{number}{1} (\bibinfo{year}{2019}).
\newblock


\bibitem[\protect\citeauthoryear{Gao, Wang, Xiao, Wang, Wang, Cao, Hu, Liu,
  Dai, and Pang}{Gao et~al\mbox{.}}{2024}]%
        {gao2024coohoi}
\bibfield{author}{\bibinfo{person}{Jiawei Gao}, \bibinfo{person}{Ziqin Wang},
  \bibinfo{person}{Zeqi Xiao}, \bibinfo{person}{Jingbo Wang},
  \bibinfo{person}{Tai Wang}, \bibinfo{person}{Jinkun Cao},
  \bibinfo{person}{Xiaolin Hu}, \bibinfo{person}{Si Liu},
  \bibinfo{person}{Jifeng Dai}, {and} \bibinfo{person}{Jiangmiao Pang}.}
  \bibinfo{year}{2024}\natexlab{}.
\newblock \showarticletitle{CooHOI: Learning Cooperative Human-Object
  Interaction with Manipulated Object Dynamics}. In
  \bibinfo{booktitle}{\emph{NeurIPS}}.
\newblock


\bibitem[\protect\citeauthoryear{Ghosh, Dabral, Golyanik, Theobalt, and
  Slusallek}{Ghosh et~al\mbox{.}}{2023}]%
        {ghosh2023imos}
\bibfield{author}{\bibinfo{person}{Anindita Ghosh}, \bibinfo{person}{Rishabh
  Dabral}, \bibinfo{person}{Vladislav Golyanik}, \bibinfo{person}{Christian
  Theobalt}, {and} \bibinfo{person}{Philipp Slusallek}.}
  \bibinfo{year}{2023}\natexlab{}.
\newblock \showarticletitle{IMoS: Intent-Driven Full-Body Motion Synthesis for
  Human-Object Interactions}. In \bibinfo{booktitle}{\emph{Computer Graphics
  Forum}}.
\newblock


\bibitem[\protect\citeauthoryear{Ghosh, Dabral, Golyanik, Theobalt, and
  Slusallek}{Ghosh et~al\mbox{.}}{2024}]%
        {ghosh2024remos}
\bibfield{author}{\bibinfo{person}{Anindita Ghosh}, \bibinfo{person}{Rishabh
  Dabral}, \bibinfo{person}{Vladislav Golyanik}, \bibinfo{person}{Christian
  Theobalt}, {and} \bibinfo{person}{Philipp Slusallek}.}
  \bibinfo{year}{2024}\natexlab{}.
\newblock \showarticletitle{Remos: 3d motion-conditioned reaction synthesis for
  two-person interactions}. In \bibinfo{booktitle}{\emph{European Conference on
  Computer Vision}}.
\newblock


\bibitem[\protect\citeauthoryear{Grady, Tang, Twigg, Vo, Brahmbhatt, and
  Kemp}{Grady et~al\mbox{.}}{2021}]%
        {grady2021contactopt}
\bibfield{author}{\bibinfo{person}{Patrick Grady}, \bibinfo{person}{Chengcheng
  Tang}, \bibinfo{person}{Christopher~D Twigg}, \bibinfo{person}{Minh Vo},
  \bibinfo{person}{Samarth Brahmbhatt}, {and} \bibinfo{person}{Charles~C
  Kemp}.} \bibinfo{year}{2021}\natexlab{}.
\newblock \showarticletitle{Contactopt: Optimizing contact to improve grasps}.
  In \bibinfo{booktitle}{\emph{CVPR}}.
\newblock


\bibitem[\protect\citeauthoryear{Hassan, Ceylan, Villegas, Saito, Yang, Zhou,
  and Black}{Hassan et~al\mbox{.}}{2021}]%
        {hassan2021samp}
\bibfield{author}{\bibinfo{person}{Mohamed Hassan}, \bibinfo{person}{Duygu
  Ceylan}, \bibinfo{person}{Ruben Villegas}, \bibinfo{person}{Jun Saito},
  \bibinfo{person}{Jimei Yang}, \bibinfo{person}{Yi Zhou}, {and}
  \bibinfo{person}{Michael Black}.} \bibinfo{year}{2021}\natexlab{}.
\newblock \showarticletitle{Stochastic Scene-Aware Motion Prediction}. In
  \bibinfo{booktitle}{\emph{ICCV}}.
\newblock


\bibitem[\protect\citeauthoryear{Hassan, Guo, Wang, Black, Fidler, and
  Peng}{Hassan et~al\mbox{.}}{2023}]%
        {hassan2023synthesizing}
\bibfield{author}{\bibinfo{person}{Mohamed Hassan}, \bibinfo{person}{Yunrong
  Guo}, \bibinfo{person}{Tingwu Wang}, \bibinfo{person}{Michael Black},
  \bibinfo{person}{Sanja Fidler}, {and} \bibinfo{person}{Xue~Bin Peng}.}
  \bibinfo{year}{2023}\natexlab{}.
\newblock \showarticletitle{Synthesizing physical character-scene
  interactions}. In \bibinfo{booktitle}{\emph{ACM SIGGRAPH 2023 Conference
  Proceedings}}.
\newblock


\bibitem[\protect\citeauthoryear{He, Saito, Zachary, Rushmeier, and Zhou}{He
  et~al\mbox{.}}{2022}]%
        {he2022nemf}
\bibfield{author}{\bibinfo{person}{Chengan He}, \bibinfo{person}{Jun Saito},
  \bibinfo{person}{James Zachary}, \bibinfo{person}{Holly Rushmeier}, {and}
  \bibinfo{person}{Yi Zhou}.} \bibinfo{year}{2022}\natexlab{}.
\newblock \showarticletitle{Nemf: Neural motion fields for kinematic
  animation}.
\newblock \bibinfo{journal}{\emph{NeurIPS}} (\bibinfo{year}{2022}).
\newblock


\bibitem[\protect\citeauthoryear{Ho, Komura, and Tai}{Ho et~al\mbox{.}}{2010}]%
        {ho2010spatial}
\bibfield{author}{\bibinfo{person}{Edmond~SL Ho}, \bibinfo{person}{Taku
  Komura}, {and} \bibinfo{person}{Chiew-Lan Tai}.}
  \bibinfo{year}{2010}\natexlab{}.
\newblock \showarticletitle{Spatial relationship preserving character motion
  adaptation}.
\newblock In \bibinfo{booktitle}{\emph{ACM SIGGRAPH 2010 papers}}.
\newblock


\bibitem[\protect\citeauthoryear{Ho, Jain, and Abbeel}{Ho
  et~al\mbox{.}}{2020}]%
        {ho2020denoising}
\bibfield{author}{\bibinfo{person}{Jonathan Ho}, \bibinfo{person}{Ajay Jain},
  {and} \bibinfo{person}{Pieter Abbeel}.} \bibinfo{year}{2020}\natexlab{}.
\newblock \showarticletitle{Denoising diffusion probabilistic models}.
\newblock \bibinfo{journal}{\emph{NeurIPS}} (\bibinfo{year}{2020}).
\newblock


\bibitem[\protect\citeauthoryear{Holden}{Holden}{2018}]%
        {holden2018robust}
\bibfield{author}{\bibinfo{person}{Daniel Holden}.}
  \bibinfo{year}{2018}\natexlab{}.
\newblock \showarticletitle{Robust solving of optical motion capture data by
  denoising}.
\newblock \bibinfo{journal}{\emph{ACM Transactions on Graphics (TOG)}}
  (\bibinfo{year}{2018}).
\newblock


\bibitem[\protect\citeauthoryear{Holden, Komura, and Saito}{Holden
  et~al\mbox{.}}{2017}]%
        {holden2017pfnn}
\bibfield{author}{\bibinfo{person}{Daniel Holden}, \bibinfo{person}{Taku
  Komura}, {and} \bibinfo{person}{Jun Saito}.} \bibinfo{year}{2017}\natexlab{}.
\newblock \showarticletitle{Phase-functioned neural networks for character
  control}.
\newblock \bibinfo{journal}{\emph{ACM Transactions on Graphics (TOG)}}
  (\bibinfo{year}{2017}).
\newblock


\bibitem[\protect\citeauthoryear{Jiang, He, Wang, Li, Chen, Huang, and
  Zhu}{Jiang et~al\mbox{.}}{2024a}]%
        {jiang2024lingo}
\bibfield{author}{\bibinfo{person}{Nan Jiang}, \bibinfo{person}{Zimo He},
  \bibinfo{person}{Zi Wang}, \bibinfo{person}{Hongjie Li},
  \bibinfo{person}{Yixin Chen}, \bibinfo{person}{Siyuan Huang}, {and}
  \bibinfo{person}{Yixin Zhu}.} \bibinfo{year}{2024}\natexlab{a}.
\newblock \showarticletitle{Autonomous character-scene interaction synthesis
  from text instruction}. In \bibinfo{booktitle}{\emph{ACM SIGGRAPH Asia 2024
  Conference Proceedings}}.
\newblock


\bibitem[\protect\citeauthoryear{Jiang, Zhang, Li, Ma, Wang, Chen, Liu, Zhu,
  and Huang}{Jiang et~al\mbox{.}}{2024b}]%
        {jiang2024trumans}
\bibfield{author}{\bibinfo{person}{Nan Jiang}, \bibinfo{person}{Zhiyuan Zhang},
  \bibinfo{person}{Hongjie Li}, \bibinfo{person}{Xiaoxuan Ma},
  \bibinfo{person}{Zan Wang}, \bibinfo{person}{Yixin Chen},
  \bibinfo{person}{Tengyu Liu}, \bibinfo{person}{Yixin Zhu}, {and}
  \bibinfo{person}{Siyuan Huang}.} \bibinfo{year}{2024}\natexlab{b}.
\newblock \showarticletitle{Scaling up dynamic human-scene interaction
  modeling}. In \bibinfo{booktitle}{\emph{CVPR}}.
\newblock


\bibitem[\protect\citeauthoryear{Karunratanakul, Preechakul, Aksan, Beeler,
  Suwajanakorn, and Tang}{Karunratanakul et~al\mbox{.}}{2024}]%
        {karunratanakul2024dno}
\bibfield{author}{\bibinfo{person}{Korrawe Karunratanakul},
  \bibinfo{person}{Konpat Preechakul}, \bibinfo{person}{Emre Aksan},
  \bibinfo{person}{Thabo Beeler}, \bibinfo{person}{Supasorn Suwajanakorn},
  {and} \bibinfo{person}{Siyu Tang}.} \bibinfo{year}{2024}\natexlab{}.
\newblock \showarticletitle{Optimizing diffusion noise can serve as universal
  motion priors}. In \bibinfo{booktitle}{\emph{CVPR}}.
\newblock


\bibitem[\protect\citeauthoryear{Karunratanakul, Preechakul, Suwajanakorn, and
  Tang}{Karunratanakul et~al\mbox{.}}{2023}]%
        {karunratanakul2023gmd}
\bibfield{author}{\bibinfo{person}{Korrawe Karunratanakul},
  \bibinfo{person}{Konpat Preechakul}, \bibinfo{person}{Supasorn Suwajanakorn},
  {and} \bibinfo{person}{Siyu Tang}.} \bibinfo{year}{2023}\natexlab{}.
\newblock \showarticletitle{Guided motion diffusion for controllable human
  motion synthesis}. In \bibinfo{booktitle}{\emph{ICCV}}.
\newblock


\bibitem[\protect\citeauthoryear{Kim, Hwang, Hyun, and Lee}{Kim
  et~al\mbox{.}}{2012}]%
        {kim2012tiling}
\bibfield{author}{\bibinfo{person}{Manmyung Kim}, \bibinfo{person}{Youngseok
  Hwang}, \bibinfo{person}{Kyunglyul Hyun}, {and} \bibinfo{person}{Jehee Lee}.}
  \bibinfo{year}{2012}\natexlab{}.
\newblock \showarticletitle{Tiling motion patches}. In
  \bibinfo{booktitle}{\emph{Proceedings of the ACM SIGGRAPH/Eurographics
  Symposium on Computer Animation}}.
\newblock


\bibitem[\protect\citeauthoryear{Lee and Joo}{Lee and Joo}{2023}]%
        {lee2023lama}
\bibfield{author}{\bibinfo{person}{Jiye Lee} {and} \bibinfo{person}{Hanbyul
  Joo}.} \bibinfo{year}{2023}\natexlab{}.
\newblock \showarticletitle{Locomotion-Action-Manipulation: Synthesizing
  Human-Scene Interactions in Complex 3D Environments}. In
  \bibinfo{booktitle}{\emph{ICCV}}.
\newblock


\bibitem[\protect\citeauthoryear{Lee, Choi, and Lee}{Lee et~al\mbox{.}}{2006}]%
        {lee2006motionpatch}
\bibfield{author}{\bibinfo{person}{Kang~Hoon Lee}, \bibinfo{person}{Myung~Geol
  Choi}, {and} \bibinfo{person}{Jehee Lee}.} \bibinfo{year}{2006}\natexlab{}.
\newblock \showarticletitle{Motion patches: building blocks for virtual
  environments annotated with motion data}.
\newblock In \bibinfo{booktitle}{\emph{ACM SIGGRAPH 2006 Papers}}.
\newblock


\bibitem[\protect\citeauthoryear{Li, Cao, Zhang, Rempe, Kautz, Iqbal, and
  Yuan}{Li et~al\mbox{.}}{2025}]%
        {li2025genmo}
\bibfield{author}{\bibinfo{person}{Jiefeng Li}, \bibinfo{person}{Jinkun Cao},
  \bibinfo{person}{Haotian Zhang}, \bibinfo{person}{Davis Rempe},
  \bibinfo{person}{Jan Kautz}, \bibinfo{person}{Umar Iqbal}, {and}
  \bibinfo{person}{Ye Yuan}.} \bibinfo{year}{2025}\natexlab{}.
\newblock \showarticletitle{GENMO: A GENeralist Model for Human MOtion}.
\newblock \bibinfo{journal}{\emph{arXiv preprint arXiv:2505.01425}}
  (\bibinfo{year}{2025}).
\newblock


\bibitem[\protect\citeauthoryear{Li, Clegg, Mottaghi, Wu, Puig, and Liu}{Li
  et~al\mbox{.}}{2024}]%
        {li2023chois}
\bibfield{author}{\bibinfo{person}{Jiaman Li}, \bibinfo{person}{Alexander
  Clegg}, \bibinfo{person}{Roozbeh Mottaghi}, \bibinfo{person}{Jiajun Wu},
  \bibinfo{person}{Xavier Puig}, {and} \bibinfo{person}{C.~Karen Liu}.}
  \bibinfo{year}{2024}\natexlab{}.
\newblock \showarticletitle{Controllable human-object interaction synthesis}.
  In \bibinfo{booktitle}{\emph{ECCV}}.
\newblock


\bibitem[\protect\citeauthoryear{Li, Wu, and Liu}{Li et~al\mbox{.}}{2023}]%
        {li2023omomo}
\bibfield{author}{\bibinfo{person}{Jiaman Li}, \bibinfo{person}{Jiajun Wu},
  {and} \bibinfo{person}{C~Karen Liu}.} \bibinfo{year}{2023}\natexlab{}.
\newblock \showarticletitle{Object Motion Guided Human Motion Synthesis}.
\newblock \bibinfo{journal}{\emph{ACM Trans. Graph.}} \bibinfo{volume}{42},
  \bibinfo{number}{6} (\bibinfo{year}{2023}).
\newblock


\bibitem[\protect\citeauthoryear{Li, Fu, and Pollard}{Li et~al\mbox{.}}{2007}]%
        {li2007datagrasp}
\bibfield{author}{\bibinfo{person}{Ying Li}, \bibinfo{person}{Jiaxin~L Fu},
  {and} \bibinfo{person}{Nancy~S Pollard}.} \bibinfo{year}{2007}\natexlab{}.
\newblock \showarticletitle{Data-driven grasp synthesis using shape matching
  and task-based pruning}.
\newblock \bibinfo{journal}{\emph{IEEE Transactions on visualization and
  computer graphics}} \bibinfo{volume}{13}, \bibinfo{number}{4}
  (\bibinfo{year}{2007}), \bibinfo{pages}{732--747}.
\newblock


\bibitem[\protect\citeauthoryear{Liang, Zhang, Li, Yu, and Xu}{Liang
  et~al\mbox{.}}{2024}]%
        {liang2024intergen}
\bibfield{author}{\bibinfo{person}{Han Liang}, \bibinfo{person}{Wenqian Zhang},
  \bibinfo{person}{Wenxuan Li}, \bibinfo{person}{Jingyi Yu}, {and}
  \bibinfo{person}{Lan Xu}.} \bibinfo{year}{2024}\natexlab{}.
\newblock \showarticletitle{Intergen: Diffusion-based multi-human motion
  generation under complex interactions}.
\newblock \bibinfo{journal}{\emph{IJCV}} (\bibinfo{year}{2024}).
\newblock


\bibitem[\protect\citeauthoryear{Ling, Zinno, Cheng, and Van De~Panne}{Ling
  et~al\mbox{.}}{2020}]%
        {ling2020motionvae}
\bibfield{author}{\bibinfo{person}{Hung~Yu Ling}, \bibinfo{person}{Fabio
  Zinno}, \bibinfo{person}{George Cheng}, {and} \bibinfo{person}{Michiel Van
  De~Panne}.} \bibinfo{year}{2020}\natexlab{}.
\newblock \showarticletitle{Character controllers using motion vaes}.
\newblock \bibinfo{journal}{\emph{ACM Transactions on Graphics (TOG)}}
  (\bibinfo{year}{2020}).
\newblock


\bibitem[\protect\citeauthoryear{Lionar and Lee}{Lionar and Lee}{2026}]%
        {lionar2026teamhoi}
\bibfield{author}{\bibinfo{person}{Stefan Lionar} {and}
  \bibinfo{person}{Gim~Hee Lee}.} \bibinfo{year}{2026}\natexlab{}.
\newblock \showarticletitle{TeamHOI: Learning a Unified Policy for Cooperative
  Human-Object Interactions with Any Team Size}.
\newblock \bibinfo{journal}{\emph{arXiv preprint arXiv:2603.07988}}
  (\bibinfo{year}{2026}).
\newblock


\bibitem[\protect\citeauthoryear{Liu}{Liu}{2009}]%
        {liu2009dextrous}
\bibfield{author}{\bibinfo{person}{C~Karen Liu}.}
  \bibinfo{year}{2009}\natexlab{}.
\newblock \showarticletitle{Dextrous manipulation from a grasping pose}.
\newblock In \bibinfo{booktitle}{\emph{ACM SIGGRAPH 2009 papers}}.
\newblock


\bibitem[\protect\citeauthoryear{Liu, Hertzmann, and Popovi{\'c}}{Liu
  et~al\mbox{.}}{2006}]%
        {liu2006composition}
\bibfield{author}{\bibinfo{person}{C~Karen Liu}, \bibinfo{person}{Aaron
  Hertzmann}, {and} \bibinfo{person}{Zoran Popovi{\'c}}.}
  \bibinfo{year}{2006}\natexlab{}.
\newblock \showarticletitle{Composition of complex optimal multi-character
  motions}. In \bibinfo{booktitle}{\emph{Proceedings of the 2006 ACM
  SIGGRAPH/Eurographics symposium on Computer animation}}.
\newblock


\bibitem[\protect\citeauthoryear{Liu, Liang, Wang, Du, Zhang, and Li}{Liu
  et~al\mbox{.}}{2025}]%
        {liu2025uni}
\bibfield{author}{\bibinfo{person}{Sheng Liu}, \bibinfo{person}{Yuanzhi Liang},
  \bibinfo{person}{Jiepeng Wang}, \bibinfo{person}{Sidan Du},
  \bibinfo{person}{Chi Zhang}, {and} \bibinfo{person}{Xuelong Li}.}
  \bibinfo{year}{2025}\natexlab{}.
\newblock \showarticletitle{Uni-Inter: Unifying 3D Human Motion Synthesis
  Across Diverse Interaction Contexts}. In
  \bibinfo{booktitle}{\emph{Proceedings of the SIGGRAPH Asia 2025 Conference
  Papers}}.
\newblock


\bibitem[\protect\citeauthoryear{Liu, Liu, Jiao, Zhu, and Zhu}{Liu
  et~al\mbox{.}}{2021}]%
        {liu202fc}
\bibfield{author}{\bibinfo{person}{Tengyu Liu}, \bibinfo{person}{Zeyu Liu},
  \bibinfo{person}{Ziyuan Jiao}, \bibinfo{person}{Yixin Zhu}, {and}
  \bibinfo{person}{Song-Chun Zhu}.} \bibinfo{year}{2021}\natexlab{}.
\newblock \showarticletitle{Synthesizing diverse and physically stable grasps
  with arbitrary hand structures using differentiable force closure estimator}.
\newblock \bibinfo{journal}{\emph{IEEE Robotics and Automation Letters}}
  (\bibinfo{year}{2021}).
\newblock


\bibitem[\protect\citeauthoryear{Liu and Yi}{Liu and Yi}{2024}]%
        {liu2024geneoh}
\bibfield{author}{\bibinfo{person}{Xueyi Liu} {and} \bibinfo{person}{Li Yi}.}
  \bibinfo{year}{2024}\natexlab{}.
\newblock \showarticletitle{Geneoh diffusion: Towards generalizable hand-object
  interaction denoising via denoising diffusion}.
\newblock \bibinfo{journal}{\emph{arXiv preprint arXiv:2402.14810}}
  (\bibinfo{year}{2024}).
\newblock


\bibitem[\protect\citeauthoryear{Liu, Zhang, Xing, Tang, Yang, and Yi}{Liu
  et~al\mbox{.}}{2024}]%
        {liu2024core4d}
\bibfield{author}{\bibinfo{person}{Yun Liu}, \bibinfo{person}{Chengwen Zhang},
  \bibinfo{person}{Ruofan Xing}, \bibinfo{person}{Bingda Tang},
  \bibinfo{person}{Bowen Yang}, {and} \bibinfo{person}{Li Yi}.}
  \bibinfo{year}{2024}\natexlab{}.
\newblock \showarticletitle{Core4d: A 4d human-object-human interaction dataset
  for collaborative object rearrangement}.
\newblock \bibinfo{journal}{\emph{arXiv preprint arXiv:2406.19353}}
  (\bibinfo{year}{2024}).
\newblock


\bibitem[\protect\citeauthoryear{Lu, Zhang, Ye, Shiratori, Starke, and
  Komura}{Lu et~al\mbox{.}}{2025}]%
        {lu2025choice}
\bibfield{author}{\bibinfo{person}{Jintao Lu}, \bibinfo{person}{He Zhang},
  \bibinfo{person}{Yuting Ye}, \bibinfo{person}{Takaaki Shiratori},
  \bibinfo{person}{Sebastian Starke}, {and} \bibinfo{person}{Taku Komura}.}
  \bibinfo{year}{2025}\natexlab{}.
\newblock \showarticletitle{CHOICE: Coordinated human-object interaction in
  cluttered environments for pick-and-place actions}.
\newblock \bibinfo{journal}{\emph{ACM Transactions on Graphics}}
  (\bibinfo{year}{2025}).
\newblock


\bibitem[\protect\citeauthoryear{{Luma AI}}{{Luma AI}}{2025}]%
        {lumaai}
\bibfield{author}{\bibinfo{person}{{Luma AI}}.}
  \bibinfo{year}{2025}\natexlab{}.
\newblock \bibinfo{title}{Luma AI: Genie}.
\newblock \bibinfo{howpublished}{\url{https://lumalabs.ai/}}.
\newblock


\bibitem[\protect\citeauthoryear{Luo, Cao, Christen, Winkler, Kitani, and
  Xu}{Luo et~al\mbox{.}}{2024}]%
        {luo2024omnigrasp}
\bibfield{author}{\bibinfo{person}{Zhengyi Luo}, \bibinfo{person}{Jinkun Cao},
  \bibinfo{person}{Sammy Christen}, \bibinfo{person}{Alexander Winkler},
  \bibinfo{person}{Kris Kitani}, {and} \bibinfo{person}{Weipeng Xu}.}
  \bibinfo{year}{2024}\natexlab{}.
\newblock \showarticletitle{Omnigrasp: Grasping diverse objects with simulated
  humanoids}.
\newblock \bibinfo{journal}{\emph{NeurIPS}} (\bibinfo{year}{2024}).
\newblock


\bibitem[\protect\citeauthoryear{Merel, Tunyasuvunakool, Ahuja, Tassa,
  Hasenclever, Pham, Erez, Wayne, and Heess}{Merel et~al\mbox{.}}{2020}]%
        {merel2020catch}
\bibfield{author}{\bibinfo{person}{Josh Merel}, \bibinfo{person}{Saran
  Tunyasuvunakool}, \bibinfo{person}{Arun Ahuja}, \bibinfo{person}{Yuval
  Tassa}, \bibinfo{person}{Leonard Hasenclever}, \bibinfo{person}{Vu Pham},
  \bibinfo{person}{Tom Erez}, \bibinfo{person}{Greg Wayne}, {and}
  \bibinfo{person}{Nicolas Heess}.} \bibinfo{year}{2020}\natexlab{}.
\newblock \showarticletitle{Catch \& Carry: reusable neural controllers for
  vision-guided whole-body tasks}.
\newblock \bibinfo{journal}{\emph{ACM Trans. Graph.}} \bibinfo{volume}{39},
  \bibinfo{number}{4} (\bibinfo{year}{2020}).
\newblock


\bibitem[\protect\citeauthoryear{Pan, Yang, Dou, Wang, Huang, Dai, Komura, and
  Wang}{Pan et~al\mbox{.}}{2025}]%
        {pan2025tokenhsi}
\bibfield{author}{\bibinfo{person}{Liang Pan}, \bibinfo{person}{Zeshi Yang},
  \bibinfo{person}{Zhiyang Dou}, \bibinfo{person}{Wenjia Wang},
  \bibinfo{person}{Buzhen Huang}, \bibinfo{person}{Bo Dai},
  \bibinfo{person}{Taku Komura}, {and} \bibinfo{person}{Jingbo Wang}.}
  \bibinfo{year}{2025}\natexlab{}.
\newblock \showarticletitle{TokenHSI: Unified Synthesis of Physical Human-Scene
  Interactions through Task Tokenization}.
\newblock \bibinfo{journal}{\emph{arXiv preprint arXiv:2503.19901}}
  (\bibinfo{year}{2025}).
\newblock


\bibitem[\protect\citeauthoryear{Pavlakos, Choutas, Ghorbani, Bolkart, Osman,
  Tzionas, and Black}{Pavlakos et~al\mbox{.}}{2019}]%
        {pavlakos2019smplx}
\bibfield{author}{\bibinfo{person}{Georgios Pavlakos},
  \bibinfo{person}{Vasileios Choutas}, \bibinfo{person}{Nima Ghorbani},
  \bibinfo{person}{Timo Bolkart}, \bibinfo{person}{Ahmed A.~A. Osman},
  \bibinfo{person}{Dimitrios Tzionas}, {and} \bibinfo{person}{Michael~J.
  Black}.} \bibinfo{year}{2019}\natexlab{}.
\newblock \showarticletitle{Expressive Body Capture: {3D} Hands, Face, and Body
  from a Single Image}. In \bibinfo{booktitle}{\emph{CVPR}}.
\newblock


\bibitem[\protect\citeauthoryear{Pollard and Zordan}{Pollard and
  Zordan}{2005}]%
        {pollard2005physically}
\bibfield{author}{\bibinfo{person}{Nancy~S Pollard} {and}
  \bibinfo{person}{Victor~Brian Zordan}.} \bibinfo{year}{2005}\natexlab{}.
\newblock \showarticletitle{Physically based grasping control from example}. In
  \bibinfo{booktitle}{\emph{Proceedings of the 2005 ACM SIGGRAPH/Eurographics
  symposium on Computer animation}}.
\newblock


\bibitem[\protect\citeauthoryear{Rempe, Birdal, Hertzmann, Yang, Sridhar, and
  Guibas}{Rempe et~al\mbox{.}}{2021}]%
        {rempe2021humor}
\bibfield{author}{\bibinfo{person}{Davis Rempe}, \bibinfo{person}{Tolga
  Birdal}, \bibinfo{person}{Aaron Hertzmann}, \bibinfo{person}{Jimei Yang},
  \bibinfo{person}{Srinath Sridhar}, {and} \bibinfo{person}{Leonidas~J
  Guibas}.} \bibinfo{year}{2021}\natexlab{}.
\newblock \showarticletitle{Humor: 3d human motion model for robust pose
  estimation}. In \bibinfo{booktitle}{\emph{ICCV}}.
\newblock


\bibitem[\protect\citeauthoryear{Ron, Tevet, Sawdayee, and Bermano}{Ron
  et~al\mbox{.}}{2025}]%
        {ron2025hoidini}
\bibfield{author}{\bibinfo{person}{Roey Ron}, \bibinfo{person}{Guy Tevet},
  \bibinfo{person}{Haim Sawdayee}, {and} \bibinfo{person}{Amit~H Bermano}.}
  \bibinfo{year}{2025}\natexlab{}.
\newblock \showarticletitle{Hoidini: Human-object interaction through diffusion
  noise optimization}.
\newblock \bibinfo{journal}{\emph{arXiv preprint arXiv:2506.15625}}
  (\bibinfo{year}{2025}).
\newblock


\bibitem[\protect\citeauthoryear{Shafir, Tevet, Kapon, and Bermano}{Shafir
  et~al\mbox{.}}{2023}]%
        {shafir2023priormdm}
\bibfield{author}{\bibinfo{person}{Yonatan Shafir}, \bibinfo{person}{Guy
  Tevet}, \bibinfo{person}{Roy Kapon}, {and} \bibinfo{person}{Amit~H Bermano}.}
  \bibinfo{year}{2023}\natexlab{}.
\newblock \showarticletitle{Human motion diffusion as a generative prior}.
\newblock \bibinfo{journal}{\emph{arXiv preprint arXiv:2303.01418}}
  (\bibinfo{year}{2023}).
\newblock


\bibitem[\protect\citeauthoryear{Shi, Starke, Ye, Komura, and Won}{Shi
  et~al\mbox{.}}{2023}]%
        {shi2023phasemp}
\bibfield{author}{\bibinfo{person}{Mingyi Shi}, \bibinfo{person}{Sebastian
  Starke}, \bibinfo{person}{Yuting Ye}, \bibinfo{person}{Taku Komura}, {and}
  \bibinfo{person}{Jungdam Won}.} \bibinfo{year}{2023}\natexlab{}.
\newblock \showarticletitle{PhaseMP: Robust 3D Pose Estimation via
  Phase-conditioned Human Motion Prior}. In \bibinfo{booktitle}{\emph{ICCV}}.
\newblock


\bibitem[\protect\citeauthoryear{Shum, Komura, Shiraishi, and Yamazaki}{Shum
  et~al\mbox{.}}{2008b}]%
        {shum2008interactionpatch}
\bibfield{author}{\bibinfo{person}{Hubert~PH Shum}, \bibinfo{person}{Taku
  Komura}, \bibinfo{person}{Masashi Shiraishi}, {and} \bibinfo{person}{Shuntaro
  Yamazaki}.} \bibinfo{year}{2008}\natexlab{b}.
\newblock \showarticletitle{Interaction patches for multi-character animation}.
\newblock \bibinfo{journal}{\emph{ACM Transactions on Graphics (TOG)}}
  (\bibinfo{year}{2008}).
\newblock


\bibitem[\protect\citeauthoryear{Shum, Komura, and Yamazaki}{Shum
  et~al\mbox{.}}{2007}]%
        {shum2007simulating}
\bibfield{author}{\bibinfo{person}{Hubert~PH Shum}, \bibinfo{person}{Taku
  Komura}, {and} \bibinfo{person}{Shuntaro Yamazaki}.}
  \bibinfo{year}{2007}\natexlab{}.
\newblock \showarticletitle{Simulating competitive interactions using singly
  captured motions}. In \bibinfo{booktitle}{\emph{Proceedings of the 2007 ACM
  symposium on Virtual reality software and technology}}.
\newblock


\bibitem[\protect\citeauthoryear{Shum, Komura, and Yamazaki}{Shum
  et~al\mbox{.}}{2008a}]%
        {shum2008simulatingspace}
\bibfield{author}{\bibinfo{person}{Hubert~PH Shum}, \bibinfo{person}{Taku
  Komura}, {and} \bibinfo{person}{Shuntaro Yamazaki}.}
  \bibinfo{year}{2008}\natexlab{a}.
\newblock \showarticletitle{Simulating interactions of avatars in high
  dimensional state space}. In \bibinfo{booktitle}{\emph{Proceedings of the
  2008 Symposium on interactive 3D Graphics and Games}}.
\newblock


\bibitem[\protect\citeauthoryear{Song, Meng, and Ermon}{Song
  et~al\mbox{.}}{2020}]%
        {song2020ddim}
\bibfield{author}{\bibinfo{person}{Jiaming Song}, \bibinfo{person}{Chenlin
  Meng}, {and} \bibinfo{person}{Stefano Ermon}.}
  \bibinfo{year}{2020}\natexlab{}.
\newblock \showarticletitle{Denoising diffusion implicit models}.
\newblock \bibinfo{journal}{\emph{arXiv preprint arXiv:2010.02502}}
  (\bibinfo{year}{2020}).
\newblock


\bibitem[\protect\citeauthoryear{Starke, Mason, and Komura}{Starke
  et~al\mbox{.}}{2022}]%
        {starke2022deepphase}
\bibfield{author}{\bibinfo{person}{Sebastian Starke}, \bibinfo{person}{Ian
  Mason}, {and} \bibinfo{person}{Taku Komura}.}
  \bibinfo{year}{2022}\natexlab{}.
\newblock \showarticletitle{DeepPhase: periodic autoencoders for learning
  motion phase manifolds}.
\newblock \bibinfo{journal}{\emph{ACM Trans. Graph.}} (\bibinfo{year}{2022}).
\newblock


\bibitem[\protect\citeauthoryear{Starke, Starke, He, Komura, and Ye}{Starke
  et~al\mbox{.}}{2024}]%
        {starke2024categorical}
\bibfield{author}{\bibinfo{person}{Sebastian Starke}, \bibinfo{person}{Paul
  Starke}, \bibinfo{person}{Nicky He}, \bibinfo{person}{Taku Komura}, {and}
  \bibinfo{person}{Yuting Ye}.} \bibinfo{year}{2024}\natexlab{}.
\newblock \showarticletitle{Categorical codebook matching for embodied
  character controllers}.
\newblock \bibinfo{journal}{\emph{ACM Transactions on Graphics (TOG)}}
  (\bibinfo{year}{2024}).
\newblock


\bibitem[\protect\citeauthoryear{Starke, Zhang, Komura, and Saito}{Starke
  et~al\mbox{.}}{2019}]%
        {starke2019neural}
\bibfield{author}{\bibinfo{person}{Sebastian Starke}, \bibinfo{person}{He
  Zhang}, \bibinfo{person}{Taku Komura}, {and} \bibinfo{person}{Jun Saito}.}
  \bibinfo{year}{2019}\natexlab{}.
\newblock \showarticletitle{Neural state machine for character-scene
  interactions.}
\newblock \bibinfo{journal}{\emph{ACM Trans. Graph.}} \bibinfo{volume}{38},
  \bibinfo{number}{6} (\bibinfo{year}{2019}).
\newblock


\bibitem[\protect\citeauthoryear{Starke, Zhao, Zinno, and Komura}{Starke
  et~al\mbox{.}}{2021}]%
        {starke2021neural}
\bibfield{author}{\bibinfo{person}{Sebastian Starke}, \bibinfo{person}{Yiwei
  Zhao}, \bibinfo{person}{Fabio Zinno}, {and} \bibinfo{person}{Taku Komura}.}
  \bibinfo{year}{2021}\natexlab{}.
\newblock \showarticletitle{Neural animation layering for synthesizing martial
  arts movements}.
\newblock \bibinfo{journal}{\emph{ACM Transactions on Graphics (TOG)}}
  (\bibinfo{year}{2021}).
\newblock


\bibitem[\protect\citeauthoryear{Taheri, Choutas, Black, and Tzionas}{Taheri
  et~al\mbox{.}}{2022}]%
        {taheri2022goal}
\bibfield{author}{\bibinfo{person}{Omid Taheri}, \bibinfo{person}{Vasileios
  Choutas}, \bibinfo{person}{Michael~J Black}, {and} \bibinfo{person}{Dimitrios
  Tzionas}.} \bibinfo{year}{2022}\natexlab{}.
\newblock \showarticletitle{Goal: Generating 4d whole-body motion for
  hand-object grasping}. In \bibinfo{booktitle}{\emph{CVPR}}.
\newblock


\bibitem[\protect\citeauthoryear{Taheri, Ghorbani, Black, and Tzionas}{Taheri
  et~al\mbox{.}}{2020}]%
        {taheri2020grab}
\bibfield{author}{\bibinfo{person}{Omid Taheri}, \bibinfo{person}{Nima
  Ghorbani}, \bibinfo{person}{Michael~J Black}, {and}
  \bibinfo{person}{Dimitrios Tzionas}.} \bibinfo{year}{2020}\natexlab{}.
\newblock \showarticletitle{GRAB: A dataset of whole-body human grasping of
  objects}. In \bibinfo{booktitle}{\emph{ECCV}}.
\newblock


\bibitem[\protect\citeauthoryear{Taheri, Zhou, Tzionas, Zhou, Ceylan, Pirk, and
  Black}{Taheri et~al\mbox{.}}{2023}]%
        {taheri2023grip}
\bibfield{author}{\bibinfo{person}{Omid Taheri}, \bibinfo{person}{Yi Zhou},
  \bibinfo{person}{Dimitrios Tzionas}, \bibinfo{person}{Yang Zhou},
  \bibinfo{person}{Duygu Ceylan}, \bibinfo{person}{Soren Pirk}, {and}
  \bibinfo{person}{Michael~J Black}.} \bibinfo{year}{2023}\natexlab{}.
\newblock \showarticletitle{GRIP: Generating Interaction Poses Using Spatial
  Cues and Latent Consistency}.
\newblock \bibinfo{journal}{\emph{arXiv preprint arXiv:2308.11617}}
  (\bibinfo{year}{2023}).
\newblock


\bibitem[\protect\citeauthoryear{Tessler, Jiang, Coumans, Luo, Peng, and
  Chechik}{Tessler et~al\mbox{.}}{2025}]%
        {tessler2025maskedmanipulator}
\bibfield{author}{\bibinfo{person}{Chen Tessler}, \bibinfo{person}{Yifeng
  Jiang}, \bibinfo{person}{Erwin Coumans}, \bibinfo{person}{Zhengyi Luo},
  \bibinfo{person}{Xue~Bin Peng}, {and} \bibinfo{person}{Gal Chechik}.}
  \bibinfo{year}{2025}\natexlab{}.
\newblock \showarticletitle{Maskedmanipulator: Versatile whole-body control for
  loco-manipulation}. In \bibinfo{booktitle}{\emph{Proceedings of the SIGGRAPH
  Asia 2025 Conference Papers}}.
\newblock


\bibitem[\protect\citeauthoryear{Tevet, Raab, Gordon, Shafir, Cohen-Or, and
  Bermano}{Tevet et~al\mbox{.}}{2022}]%
        {tevet2022mdm}
\bibfield{author}{\bibinfo{person}{Guy Tevet}, \bibinfo{person}{Sigal Raab},
  \bibinfo{person}{Brian Gordon}, \bibinfo{person}{Yonatan Shafir},
  \bibinfo{person}{Daniel Cohen-Or}, {and} \bibinfo{person}{Amit~H Bermano}.}
  \bibinfo{year}{2022}\natexlab{}.
\newblock \showarticletitle{Human motion diffusion model}.
\newblock \bibinfo{journal}{\emph{arXiv preprint arXiv:2209.14916}}
  (\bibinfo{year}{2022}).
\newblock


\bibitem[\protect\citeauthoryear{Wang, Rong, Liu, Yan, Lin, and Dai}{Wang
  et~al\mbox{.}}{2022b}]%
        {wang2022towards}
\bibfield{author}{\bibinfo{person}{Jingbo Wang}, \bibinfo{person}{Yu Rong},
  \bibinfo{person}{Jingyuan Liu}, \bibinfo{person}{Sijie Yan},
  \bibinfo{person}{Dahua Lin}, {and} \bibinfo{person}{Bo Dai}.}
  \bibinfo{year}{2022}\natexlab{b}.
\newblock \showarticletitle{Towards Diverse and Natural Scene-aware 3D Human
  Motion Synthesis}. In \bibinfo{booktitle}{\emph{CVPR}}.
\newblock


\bibitem[\protect\citeauthoryear{Wang, Xu, Xu, Liu, and Wang}{Wang
  et~al\mbox{.}}{2021}]%
        {wang2021synthesizing}
\bibfield{author}{\bibinfo{person}{Jiashun Wang}, \bibinfo{person}{Huazhe Xu},
  \bibinfo{person}{Jingwei Xu}, \bibinfo{person}{Sifei Liu}, {and}
  \bibinfo{person}{Xiaolong Wang}.} \bibinfo{year}{2021}\natexlab{}.
\newblock \showarticletitle{Synthesizing long-term 3d human motion and
  interaction in 3d scenes}. In \bibinfo{booktitle}{\emph{CVPR}}.
\newblock


\bibitem[\protect\citeauthoryear{Wang, Zhang, Chen, Xu, Li, Liu, and Wang}{Wang
  et~al\mbox{.}}{2022c}]%
        {wang2022dexgraspnet}
\bibfield{author}{\bibinfo{person}{Ruicheng Wang}, \bibinfo{person}{Jialiang
  Zhang}, \bibinfo{person}{Jiayi Chen}, \bibinfo{person}{Yinzhen Xu},
  \bibinfo{person}{Puhao Li}, \bibinfo{person}{Tengyu Liu}, {and}
  \bibinfo{person}{He Wang}.} \bibinfo{year}{2022}\natexlab{c}.
\newblock \showarticletitle{Dexgraspnet: A large-scale robotic dexterous grasp
  dataset for general objects based on simulation}.
\newblock \bibinfo{journal}{\emph{arXiv preprint arXiv:2210.02697}}
  (\bibinfo{year}{2022}).
\newblock


\bibitem[\protect\citeauthoryear{Wang, Lin, Zeng, Luo, Zhang, and Zhang}{Wang
  et~al\mbox{.}}{2023}]%
        {wang2023physhoi}
\bibfield{author}{\bibinfo{person}{Yinhuai Wang}, \bibinfo{person}{Jing Lin},
  \bibinfo{person}{Ailing Zeng}, \bibinfo{person}{Zhengyi Luo},
  \bibinfo{person}{Jian Zhang}, {and} \bibinfo{person}{Lei Zhang}.}
  \bibinfo{year}{2023}\natexlab{}.
\newblock \showarticletitle{Physhoi: Physics-based imitation of dynamic
  human-object interaction}.
\newblock \bibinfo{journal}{\emph{arXiv preprint arXiv:2312.04393}}
  (\bibinfo{year}{2023}).
\newblock


\bibitem[\protect\citeauthoryear{Wang, Chen, Liu, Zhu, Liang, and Huang}{Wang
  et~al\mbox{.}}{2022a}]%
        {wang2022humanise}
\bibfield{author}{\bibinfo{person}{Zan Wang}, \bibinfo{person}{Yixin Chen},
  \bibinfo{person}{Tengyu Liu}, \bibinfo{person}{Yixin Zhu},
  \bibinfo{person}{Wei Liang}, {and} \bibinfo{person}{Siyuan Huang}.}
  \bibinfo{year}{2022}\natexlab{a}.
\newblock \showarticletitle{Humanise: Language-conditioned human motion
  generation in 3d scenes}.
\newblock \bibinfo{journal}{\emph{NeurIPS}} (\bibinfo{year}{2022}).
\newblock


\bibitem[\protect\citeauthoryear{Wang, Li, Sui, Zhou, Jiang, Nie, and Liu}{Wang
  et~al\mbox{.}}{2025}]%
        {wang2025stablemotion}
\bibfield{author}{\bibinfo{person}{Ziyi Wang}, \bibinfo{person}{Haipeng Li},
  \bibinfo{person}{Lin Sui}, \bibinfo{person}{Tianhao Zhou},
  \bibinfo{person}{Hai Jiang}, \bibinfo{person}{Lang Nie}, {and}
  \bibinfo{person}{Shuaicheng Liu}.} \bibinfo{year}{2025}\natexlab{}.
\newblock \showarticletitle{StableMotion: Repurposing Diffusion-Based Image
  Priors for Motion Estimation}.
\newblock \bibinfo{journal}{\emph{arXiv preprint arXiv:2505.06668}}
  (\bibinfo{year}{2025}).
\newblock


\bibitem[\protect\citeauthoryear{Won, Gopinath, and Hodgins}{Won
  et~al\mbox{.}}{2021}]%
        {won2021control}
\bibfield{author}{\bibinfo{person}{Jungdam Won}, \bibinfo{person}{Deepak
  Gopinath}, {and} \bibinfo{person}{Jessica Hodgins}.}
  \bibinfo{year}{2021}\natexlab{}.
\newblock \showarticletitle{Control strategies for physically simulated
  characters performing two-player competitive sports}.
\newblock \bibinfo{journal}{\emph{ACM Transactions on Graphics (TOG)}}
  (\bibinfo{year}{2021}).
\newblock


\bibitem[\protect\citeauthoryear{Won, Lee, O'Sullivan, Hodgins, and Lee}{Won
  et~al\mbox{.}}{2014}]%
        {won2014generating}
\bibfield{author}{\bibinfo{person}{Jungdam Won}, \bibinfo{person}{Kyungho Lee},
  \bibinfo{person}{Carol O'Sullivan}, \bibinfo{person}{Jessica~K Hodgins},
  {and} \bibinfo{person}{Jehee Lee}.} \bibinfo{year}{2014}\natexlab{}.
\newblock \showarticletitle{Generating and ranking diverse multi-character
  interactions}.
\newblock \bibinfo{journal}{\emph{ACM Transactions on Graphics (TOG)}}
  (\bibinfo{year}{2014}).
\newblock


\bibitem[\protect\citeauthoryear{Wu, Li, Xu, and Liu}{Wu et~al\mbox{.}}{2025}]%
        {wu2025hoifhli}
\bibfield{author}{\bibinfo{person}{Zhen Wu}, \bibinfo{person}{Jiaman Li},
  \bibinfo{person}{Pei Xu}, {and} \bibinfo{person}{C~Karen Liu}.}
  \bibinfo{year}{2025}\natexlab{}.
\newblock \showarticletitle{Human-object interaction from human-level
  instructions}. In \bibinfo{booktitle}{\emph{ICCV}}.
\newblock


\bibitem[\protect\citeauthoryear{Xie, Jampani, Zhong, Sun, and Jiang}{Xie
  et~al\mbox{.}}{2024}]%
        {xie2024omnicontrol}
\bibfield{author}{\bibinfo{person}{Yiming Xie}, \bibinfo{person}{Varun
  Jampani}, \bibinfo{person}{Lei Zhong}, \bibinfo{person}{Deqing Sun}, {and}
  \bibinfo{person}{Huaizu Jiang}.} \bibinfo{year}{2024}\natexlab{}.
\newblock \showarticletitle{OmniControl: Control Any Joint at Any Time for
  Human Motion Generation}. In \bibinfo{booktitle}{\emph{ICLR}}.
\newblock


\bibitem[\protect\citeauthoryear{Xu, Lv, Yan, Jin, Wu, Xu, Liu, Zhou, Rao,
  Sheng, et~al\mbox{.}}{Xu et~al\mbox{.}}{2024}]%
        {xu2024interx}
\bibfield{author}{\bibinfo{person}{Liang Xu}, \bibinfo{person}{Xintao Lv},
  \bibinfo{person}{Yichao Yan}, \bibinfo{person}{Xin Jin},
  \bibinfo{person}{Shuwen Wu}, \bibinfo{person}{Congsheng Xu},
  \bibinfo{person}{Yifan Liu}, \bibinfo{person}{Yizhou Zhou},
  \bibinfo{person}{Fengyun Rao}, \bibinfo{person}{Xingdong Sheng},
  {et~al\mbox{.}}} \bibinfo{year}{2024}\natexlab{}.
\newblock \showarticletitle{Inter-x: Towards versatile human-human interaction
  analysis}. In \bibinfo{booktitle}{\emph{CVPR}}.
\newblock


\bibitem[\protect\citeauthoryear{Xu, Wu, Wang, Sarukkai, Fatahalian,
  Karamouzas, Zordan, and Liu}{Xu et~al\mbox{.}}{2025b}]%
        {xu2025learningtoball}
\bibfield{author}{\bibinfo{person}{Pei Xu}, \bibinfo{person}{Zhen Wu},
  \bibinfo{person}{Ruocheng Wang}, \bibinfo{person}{Vishnu Sarukkai},
  \bibinfo{person}{Kayvon Fatahalian}, \bibinfo{person}{Ioannis Karamouzas},
  \bibinfo{person}{Victor Zordan}, {and} \bibinfo{person}{C~Karen Liu}.}
  \bibinfo{year}{2025}\natexlab{b}.
\newblock \showarticletitle{Learning to Ball: Composing Policies for
  Long-Horizon Basketball Moves}.
\newblock \bibinfo{journal}{\emph{ACM Transactions on Graphics (TOG)}}
  (\bibinfo{year}{2025}).
\newblock


\bibitem[\protect\citeauthoryear{Xu, Li, Wang, and Gui}{Xu
  et~al\mbox{.}}{2023}]%
        {xu2023interdiff}
\bibfield{author}{\bibinfo{person}{Sirui Xu}, \bibinfo{person}{Zhengyuan Li},
  \bibinfo{person}{Yu-Xiong Wang}, {and} \bibinfo{person}{Liang-Yan Gui}.}
  \bibinfo{year}{2023}\natexlab{}.
\newblock \showarticletitle{{InterDiff}: Generating 3D Human-Object
  Interactions with Physics-Informed Diffusion}. In
  \bibinfo{booktitle}{\emph{ICCV}}.
\newblock


\bibitem[\protect\citeauthoryear{Xu, Ling, Wang, and Gui}{Xu
  et~al\mbox{.}}{2025a}]%
        {xu2025intermimic}
\bibfield{author}{\bibinfo{person}{Sirui Xu}, \bibinfo{person}{Hung~Yu Ling},
  \bibinfo{person}{Yu-Xiong Wang}, {and} \bibinfo{person}{Liang-Yan Gui}.}
  \bibinfo{year}{2025}\natexlab{a}.
\newblock \showarticletitle{Intermimic: Towards universal whole-body control
  for physics-based human-object interactions}.
\newblock \bibinfo{journal}{\emph{arXiv preprint arXiv:2502.20390}}
  (\bibinfo{year}{2025}).
\newblock


\bibitem[\protect\citeauthoryear{Ye, Pavlakos, Malik, and Kanazawa}{Ye
  et~al\mbox{.}}{2023}]%
        {ye2023slahmr}
\bibfield{author}{\bibinfo{person}{Vickie Ye}, \bibinfo{person}{Georgios
  Pavlakos}, \bibinfo{person}{Jitendra Malik}, {and} \bibinfo{person}{Angjoo
  Kanazawa}.} \bibinfo{year}{2023}\natexlab{}.
\newblock \showarticletitle{Decoupling Human and Camera Motion from Videos in
  the Wild}. In \bibinfo{booktitle}{\emph{CVPR}}.
\newblock


\bibitem[\protect\citeauthoryear{Ye and Liu}{Ye and Liu}{2012}]%
        {ye2012synthesishand}
\bibfield{author}{\bibinfo{person}{Yuting Ye} {and} \bibinfo{person}{C~Karen
  Liu}.} \bibinfo{year}{2012}\natexlab{}.
\newblock \showarticletitle{Synthesis of detailed hand manipulations using
  contact sampling}.
\newblock \bibinfo{journal}{\emph{ACM Transactions on Graphics (ToG)}}
  (\bibinfo{year}{2012}).
\newblock


\bibitem[\protect\citeauthoryear{Yu, Wang, Zhao, Tsui, Wang, Tan, and Chen}{Yu
  et~al\mbox{.}}{2025}]%
        {yu2025skillmimicv2}
\bibfield{author}{\bibinfo{person}{Runyi Yu}, \bibinfo{person}{Yinhuai Wang},
  \bibinfo{person}{Qihan Zhao}, \bibinfo{person}{Hok~Wai Tsui},
  \bibinfo{person}{Jingbo Wang}, \bibinfo{person}{Ping Tan}, {and}
  \bibinfo{person}{Qifeng Chen}.} \bibinfo{year}{2025}\natexlab{}.
\newblock \showarticletitle{Skillmimic-v2: Learning robust and generalizable
  interaction skills from sparse and noisy demonstrations}. In
  \bibinfo{booktitle}{\emph{SIGGRAPH 2025 Conference Papers}}.
\newblock


\bibitem[\protect\citeauthoryear{Zhang, Ye, Shiratori, and Komura}{Zhang
  et~al\mbox{.}}{2021a}]%
        {zhang2021manipnet}
\bibfield{author}{\bibinfo{person}{He Zhang}, \bibinfo{person}{Yuting Ye},
  \bibinfo{person}{Takaaki Shiratori}, {and} \bibinfo{person}{Taku Komura}.}
  \bibinfo{year}{2021}\natexlab{a}.
\newblock \showarticletitle{Manipnet: neural manipulation synthesis with a
  hand-object spatial representation}.
\newblock \bibinfo{journal}{\emph{ACM Transactions on Graphics (ToG)}}
  (\bibinfo{year}{2021}).
\newblock


\bibitem[\protect\citeauthoryear{Zhang, Zhang, Song, Shi, Zhao, Shi, Yu, Xu,
  and Wang}{Zhang et~al\mbox{.}}{2024c}]%
        {zhang2024hoim3}
\bibfield{author}{\bibinfo{person}{Juze Zhang}, \bibinfo{person}{Jingyan
  Zhang}, \bibinfo{person}{Zining Song}, \bibinfo{person}{Zhanhe Shi},
  \bibinfo{person}{Chengfeng Zhao}, \bibinfo{person}{Ye Shi},
  \bibinfo{person}{Jingyi Yu}, \bibinfo{person}{Lan Xu}, {and}
  \bibinfo{person}{Jingya Wang}.} \bibinfo{year}{2024}\natexlab{c}.
\newblock \showarticletitle{HOI-M3: Capture Multiple Humans and Objects
  Interaction within Contextual Environment}.
\newblock \bibinfo{journal}{\emph{arXiv preprint arXiv:2404.00299}}
  (\bibinfo{year}{2024}).
\newblock


\bibitem[\protect\citeauthoryear{Zhang, Bhatnagar, Xu, Winkler, Kadlecek, Tang,
  and Bogo}{Zhang et~al\mbox{.}}{2024b}]%
        {zhang2024rohm}
\bibfield{author}{\bibinfo{person}{Siwei Zhang}, \bibinfo{person}{Bharat~Lal
  Bhatnagar}, \bibinfo{person}{Yuanlu Xu}, \bibinfo{person}{Alexander Winkler},
  \bibinfo{person}{Petr Kadlecek}, \bibinfo{person}{Siyu Tang}, {and}
  \bibinfo{person}{Federica Bogo}.} \bibinfo{year}{2024}\natexlab{b}.
\newblock \showarticletitle{RoHM: Robust Human Motion Reconstruction via
  Diffusion}. In \bibinfo{booktitle}{\emph{CVPR}}.
\newblock


\bibitem[\protect\citeauthoryear{Zhang, Zhang, Bogo, Pollefeys, and Tang}{Zhang
  et~al\mbox{.}}{2021b}]%
        {zhang2021lemo}
\bibfield{author}{\bibinfo{person}{Siwei Zhang}, \bibinfo{person}{Yan Zhang},
  \bibinfo{person}{Federica Bogo}, \bibinfo{person}{Marc Pollefeys}, {and}
  \bibinfo{person}{Siyu Tang}.} \bibinfo{year}{2021}\natexlab{b}.
\newblock \showarticletitle{Learning motion priors for 4d human body capture in
  3d scenes}. In \bibinfo{booktitle}{\emph{ICCV}}.
\newblock


\bibitem[\protect\citeauthoryear{Zhang, Bhatnagar, Starke, Guzov, and
  Pons-Moll}{Zhang et~al\mbox{.}}{2022}]%
        {zhang2022couch}
\bibfield{author}{\bibinfo{person}{Xiaohan Zhang}, \bibinfo{person}{Bharat~Lal
  Bhatnagar}, \bibinfo{person}{Sebastian Starke}, \bibinfo{person}{Vladimir
  Guzov}, {and} \bibinfo{person}{Gerard Pons-Moll}.}
  \bibinfo{year}{2022}\natexlab{}.
\newblock \showarticletitle{COUCH: Towards Controllable Human-Chair
  Interactions}. In \bibinfo{booktitle}{\emph{ECCV}}.
\newblock


\bibitem[\protect\citeauthoryear{Zhang, Bhatnagar, Starke, Petrov, Guzov,
  Dhamo, P{\'e}rez-Pellitero, and Pons-Moll}{Zhang et~al\mbox{.}}{2024a}]%
        {zhang2024force}
\bibfield{author}{\bibinfo{person}{Xiaohan Zhang}, \bibinfo{person}{Bharat~Lal
  Bhatnagar}, \bibinfo{person}{Sebastian Starke}, \bibinfo{person}{Ilya
  Petrov}, \bibinfo{person}{Vladimir Guzov}, \bibinfo{person}{Helisa Dhamo},
  \bibinfo{person}{Eduardo P{\'e}rez-Pellitero}, {and} \bibinfo{person}{Gerard
  Pons-Moll}.} \bibinfo{year}{2024}\natexlab{a}.
\newblock \showarticletitle{FORCE: Physics-aware Human-object Interaction}.
\newblock \bibinfo{journal}{\emph{arXiv preprint arXiv:2403.11237}}
  (\bibinfo{year}{2024}).
\newblock


\bibitem[\protect\citeauthoryear{Zhang, Gopinath, Ye, Hodgins, Turk, and
  Won}{Zhang et~al\mbox{.}}{2023}]%
        {zhang2023ig}
\bibfield{author}{\bibinfo{person}{Yunbo Zhang}, \bibinfo{person}{Deepak
  Gopinath}, \bibinfo{person}{Yuting Ye}, \bibinfo{person}{Jessica Hodgins},
  \bibinfo{person}{Greg Turk}, {and} \bibinfo{person}{Jungdam Won}.}
  \bibinfo{year}{2023}\natexlab{}.
\newblock \showarticletitle{Simulation and retargeting of complex
  multi-character interactions}. In \bibinfo{booktitle}{\emph{ACM SIGGRAPH 2023
  Conference Proceedings}}.
\newblock


\bibitem[\protect\citeauthoryear{Zheng, Yang, Mo, Li, Yu, Liu, Liu, and
  Guibas}{Zheng et~al\mbox{.}}{2022}]%
        {zheng2022gimo}
\bibfield{author}{\bibinfo{person}{Yang Zheng}, \bibinfo{person}{Yanchao Yang},
  \bibinfo{person}{Kaichun Mo}, \bibinfo{person}{Jiaman Li},
  \bibinfo{person}{Tao Yu}, \bibinfo{person}{Yebin Liu},
  \bibinfo{person}{C~Karen Liu}, {and} \bibinfo{person}{Leonidas~J Guibas}.}
  \bibinfo{year}{2022}\natexlab{}.
\newblock \showarticletitle{Gimo: Gaze-informed human motion prediction in
  context}. In \bibinfo{booktitle}{\emph{ECCV}}.
\newblock


\bibitem[\protect\citeauthoryear{Zhou, Bhatnagar, Lenssen, and Pons-Moll}{Zhou
  et~al\mbox{.}}{2022}]%
        {zhou2022toch}
\bibfield{author}{\bibinfo{person}{Keyang Zhou}, \bibinfo{person}{Bharat~Lal
  Bhatnagar}, \bibinfo{person}{Jan~Eric Lenssen}, {and} \bibinfo{person}{Gerard
  Pons-Moll}.} \bibinfo{year}{2022}\natexlab{}.
\newblock \showarticletitle{Toch: Spatio-temporal object-to-hand correspondence
  for motion refinement}. In \bibinfo{booktitle}{\emph{ECCV}}.
\newblock


\end{thebibliography}
\end{document}